\documentclass{article} % For LaTeX2e
\usepackage{iclr2027_conference,times}

\usepackage{amsmath,amsfonts,bm}

\def\eqref#1{equation~\ref{#1}}
\def\1{\bm{1}}

\DeclareMathAlphabet{\mathsfit}{\encodingdefault}{\sfdefault}{m}{sl}
\SetMathAlphabet{\mathsfit}{bold}{\encodingdefault}{\sfdefault}{bx}{n}

\usepackage{hyperref}
\usepackage{url}
\usepackage{booktabs} % \toprule, \midrule, \bottomrule, \cmidrule
\usepackage{multirow} % \multirow
\usepackage{graphicx} % \resizebox
\usepackage{lipsum}
\usepackage{xspace}
\usepackage{pifont}
\usepackage[table]{xcolor}
\usepackage{array}
\usepackage{makecell}
\usepackage{duckuments}
\usepackage{algorithm}
\usepackage{algpseudocode}
\usepackage{subcaption}
\usepackage{wrapfig}
\usepackage{caption}
\usepackage[most]{tcolorbox}
\usepackage{float}
\usepackage{tabularx}
\usepackage{enumitem}

\newcolumntype{Y}{>{\centering\arraybackslash}X}

\title{PLAYGROUND: Progressive Layout\\Generation with Render-Grounded Planning}

\author{\textbf{Jaeho Lee} \quad
\textbf{Junhwan Heo} \quad
\textbf{Junghyun Park} \quad
\textbf{WonJun Moon} \quad
\textbf{Eunju Yang} \\
\textbf{Seungho Jang} \quad
\textbf{Seongchan Kim} \quad
\textbf{Seungryong Kim}$^{\dagger}$ \\[4pt]
KAIST AI \\[4pt]
\url{https://cvlab-kaist.github.io/PlayGround/}
}

\newcommand{\ours}{\mbox{PlayGround}\xspace}

\definecolor{hlours}{HTML}{e8f0f8}
\newcommand{\hlrow}{\rowcolor{hlours}}
\newcommand{\metricrowstrut}{\rule[-0.75ex]{0pt}{2.8ex}}
\newcommand{\datasetseparator}{\specialrule{0.65pt}{0pt}{0pt}}

\usepackage{pifont}
\newcommand{\cmark}{\ding{51}}
\newcommand{\xmark}{\ding{55}}

\newtcolorbox{vlmpromptbox_stage1}{
    enhanced,
    breakable,
    colback=black!1,
    colframe=black!35,
    colbacktitle=black!8,
    coltitle=black,
    title={Prompt Used for Stage 1},
    fonttitle=\bfseries,
    boxrule=0.5pt,
    arc=1.5mm,
    left=2mm,
    right=2mm,
    top=1.5mm,
    bottom=1.5mm,
    before skip=6pt,
    after skip=6pt,
    before upper={\footnotesize\ttfamily\raggedright\sloppy}
}

\newtcolorbox{vlmpromptbox_stage2}{
    enhanced,
    breakable,
    colback=black!1,
    colframe=black!35,
    colbacktitle=black!8,
    coltitle=black,
    title={Prompt Used for Stage 2},
    fonttitle=\bfseries,
    boxrule=0.5pt,
    arc=1.5mm,
    left=2mm,
    right=2mm,
    top=1.5mm,
    bottom=1.5mm,
    before skip=6pt,
    after skip=6pt,
    before upper={\footnotesize\ttfamily\raggedright\sloppy}
}

\newtcolorbox{vlmpromptbox_retriever}{
    enhanced,
    breakable,
    colback=black!1,
    colframe=black!35,
    colbacktitle=black!8,
    coltitle=black,
    title={Prompt Used for Structure-Aware Retriever},
    fonttitle=\bfseries,
    boxrule=0.5pt,
    arc=1.5mm,
    left=2mm,
    right=2mm,
    top=1.5mm,
    bottom=1.5mm,
    before skip=6pt,
    after skip=6pt,
    before upper={\fontsize{6.5}{7.2}\selectfont\ttfamily\raggedright\sloppy}
}

\newtcolorbox{vlmpromptbox_planner}{
    enhanced,
    breakable,
    colback=black!1,
    colframe=black!35,
    colbacktitle=black!8,
    coltitle=black,
    title={Prompt Used for Progressive Planner},
    fonttitle=\bfseries,
    boxrule=0.5pt,
    arc=1.5mm,
    left=2mm,
    right=2mm,
    top=1.5mm,
    bottom=1.5mm,
    before skip=6pt,
    after skip=6pt,
    before upper={\fontsize{6.5}{7.2}\selectfont\ttfamily\raggedright\sloppy}
}

\newtcolorbox{vlmpromptbox_generator}{
    enhanced,
    breakable,
    colback=black!1,
    colframe=black!35,
    colbacktitle=black!8,
    coltitle=black,
    title={Prompt Used for Layout Generator},
    fonttitle=\bfseries,
    boxrule=0.5pt,
    arc=1.5mm,
    left=2mm,
    right=2mm,
    top=1.5mm,
    bottom=1.5mm,
    before skip=6pt,
    after skip=6pt,
    before upper={\fontsize{6.5}{7.2}\selectfont\ttfamily\raggedright\sloppy}
}

\newtcolorbox{vlmpromptbox_verifier}{
    enhanced,
    breakable,
    colback=black!1,
    colframe=black!35,
    colbacktitle=black!8,
    coltitle=black,
    title={Prompt Used for Visual Verifier},
    fonttitle=\bfseries,
    boxrule=0.5pt,
    arc=1.5mm,
    left=2mm,
    right=2mm,
    top=1.5mm,
    bottom=1.5mm,
    before skip=6pt,
    after skip=6pt,
    before upper={\fontsize{6.5}{7.2}\selectfont\ttfamily\raggedright\sloppy}
}

\newtcolorbox{vlmpromptbox_vlm_eval}{
    enhanced,
    breakable,
    colback=black!1,
    colframe=black!35,
    colbacktitle=black!8,
    coltitle=black,
    title={Prompt Used for VLM Evaluation},
    fonttitle=\bfseries,
    boxrule=0.5pt,
    arc=1.5mm,
    left=2mm,
    right=2mm,
    top=1.5mm,
    bottom=1.5mm,
    before skip=6pt,
    after skip=6pt,
    before upper={
        \fontsize{6.5}{7.2}\selectfont
        \ttfamily
        \raggedright
        \sloppy
        \setlength{\parindent}{0pt}
        \setlength{\parskip}{1.5pt}
    }
}

\newtcolorbox{vlmpromptbox_i2d}{
    enhanced,
    breakable,
    colback=black!1,
    colframe=black!35,
    colbacktitle=black!8,
    coltitle=black,
    title={Prompt Used for Element Generation},
    fonttitle=\bfseries,
    boxrule=0.5pt,
    arc=1.5mm,
    left=2mm,
    right=2mm,
    top=1.5mm,
    bottom=1.5mm,
    before skip=6pt,
    after skip=6pt,
    before upper={\footnotesize\ttfamily\raggedright\sloppy}
}

\iclrfinalcopy % Uncomment for camera-ready version, but NOT for submission.
\begin{document}

\maketitle
\begin{figure}[h]
    \centering
    \includegraphics[width=1.0\linewidth]{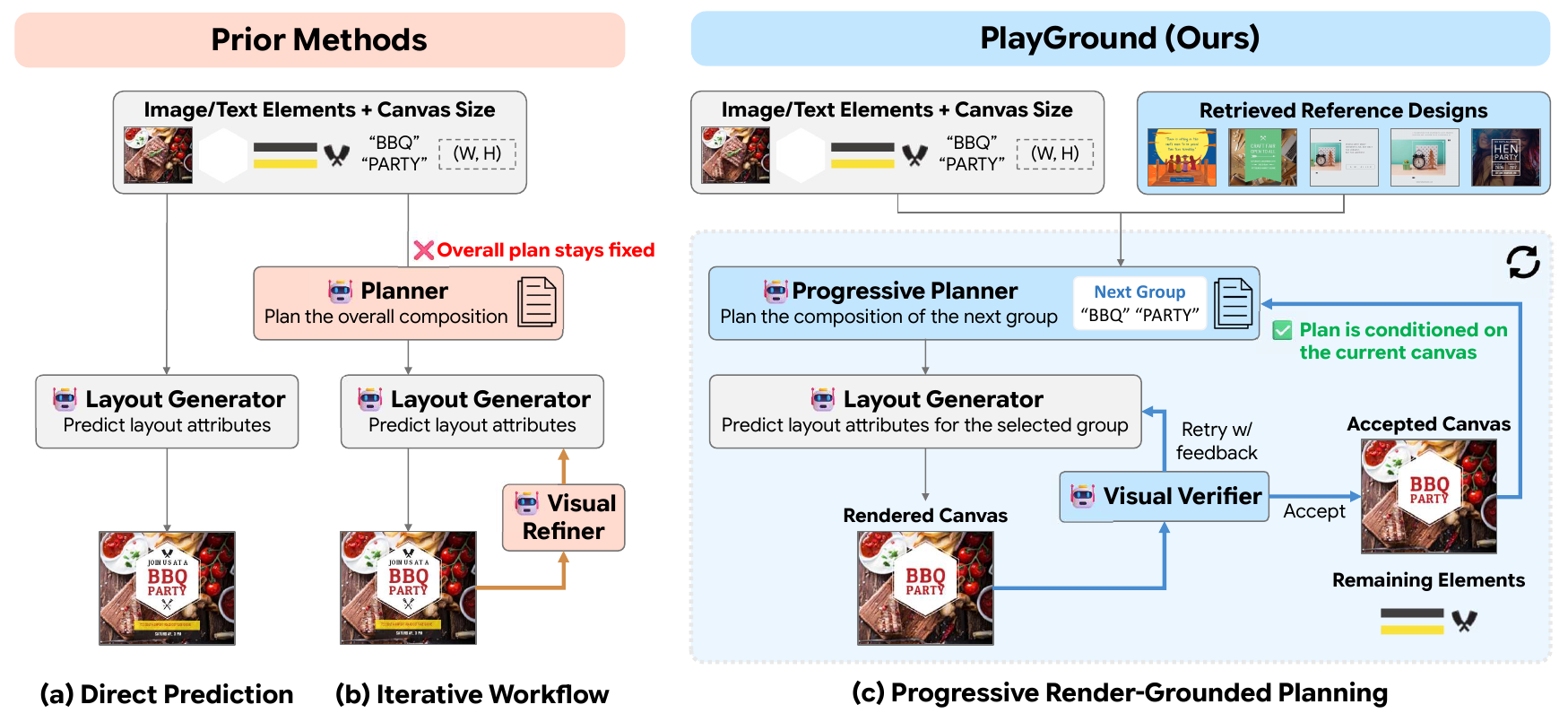}
    \vspace{-15pt}
    \caption{Prior methods~\citep{cheng2024graphicdesignlargemultimodal,liu2026designascode} either (a) generate layout in a single forward pass, or (b) refine results iteratively while overall design plan stays fixed outside the feedback loop. 
    In contrast, (c) \textbf{\ours} decomposes planning into steps and places them inside the iterative loop. 
    As \ours plans and fills the canvas progressively based on the current canvas state, \ours is robust to intermediate design errors and produces more coherent final designs.
    % \textbf{Teaser.} Prior methods either generate layouts without an iterative workflow (a), or refine generated results iteratively while keeping the Planner outside the feedback loop (b). In contrast, \ours incorporates planning into the iterative process, repeatedly planning and generating functional groups based on the current rendered canvas (c).
    }
    \label{fig:teaser}
    \vspace{-5pt}
\end{figure}
\begin{abstract}
\label{sec:abstract}

Generating editable graphic designs requires arranging visual and textual elements into a coherent composition while preserving their content and editability. 
Although recent MLLM-based approaches employ iterative generation and visual refinement to reflect the real-world design workflows, their high-level plans are predetermined prior to execution, limiting their adaptability as the design evolves. 
To address this limitation, we propose \textbf{PlayGround}, a progressive layout generation framework that dynamically adapts to the current rendered canvas.
PlayGround establishes a progressive design loop comprising a Progressive Planner, a Layout Generator, and a Visual Verifier, guided by a Structure-Aware Retriever. 
Given an initial set of unordered elements, the Structure-Aware Retriever initially infers a coarse layout structure to retrieve structurally similar reference designs. 
Subsequently, at each turn, the Progressive Planner examines the current rendered canvas alongside the remaining elements, forms the next functional group, and plans its composition. 
The Layout Generator then predicts editable layout attributes, while the Visual Verifier evaluates rendered candidates and provides feedback when refinement is necessary. 
To train the Progressive Planner with step-wise supervision, we construct a two-stage data generation pipeline that extracts functional groups and turn-level planning trajectories from completed designs. 
Experiments on Crello and LICA demonstrate the effectiveness of \textbf{PlayGround} over existing methods. 
Our code and weights will be publicly released.
% Generating editable graphic designs requires arranging visual and textual elements into a coherent composition while preserving their content and editability. 
% Recent MLLM-based approaches increasingly use iterative generation and visual refinement, but fix the overall plan or element grouping upfront, limiting adaptation to the evolving design. 
% To address this limitation, we propose \textbf{PlayGround}, a progressive layout generation framework with render-grounded planning. 
% At each turn, a Progressive Planner observes the current rendered canvas and remaining elements, forms the next functional group, and plans its composition. 
% A Layout Generator predicts editable layout attributes, while a Visual Verifier evaluates rendered candidates and provides feedback when needed. 
% We further introduce a Structure-Aware Retriever that infers coarse layout structure from unordered elements to retrieve structurally similar references. 
% To obtain step-wise supervision for progressive planning, we develop a two-stage data construction pipeline that derives functional groups and turn-level plans from completed designs. 
% Experiments on Crello and LICA demonstrate improved graphic design quality and generalization over existing methods, with ablations validating our key components. Our code and weights will be publicly released.
\end{abstract}
\section{Introduction}
\label{sec:intro}

Graphic design is a key medium for effectively communicating visual and textual information across domains such as advertising, e-commerce, and presentation authoring, yet its production requires substantial time and expertise. 
% Graphic design is a key medium for effectively communicating visual and textual information across domains such as advertising, branding, e-commerce, and presentation authoring, yet its production requires substantial time and expertise. 
To automate this process, AI-based approaches have been extensively studied along two main directions.
% have broadly developed along two directions. 
One directly synthesizes the final design as a pixel-level image using diffusion-based generative models~\citep{li2023planning, gao2025postermaker, wang2025designdiffusion}, while the other predicts layouts of given image and text elements to produce an editable document~\citep{inoue2023flexiblemultimodaldocumentmodels, inoue2024opencolereproducibleautomaticgraphic}. 
Although text-to-image generative models achieve high visual quality, they still struggle with accurate text rendering and faithful preservation of input elements, and their raster outputs limit element-level editing. 
In contrast, layout generation models represent designs in structured formats such as JSON or HTML/CSS, supporting content fidelity and editability.

Recently, Multimodal Large Language Models~(MLLMs) have been adopted for editable layout generation due to their strong multimodal understanding and reasoning capabilities.
While pioneering works~\citep{cheng2024graphicdesignlargemultimodal, lin2024elementsdesignlayeredapproach, patnaik2025aesthetiqenhancinggraphiclayout} exploited MLLMs to directly predict layout attributes such as position, size, and ordering from multimodal elements (Fig.~\ref{fig:teaser}~(a)), recent works aim to reflect the iterative nature of real-world design workflows.
PosterCopilot~\citep{wei2025postercopilotlayoutreasoningcontrollable} enabled iterative editing and rearrangement of generated designs, while VFLM~\citep{guo2026seeingimprovingvisualfeedback} progressively refines text layouts using visual feedback from rendered results.
Other works introduced an explicit planning stage into the design workflow.
DesignAsCode~\citep{liu2026designascode} combines semantic planning, HTML/CSS generation, and visual refinement, while PSDesigner~\citep{shuai2026psdesignerautomatedgraphicdesign} integrates element collection and grouping with sequential placement and refinement.
Although these works have successfully enhanced the overall design quality, their workflows rely heavily on initial planning and element grouping decisions that cannot be revised using intermediate visual feedback (Fig.~\ref{fig:teaser}~(b)). 

Therefore, we propose \textbf{\ours} (\textbf{P}rogressive \textbf{Lay}out
Generation with Render-\textbf{Ground}ed Planning), a framework that
adaptively adjusts the design workflow based on the current rendered
layout, as illustrated in Fig.~\ref{fig:teaser}~(c).
% , where each step is conditioned on the current canvas state rather than on a fixed one-shot plan (Fig.~\ref{fig:teaser}~(c)).
Initially, given the canvas size and a set of visual and text elements, a Structure-Aware Retriever first infers a coarse layout structure from unordered elements and retrieves reference designs based on structural similarity.
Conditioned on the retrieved references, \ours then completes the design through an iterative workflow of a Progressive Planner, a Layout Generator, and a Visual Verifier.

% To enable dynamic planning, we let the Progressive Planner observe the current canvas and the remaining elements at every turn.
% We define the unit of each turn as a functional group, a set of elements that share the same role in the design, rather than an arbitrary subset.
% The Planner then selects the next functional group and plans its composition, conditioned on the current canvas.
% To enable dynamic planning, we let the Progressive Planner observe the current canvas and the remaining elements at every turn, and decide which elements to place in the current turn as a group.
% Yet, since arbitrary grouping leaves it ambiguous which elements should be placed together, we define the unit of each turn as a functional group, a set of elements that share the same role in the design. The Progressive Planner then forms a functional group and plans its composition with respect to the current canvas.
To enable dynamic planning, the Progressive Planner observes the current canvas and the remaining elements at each turn and determines which elements to place together.
However, leaving the Planner to form arbitrary groups provides no clear criterion for deciding which elements should be grouped, potentially leading to ambiguous grouping decisions.
We therefore predefine the unit of each turn as a functional group, a set of elements that share the same role~(\textit{e.g.,} background, title) in the design, and train the Planner to identify such groups.
Consequently, at each turn, the Progressive Planner selects a functional group and plans its composition in the context of the current canvas.

Subsequently, the Layout Generator realizes the plan by predicting the layout attributes of the elements in the group, and the Visual Verifier inspects the rendered result, returning feedback for the Layout Generator to revise its predictions when necessary.
The accepted canvas is then passed back to the Planner, so that later grouping and planning decisions adapt to the actual partial design rather than to an assumed one. 
By decomposing the design process into a progressive procedure, our experimental results validate that \ours produces high-quality and consistent final designs.

Such progressive planning, however, requires step-wise supervision which does not exist in conventional layout datasets.
Therefore, we additionally develop a two-stage data construction pipeline, which decomposes the layout designs into functional groups and generates supervision for group-wise turn-level planning.
Together with the original layout attributes, these annotations supervise the Structure-Aware Retriever, Progressive Planner, and Layout Generator.

Our contributions are summarized as follows:
\begin{itemize}[
    topsep=0pt,
    itemsep=3pt,
    parsep=0pt,
    partopsep=0pt
]
\item We propose \textbf{\ours}, a progressive layout generation framework that integrates dynamic planning and functional grouping to reflect the real-world design workflow.
\item We introduce a {Progressive Planner} that decomposes the complex design process into sequential steps and progressively plans each step based on the current rendered canvas.
\item We define {functional groups} as the planning unit, providing a principled grouping criterion by organizing elements that share the same role in the design.
\item We develop a two-stage data construction pipeline that derives group-wise, turn-level supervision from completed designs.
\item Experiments on Crello and LICA demonstrate improved design quality and zero-shot generalization, with ablations validating the effectiveness of our key components.
\end{itemize}
% \begin{itemize}    
%     \item We propose \textbf{\ours}, a progressive layout generation framework that adaptively adjusts the design plan and decomposes element placement into functional groups based on their roles in the design.
%     % that dynamically forms functional groups and plans their layouts based on the evolving canvas.
%     \item We introduce a Structure-Aware Retriever that infers coarse layout structure to retrieve structurally similar references from unordered elements.
%     \item We develop a two-stage data construction pipeline that derives group-wise turn-level supervision from completed designs.
%     \item Experiments on Crello and LICA demonstrate improved design quality and zero-shot generalization, with ablations validating our key components.
% \end{itemize}
\section{Related Work}
\label{sec:related_work}

Graphic design composition methods either arrange supplied visual and textual elements~\citep{inoue2023flexiblemultimodaldocumentmodels,lin2024elementsdesignlayeredapproach} or additionally generate and retrieve elements during the design process~\citep{jia2024colehierarchicalgenerationframework,inoue2024opencolereproducibleautomaticgraphic}; our work focuses on the former task. Moving beyond direct composition, iterative graphic design systems incorporate visual feedback to refine generated results. PosterCopilot~\citep{wei2025postercopilotlayoutreasoningcontrollable} combines layout reasoning with layer-controllable iterative editing, while VFLM~\citep{guo2026seeingimprovingvisualfeedback} progressively refines text layouts through a generation--rendering--reflection--refinement cycle. DesignAsCode~\citep{liu2026designascode} constructs a semantic plan, implements it as HTML/CSS, and subsequently refines the rendered design, while PSDesigner~\citep{shuai2026psdesignerautomatedgraphicdesign} combines element collection and planning with sequential placement and refinement. Despite these iterative mechanisms, their initial planning or grouping decisions remain fixed throughout generation. In contrast, \ours dynamically updates both decisions at each turn based on the current rendered canvas. Additional related works are discussed in Appendix~\ref{sec:additional_related_work}.

\section{Data Construction}
\label{sec:data_construction}

\begin{figure}[t]
    \centering
    \includegraphics[width=1.0\linewidth]{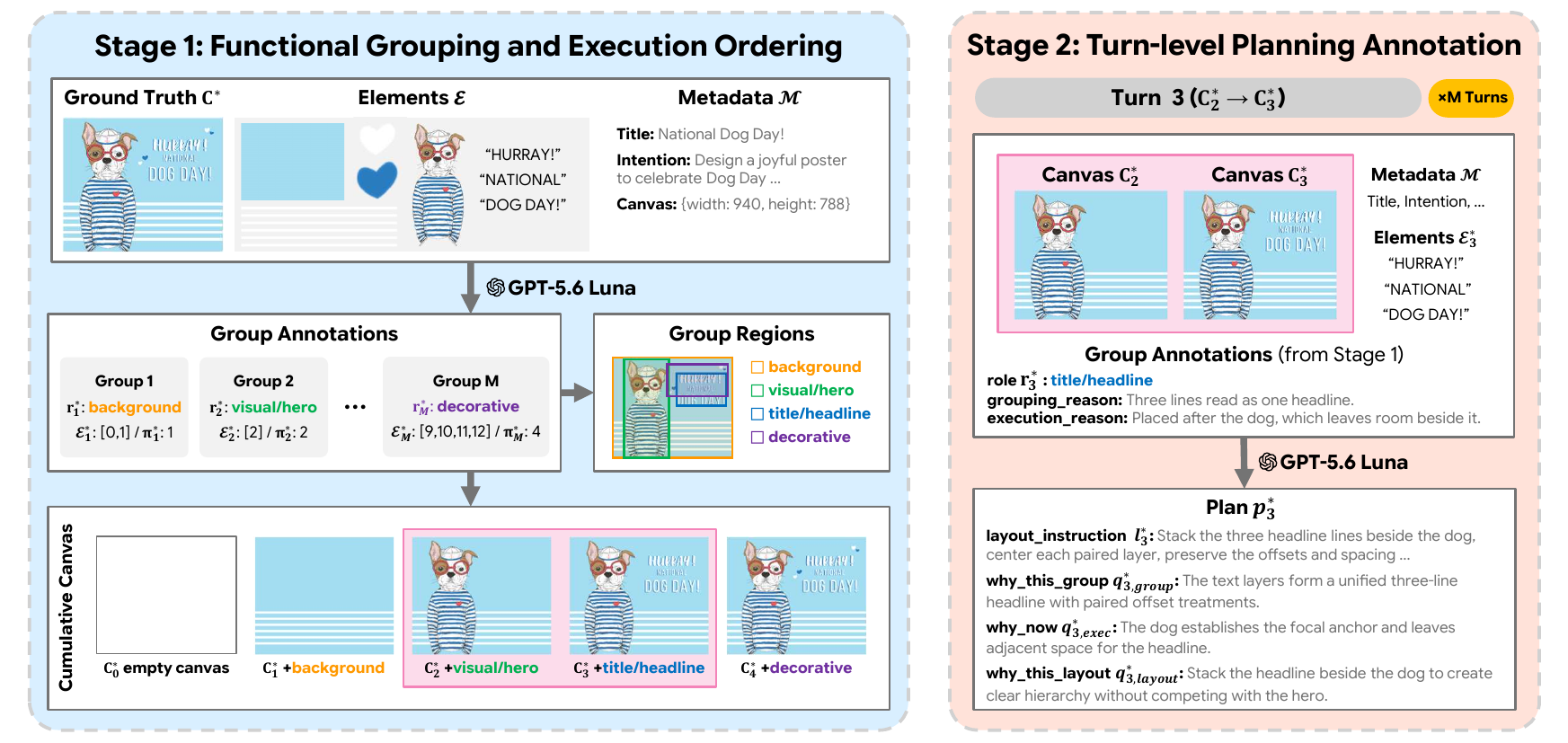}
    \vspace{-15pt}
    \caption{
    \textbf{Data Construction Pipeline.} Stage 1 constructs functional groups, execution order, coarse group regions, and cumulative canvas states from completed designs. Stage 2 uses these outputs to derive turn-level planning annotations for each functional group.
    }
    \label{fig:data_construction}
    \vspace{-10pt}
\end{figure}

To construct turn-level annotations for functional grouping and layout planning required by \ours, we build a two-stage data construction pipeline on Crello~\citep{yamaguchi2021canvasvaelearninggeneratevector} dataset using GPT-5.6 Luna~\citep{openai2026gpt56} as an MLLM annotator. 
For each design, we use the title from the Crello metadata and the intention annotation provided by OpenCOLE~\citep{inoue2024opencolereproducibleautomaticgraphic} as additional semantic context, together with the complete element inventory and rendered design.

\paragraph{Stage 1: Functional grouping and execution ordering.}
\label{sec:Data_construction_stage1}
We first decompose each design into functional groups and determine their execution order, as illustrated in Fig.~\ref{fig:data_construction}~(left).
A functional group is a set of elements that jointly serve a common design function and are intended to be planned and positioned together.
Each group is categorized into one of six roles: \texttt{background}, \texttt{visual/hero}, \texttt{title/headline}, \texttt{content}, \texttt{support}, or \texttt{decorative}.
A group may contain image elements, text elements, or both, and multiple groups may share the same role when they should be planned and placed independently.
Finally, groups that provide visual references for other groups are placed earlier in the execution order to serve as visual anchors.

Let $C^{*}$\footnote{The superscript \(^{*}\) denotes ground-truth-derived quantities and is omitted at inference.} denote the final ground-truth design,
$\mathcal{E}=\{e_i\}_{i=1}^{N}$ the complete element inventory,
and $\mathcal{M}$ the document metadata including the canvas size, title, and intention.
Given these inputs and the Stage~1 annotation prompt $\mathcal{P}_{\text{S1}}$,
the annotator assigns every element to a group, assigns each group a functional role,
and determines a dependency-aware execution order:
% We provide the annotator $(C^{*},\mathcal{E},\mathcal{M})$, together with the definition of each role and annotation instructions.
% The annotator assigns every element to a group, assigns each group a functional role, and determines a dependency-aware execution order. Denoting the Stage~1 prompt by $\mathcal{P}_\text{S1}$, we express this process as:
\begin{equation}
(\mathcal{G}^{*},\pi^{*})
=
f_\text{annot}
(\mathcal{P}_{\text{S1}},C^{*},\mathcal{E},\mathcal{M}).
\end{equation}
Here, $\mathcal{G}^{*}=\{G_j^{*}\}_{j=1}^{M}$ denotes the $M$ annotated functional groups, with each group defined as $G_j^{*}=(\mathcal{E}_j^{*},r_j^{*})$, where $\mathcal{E}_j^{*}\subseteq\mathcal{E}$ is the set of elements assigned to the $j$-th group and $r_j^{*}$ is its functional role.
The execution order is represented by $\pi^{*}=(\pi_1^{*},\ldots,\pi_M^{*})$.
The annotator also provides auxiliary explanations for grouping and execution order, which are used as context in Stage~2.
% To enrich the training supervision, we additionally generate rationales for the grouping and execution order decisions of each group, which are further used in context in Stage~2.
% Let $\mathcal{E}=\{e_i\}_{i=1}^{N}$ denote the complete element inventory. We represent each functional group as
% % $b_i^{*}$ the ground-truth bounding box of element $e_i$, $C^{*}$ the final ground-truth render, and $\mathcal{M}$ the document metadata including the canvas size, title, and intention. 
% % We represent each functional group as
% \begin{equation}
% G_j^{*}=\left(\mathcal{E}_j^{*},r_j^{*}\right),
% \end{equation}
% where $\mathcal{E}_j^{*}\subseteq\mathcal{E}$ denotes the set of elements assigned to the $j$-th group and $r_j^{*}$ denotes its functional role. The groups form a partition of the complete element inventory.

%$b_i^{*}$ denote the ground-truth bounding box of element $e_i$, 
% We provide MLLM with $(C^{*},\mathcal{E},\mathcal{M})$, together with the definitions of the functional roles and the grouping and ordering instructions. Where $C^{*}$ the final ground-truth render and $\mathcal{M}$ the document metadata including the canvas size, title, and intention. The model outputs a set of functional groups
% \begin{equation}
% \mathcal{G}^{*}=\{G_j^{*}\}_{j=1}^{M},
% \end{equation}
% and a dependency-aware execution order
% \begin{equation}
% \pi^{*}
% =
% (\pi_1^{*},\ldots,\pi_M^{*}),
% \end{equation}
% in a structured JSON format. To explicitly capture the reasoning behind the decomposition, we additionally annotate rationales for both functional grouping and execution ordering.

From the Stage~1 outputs, we further derive group regions and cumulative canvas states for downstream supervision. First, we compute a spatial region for each group, as visualized in the center of Fig.~\ref{fig:data_construction}. Let $b_i^{*}$ denote the ground-truth bounding box of element $e_i$. For each group $G_j^{*}$, we define its region as $B_j^{*}=\operatorname{BBox}\left(\{b_i^{*}\mid e_i\in\mathcal{E}_j^{*}\}\right)$, where $\operatorname{BBox}(\cdot)$ denotes the smallest bounding box covering all elements in the group.
Second, we construct cumulative canvas states following the annotated execution order, which defines the sequence of planning turns used in Stage~2. To this end, we reorder the functional groups by the annotated execution order $\pi^{*}$. We denote the resulting execution-ordered sequence as
$\widetilde{\mathcal{G}}^{*}=(\widetilde{G}_1^{*},\ldots,\widetilde{G}_M^{*})$, where $\widetilde{G}_t^{*}=G_{\pi_t^{*}}^{*}$.
We then progressively render this ordered sequence to obtain the intermediate canvas states,
$ C_t^{*} = \mathcal{R} \left( \widetilde{G}_1^{*},\ldots,\widetilde{G}_t^{*} \right)$,
where $\mathcal{R}$ denotes the document renderer~\citep{inoue2024opencolereproducibleautomaticgraphic}. Each state uses the ground-truth attributes of the included elements while preserving their original stacking order. Together with the initial empty canvas $C_0^{*}$, this yields the cumulative canvas state sequence $\{C_t^{*}\}_{t=0}^{M}$ shown at the bottom left of Fig.~\ref{fig:data_construction}, which is used in Stage~2 to construct turn-level planning annotations.

\paragraph{Stage 2: Turn-level Planning Annotation.}
\label{sec:Data_construction_stage2}
% In the second stage, we sequentially process the functional groups following the execution order from Stage~1 to obtain turn-level plans. Each group $\widetilde{G}_t^{*}$ corresponds to one planning turn, as illustrated in Fig.~\ref{fig:data_construction}.
Using the execution-ordered functional groups and cumulative canvas states constructed above, Stage~2 derives a turn-level plan for each functional group, where each group $\widetilde{G}_t^{*}$ corresponds to one planning turn, as illustrated in Fig.~\ref{fig:data_construction} (right).
For turn $t$, let $\widetilde{\mathcal{G}}_{\leq t}^{*}=(\widetilde{G}_1^{*},\ldots,\widetilde{G}_t^{*})$
denote the execution-ordered groups up to the current turn. 
The annotator is conditioned on $\widetilde{\mathcal{G}}_{\leq t}^{*}$, document metadata $\mathcal{M}$, the rendered states $C_{t-1}^{*}$ and $C_t^{*}$ before and after adding the current group, and the final design $C^{*}$. The state pair provides visual evidence for the group's placement and spatial relationships, while $C^{*}$ provides global composition context. With the Stage~2 prompt $\mathcal{P}_{\text{S2}}$, we write:
\begin{equation}
p_t^{*}
=
f_{\text{annot}}
\left(
\mathcal{P}_{\text{S2}},
\widetilde{\mathcal{G}}_{\leq t}^{*},
C_{t-1}^{*},
C_t^{*},
C^{*},
\mathcal{M}
\right),
\end{equation}
% The annotator receives $\widetilde{\mathcal{G}}_{\leq t}^{*}$, the document metadata $\mathcal{M}$, the states $C_{t-1}^{*}$ and $C_t^{*}$ rendered immediately before and after adding the current group, and the final design $C^{*}$. The pair $C_{t-1}^{*}$ and $C_t^{*}$ provide visual evidence for the placement and spatial relationships of the current group, while $C^{*}$ provides context for the overall composition and design intent.
% Denoting the Stage~2 prompt by $\mathcal{P}_\text{S2}$, we formulate the process as:
% \begin{equation}
% p_t^{*}
% =
% f_{\text{annot}}
% \left(
% \mathcal{P}_{\text{S2}},
% \widetilde{\mathcal{G}}_{\leq t}^{*},
% C_{t-1}^{*},
% C_t^{*},
% C^{*},
% \mathcal{M}
% \right),
% \end{equation}
where $p_t^{*}=(q_{t,\text{group}}^{*},q_{t,\text{exec}}^{*},q_{t,\text{layout}}^{*},\ell_t^{*})$ denotes the resulting turn-level plan.
% Here, $q_{t,\text{group}}^{*}$, $q_{t,\text{exec}}^{*}$, and $q_{t,\text{layout}}^{*}$ provide the grouping, execution, and layout rationales, respectively. 
Here, $q_{t,\text{group}}^{*}$, $q_{t,\text{exec}}^{*}$, and $q_{t,\text{layout}}^{*}$ provide rationales for the selected grouping, its execution at the current turn, and the resulting arrangement on the current canvas, respectively.
The layout instruction $\ell_t^{*}$ qualitatively describes how the current group should be placed on the canvas, including its target region and relevant spatial relations such as scale, alignment, spacing, overlap, containment, and stacking.

Overall, the two-stage pipeline produces group regions, cumulative canvas states, and group-wise turn-level plans, which are used to train the Structure-Aware Retriever, Progressive Planner, and Layout Generator.
More details of the data construction process are provided in Appendix~\ref{sec:data_construction_details}.

\section{Method}
\label{sec:method}

\begin{figure}[t]
    \centering
    \includegraphics[width=1.0\linewidth]{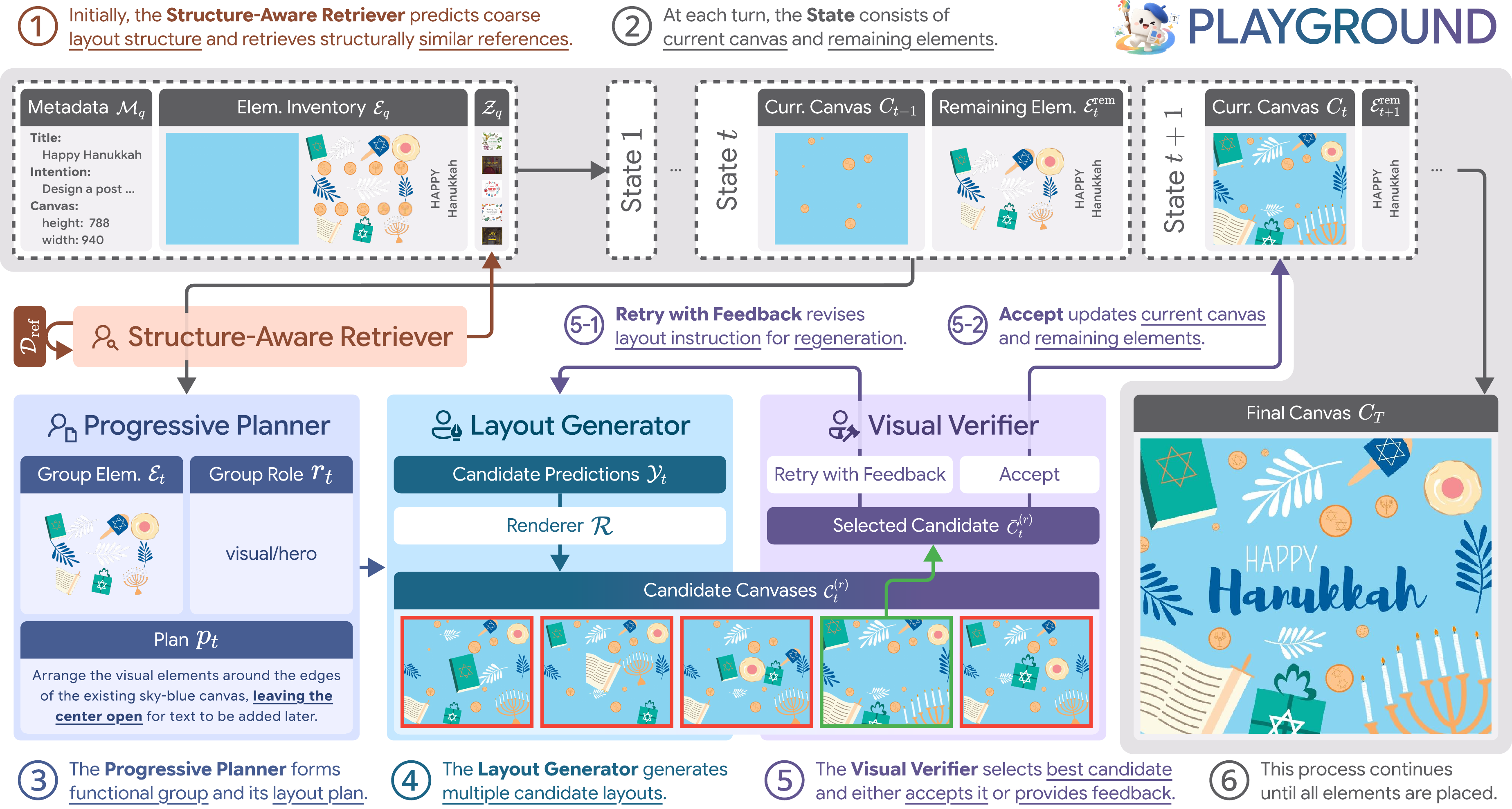}
    \vspace{-15pt}
    \caption{\textbf{Overview of \textbf{\ours}.}}
    \label{fig:framework}
\end{figure}

\subsection{Problem Formulation and Overview}
\label{sec:method_overview}
% Given a set of image and text elements, our goal is to arrange them into an editable graphic design. Following LaDeCo~\citep{lin2024elementsdesignlayeredapproach}, we predict the bounding box and text specific attributes of each element, while additionally predicting the stacking order within each functional group.

% Figure~\ref{fig:framework} illustrates the overall workflow of \ours. Before generation, the \textbf{Structure-Aware Retriever} (Sec.~\ref{method_retriever}) predicts a coarse layout structure from the input elements and retrieves five structurally similar reference designs as global composition priors. Then, at each turn, the \textbf{Progressive Planner} (Sec.~\ref{method_planner}) observes the current canvas, i.e., the canvas accepted from the preceding turn, together with the remaining elements, and forms a functional group and its layout plan. Conditioned on this plan, the \textbf{Layout Generator} (Sec.~\ref{method_predictor}) predicts the layout attributes of the selected group and generates multiple candidate layouts. The \textbf{Visual Verifier} (Sec.~\ref{method_verifier}) evaluates their rendered results, selects the best candidate, and either accepts it or provides feedback to the Layout Generator for regeneration. Once accepted, the selected canvas becomes the current canvas for the next turn, and the placed elements are removed from the remaining element set. This process continues until all elements are placed.

Given a set of image and text elements, our goal is to arrange them into an editable graphic design. Following LaDeCo~\citep{lin2024elementsdesignlayeredapproach}, we predict the bounding box and text-specific attributes of each element, while additionally predicting the stacking order within each functional group.

Figure~\ref{fig:framework} illustrates the workflow of \ours. 
Initially, the \textbf{Structure-Aware Retriever} (Sec.~\ref{method_retriever}) predicts a coarse layout structure and retrieves five structurally similar references as global composition priors. 
Then, at each turn, the \textbf{Progressive Planner} (Sec.~\ref{method_planner}) observes the current canvas, \textit{i.e.,} the canvas accepted from the preceding turn, together with the remaining elements, and forms a functional group and its layout plan. Conditioned on this plan, the \textbf{Layout Generator} (Sec.~\ref{method_predictor}) predicts the layout attributes of the elements in the group and generates multiple candidate layouts. 
Lastly, the \textbf{Visual Verifier} (Sec.~\ref{method_verifier}) selects the best candidate by evaluating their rendered results and either accepts it or provides feedback to the Layout Generator for regeneration. 
Once accepted, the selected canvas becomes the current canvas for the next turn, and the placed elements are removed from the remaining element set. This continues until all elements are placed.

\subsection{Structure-Aware Retriever}
\label{method_retriever}
The Structure-Aware Retriever retrieves reference designs that share a similar global layout structure with the input elements. 
Because no target layout is available at inference time, the Retriever must first infer a candidate layout structure to serve as a proxy for computing structural similarity. 
However, predicting precise bounding boxes for all elements at this stage would negate the benefits of progressive generation by effectively predicting the entire layout upfront.
To avoid this, we represent the coarse layout structure using functional group role--region pairs; we train the Retriever to predict the group roles and regions constructed in Stage~1 of Sec.~\ref{sec:Data_construction_stage1}.
Given the input design query $q$, retrieval prompt $\mathcal{P}_\text{ret}$, document metadata $\mathcal{M}_q$, and element inventory $\mathcal{E}_q$, the Retriever predicts the attributes of functional groups as:
% The \textbf{Structure-Aware Retriever} retrieves reference designs with a similar global layout structure from the input elements.
% Because no target layout is available at inference time, the Retriever first infers a coarse layout structure and uses it as a proxy to compute structural similarity.
% We represent the coarse layout structure using functional group roles and their corresponding regions. 
% Predicting precise bounding boxes for every element from unordered inputs would require recovering the full layout before generation, whereas group-level roles and regions provide a simpler and more intuitive structural target. We therefore train the Retriever to predict the role--region pairs constructed in Stage~1 of Sec.~\ref{sec:Data_construction_stage1}. Together, these indicate which functional components are expected and where they are likely to appear.
% Let $q$ denote the input design query. Given the retrieval prompt $\mathcal{P}_\text{ret}$, document metadata $\mathcal{M}_q$, and element inventory $\mathcal{E}_q$, the Retriever predicts:
\begin{equation}
\Phi_q
=
f_{\theta_\text{ret}}
\left(
\mathcal{P}_\text{ret},
\mathcal{M}_q,
\mathcal{E}_q
\right)
=
\left\{
\left(
r_{q,j},
B_{q,j}
\right)
\right\}_{j=1}^{J_q},
\label{eq:retriever_output}
\end{equation}
where $J_q$ is the predicted number of functional groups, and $r_{q,j}$ and $B_{q,j}$ are the functional role and spatial region of the $j$-th group, respectively.

Then, given the reference pool $\mathcal{D}_\text{ref}$ where each reference candidate is also decomposed into functional groups, we compare the predicted structure $\Phi_q$ with each candidate $c\in\mathcal{D}_\text{ref}$ as: 
% Let $\mathcal{D}_\text{ref}$ denote the reference pool, with each design represented by its Stage 1 role--region structure.
% We compare the predicted structure $\Phi_q$ with each candidate $c\in\mathcal{D}_\text{ref}$ using
\begin{equation}
\operatorname{sim}(q,c)
=
\lambda_\text{role-IoU}S_\text{role-IoU}(q,c)
+
\lambda_\text{canvas}S_\text{canvas}(q,c)
+
\lambda_\text{word/image}S_\text{word/image}(q,c).
\label{eq:retrieval_score}
\end{equation}
Here, $S_\text{role-IoU}$ measures spatial similarity between role-matched group regions,
$S_\text{canvas}$ measures similarity in canvas aspect ratio,
and $S_\text{word/image}$ compares the amount of textual and visual content.
Together, these terms capture similarity in layout structure, canvas geometry, and text--image composition. We rank the reference designs by $\operatorname{sim}(q,c)$ and retrieve the top five as $\mathcal{Z}_q$.

\subsection{Progressive Planner}
\label{method_planner}
Given the retrieved references, the \textbf{Progressive Planner} forms a functional group and plans its composition at each turn based on the remaining elements and current canvas. 
At turn $t$, let $\mathcal{E}^\text{rem}_t$ denote the remaining elements and $C_{t-1}$ the current rendered canvas, \textit{i.e.,} the canvas accepted at turn $t-1$.
Given the planning prompt $\mathcal{P}_\text{plan}$, document metadata $\mathcal{M}$, retrieved references $\mathcal{Z}$, $C_{t-1}$, and $\mathcal{E}^\text{rem}_t$, the Planner jointly predicts the current functional group $G_t$ and its turn-level plan $p_t$:
\begin{equation}
(G_t,p_t)
=
f_{\theta_\text{plan}}
\left(
\mathcal{P}_\text{plan},
\mathcal{M},
\mathcal{Z},
C_{t-1},
\mathcal{E}^\text{rem}_t
\right),
\qquad
G_t=(\mathcal{E}_t,r_t),
\label{eq:planner_output}
\end{equation}
where $\mathcal{E}_t\subseteq\mathcal{E}^\text{rem}_t$ denotes the elements selected for the current group and $r_t$ its functional role.
The plan $p_t$ follows the turn-level representation defined in Sec.~\ref{sec:Data_construction_stage2}. The Planner is supervised with the corresponding oracle group $\widetilde{G}_t^{*}$ and turn-level plan $p_t^{*}$ constructed in Sec.~\ref{sec:data_construction}, using $C_{t-1}^{*}$ as visual context.
By conditioning both group formation and layout planning on the current rendered canvas, the Planner progressively updates composition decisions as the design evolves.

% Given the reference designs, the \textbf{Planner} determines the group of elements and how they should be arranged at each turn by considering the remaining elements and the current canvas. 
% % which remaining elements should be handled as a group together at each turn and how they should be arranged. 
% At turn $t$, let $\mathcal{E}^{\mathrm{\text{rem}}}_t$ denote the remaining elements and $C_{t-1}$ the rendered canvas from the previous turn. Given the planning prompt $\mathcal{P}_{\mathrm{\text{plan}}}$, document metadata $\mathcal{M}$, retrieved references $\mathcal{Z}$, $C_{t-1}$, and $\mathcal{E}^{\mathrm{\text{rem}}}_t$, the Planner jointly predicts the current functional group $G_t$ and its turn-level plan $p_t$:
% \begin{equation}
% (G_t,p_t)
% =
% f_{\theta_\text{plan}}
% \left(
% \mathcal{P}_{\text{plan}},
% \mathcal{M},
% \mathcal{Z},
% C_{t-1},
% \mathcal{E}^{\text{rem}}_t
% \right),
% \qquad
% G_t=(\mathcal{E}_t,r_t),
% \label{eq:planner_output}
% \end{equation}
% where $\mathcal{E}_t\subseteq\mathcal{E}^{\text{rem}}_t$ denotes the
% elements assigned to the current group and $r_t$, its functional role.
% The plan $p_t$ follows the turn-level representation defined in
% Section~\ref{sec:Data_construction_stage2}. By conditioning both group formation and layout planning on the rendered state from the previous turn, the Planner progressively updates composition decisions as the design evolves.
\subsection{Layout Generator}
\label{method_predictor}
The \textbf{Layout Generator} translates the Progressive Planner's group-level plan into element-level layout attributes. 
It takes the prediction prompt $\mathcal{P}_\text{pred}$, document metadata $\mathcal{M}$, retrieved references $\mathcal{Z}$, current rendered canvas $C_{t-1}$, functional group $G_t$, and plan $p_t$, and predicts:
\begin{equation}
\mathcal{Y}_t
=
f_{\theta_\text{pred}}
\left(
\mathcal{P}_\text{pred},
\mathcal{M},
\mathcal{Z},
C_{t-1},
G_t,
p_t
\right)
=
\left\{
y_i=(b_i,z_i,\tau_i)
\mid
e_i \in \mathcal{E}_t
\right\}.
\label{eq:predictor_output}
\end{equation}
Here, $b_i$ denotes the bounding box, $z_i$ the group-local stacking order, and $\tau_i$ the text-specific attributes when applicable.
Only the attributes of the current group are predicted, while elements placed in previous turns remain unchanged.
The Layout Generator is trained with the ground-truth element-level layout annotations of $\widetilde{G}_t^{*}$, conditioned on the oracle plan $p_t^{*}$ and canvas $C_{t-1}^{*}$.

% Since a group-level plan does not uniquely determine the exact element-level attributes, different predictions under the same plan can lead to substantially different rendered outcomes. We therefore generate $K$ candidate layouts for each group rather than relying on a single prediction.
Since a group-level plan leaves element-level attributes underconstrained, a single layout prediction may yield suboptimal arrangements or spatial conflicts. 
We therefore generate $K$ candidate layouts for each group and select the optimal one, rather than relying on a single prediction.

\subsection{Visual Verifier}
\label{method_verifier}
Finally, the \textbf{Visual Verifier} evaluates the $K$ rendered candidates and determines the canvas for the next planning turn.
% At verification round $r$, let $\mathcal{C}^{(r)}_t=\{C_t^{(r,k)}\}^{K}_{k=1}$ denote the set of candidate canvases produced by the Layout Generator using the current layout instruction $\ell^{(r)}_t$, where $\ell^{(1)}_t=\ell_t$.
% For each round, the verifier first selects the candidate that best follows $\ell_t^{(r)}$ and achieves high visual quality, then decides whether to accept it or retry with a revised layout instruction.
% \begin{equation}
% \left(
% \bar{C}_t^{(r)},
% d_t^{(r)}
% \right)
% =
% f_{\mathrm{\text{ver}}}
% \left(
% \mathcal{P}_{\mathrm{\text{ver}}},
% G_t,
% p_t,
% C_{t-1},
% \mathcal{C}_t^{(r)}
% \right),
% \end{equation}
% where $\bar{C}_t^{(r)}\in\mathcal{C}_t^{(r)}$ denotes the candidate selected at round \(r\), and \(d_t^{(r)}\in\{\texttt{accept},\texttt{retry\_with\_feedback}\}\) denotes the verification decision.
% If accepted, the selected candidate becomes $C_t$ and is passed to the Planner for the next turn. Otherwise, the verifier revises the layout instruction, $\ell_t^{(r)}\rightarrow\ell_t^{(r+1)}$, and the Layout Generator generates a new candidate set for the same group. We allow up to three verification rounds; if no round is accepted, the verifier makes a final choice among the best candidate from each round.
At verification round $r$, let
$\mathcal{C}^{(r)}_t=\{C_t^{(r,k)}\}_{k=1}^{K}$
denote the candidate canvases generated using layout instruction $\ell_t^{(r)}$, with $\ell_t^{(1)}=\ell_t$.
Given the verification prompt $\mathcal{P}_\text{ver}$, current rendered canvas $C_{t-1}$, group $G_t$, current layout instruction $\ell_t^{(r)}$, and candidate set $\mathcal{C}_t^{(r)}$, the Verifier first selects the candidate that best follows the current layout instruction while maintaining high visual quality, then decides whether to accept it or retry with feedback:
\begin{equation}
\left(
\bar{C}_t^{(r)},
d_t^{(r)}
\right)
=
f_{\mathrm{ver}}
\left(
\mathcal{P}_\text{ver},
G_t,
\ell_t^{(r)},
C_{t-1},
\mathcal{C}_t^{(r)}
\right),
\label{eq:verifier}
\end{equation}
where $\bar{C}_t^{(r)}\in\mathcal{C}_t^{(r)}$ is the selected candidate and
$d_t^{(r)}\in\{\texttt{accept}, \texttt{retry\_with\_feedback}\}$ is the verification decision.
If accepted, $\bar{C}_t^{(r)}$ becomes $C_t$ and is passed to the Planner for the next turn.
Otherwise, the Verifier returns a revised layout instruction,
$\ell_t^{(r+1)}$,
and the Layout Generator generates $K$ new candidates for the same group.
We allow up to three verification rounds; if none is accepted, the Verifier makes a final selection among the best candidates from all rounds.

\subsection{Training objective}
The Structure-Aware Retriever, Progressive Planner, and Layout Generator are separately fine-tuned with autoregressive supervision on their respective
targets. For module $m\in\{\text{ret},\text{plan},\text{pred}\}$,
\begin{equation}
\mathcal{L}_{m}
=
-\log
p_{\theta_m}
\left(
o_m^{*}\mid x_m^{*}
\right),
\end{equation}
where $x_m^{*}$ and $o_m^{*}$ denote the module-specific input context and
oracle target, respectively.
The Visual Verifier is implemented with GPT-5.6 Luna~\citep{openai2026gpt56}
and is used without additional training.
Further algorithmic and implementation details are provided in Appendix~\ref{sec:main_pipeline_details}.
\section{Experiments}
\label{sec:experiments}

\begin{table}[t]
    \centering
    \setlength{\tabcolsep}{1pt}
    \caption{\small\textbf{Quantitative Results on the Crello and LICA datasets.}
    We report six VLM-based metrics (mean $\pm$ standard deviation) and two rule-based metrics, T.Occ. and Rea. \textbf{Bold} and \underline{underline} mark the best and the second best
    within each dataset, excluding GT.}    
    \label{tab:main_quan}
    \vspace{-9pt}
    \scriptsize
    \renewcommand{\arraystretch}{0.95}

    \begin{tabular}{
        >{\centering\arraybackslash}m{0.07\textwidth}|
        >{\raggedright\arraybackslash\hspace{3pt}}m{0.115\textwidth}|
        *{6}{>{\centering\arraybackslash\tiny}m{0.097\textwidth}}|
        *{2}{>{\centering\arraybackslash\tiny}m{0.083\textwidth}}}
        \toprule

        \multicolumn{1}{c|}{\multirow[c]{2}{*}{\textbf{Dataset}}}
        & \multicolumn{1}{c|}{\multirow[c]{2}{*}{\textbf{Method}}}
        & \multicolumn{6}{c|}{\textbf{VLM-based metrics $\uparrow$}}
        & \multicolumn{2}{c}{\textbf{Rule-based metrics $\downarrow$}} \\

        &
        & {\scriptsize\mbox{(i)}}
        & {\scriptsize\mbox{(ii)}}
        & {\scriptsize\mbox{(iii)}}
        & {\scriptsize\mbox{(iv)}}
        & {\scriptsize\mbox{(v)}}
        & {\scriptsize\mbox{(vi)}}
        & {\scriptsize\mbox{T.Occ.}}
        & {\scriptsize\mbox{Rea.}} \\

        \datasetseparator
        % \multirow[c]{6}{*}{\makecell{\textbf{Crello}\\\textbf{dataset}}}
        
        % & \multicolumn{9}{l}{\textcolor{gray}{\textit{Intention-to-Design Generation Models}}} \\
        % & OpenCOLE
        % & \shortstack{3.666 \textcolor{gray}{$\pm$0.710}}
        % & \shortstack{2.028 \textcolor{gray}{$\pm$0.948}}
        % & \shortstack{4.015 \textcolor{gray}{$\pm$0.551}}
        % & \shortstack{4.426 \textcolor{gray}{$\pm$0.568}}
        % & \shortstack{4.246 \textcolor{gray}{$\pm$0.655}}
        % & \shortstack{3.331 \textcolor{gray}{$\pm$1.153}}
        % & 0.0034 & 0.0737 \\

        % \cmidrule(l){2-10}
        % & \multicolumn{9}{l}{\textcolor{gray}{\textit{Graphic Design Composition Models}}} \\
        & \metricrowstrut FlexDM
        & \shortstack{2.670 \textcolor{gray}{$\pm$0.690}}
        & \shortstack{2.156 \textcolor{gray}{$\pm$0.656}}
        & \shortstack{3.133 \textcolor{gray}{$\pm$0.705}}
        & \shortstack{3.637 \textcolor{gray}{$\pm$0.581}}
        & \shortstack{3.200 \textcolor{gray}{$\pm$0.740}}
        & \shortstack{3.021 \textcolor{gray}{$\pm$0.992}}
        & 0.1519 & 0.0583 \\

        & \metricrowstrut LaDeCo
        & \shortstack{3.103 \textcolor{gray}{$\pm$0.722}}
        & \shortstack{2.615 \textcolor{gray}{$\pm$0.809}}
        & \shortstack{3.451 \textcolor{gray}{$\pm$0.721}}
        & \shortstack{\underline{3.979} \textcolor{gray}{$\pm$0.574}}
        & \shortstack{3.592 \textcolor{gray}{$\pm$0.743}}
        & \shortstack{3.528 \textcolor{gray}{$\pm$1.029}}
        & 0.0570 & 0.0432 \\

        & \metricrowstrut PosterCopilot
        & \shortstack{2.980 \textcolor{gray}{$\pm$0.729}}
        & \shortstack{2.623 \textcolor{gray}{$\pm$0.810}}
        & \shortstack{3.263 \textcolor{gray}{$\pm$0.776}}
        & \shortstack{3.809 \textcolor{gray}{$\pm$0.653}}
        & \shortstack{3.467 \textcolor{gray}{$\pm$0.784}}
        & \shortstack{3.435 \textcolor{gray}{$\pm$1.073}}
        & 0.0319 & 0.0374 \\

        & \metricrowstrut VFLM
        & \shortstack{\underline{3.335} \textcolor{gray}{$\pm$0.564}}
        & \shortstack{\underline{2.718} \textcolor{gray}{$\pm$0.645}}
        & \shortstack{\underline{3.714} \textcolor{gray}{$\pm$0.576}}
        & \shortstack{3.969 \textcolor{gray}{$\pm$0.565}}
        & \shortstack{\underline{3.759} \textcolor{gray}{$\pm$0.642}}
        & \shortstack{\underline{3.693} \textcolor{gray}{$\pm$0.947}}
        & \textbf{0.0002} & \underline{0.0360} \\

        \rowcolor{hlours}[\tabcolsep][\tabcolsep]\cellcolor{white}
        & \metricrowstrut\textbf{\ours}
        & \shortstack{\textbf{3.794} \textcolor{gray}{$\pm$0.583}}
        & \shortstack{\textbf{3.503} \textcolor{gray}{$\pm$0.660}}
        & \shortstack{\textbf{3.899} \textcolor{gray}{$\pm$0.587}}
        & \shortstack{\textbf{4.381} \textcolor{gray}{$\pm$0.542}}
        & \shortstack{\textbf{4.183} \textcolor{gray}{$\pm$0.635}}
        & \shortstack{\textbf{4.248} \textcolor{gray}{$\pm$0.919}}
        & \underline{0.0041}
        & \textbf{0.0345} \\

        \cline{2-10}
        \multirow[c]{-6}{*}[2.5pt]{\makecell{\textbf{Crello}\\\textbf{dataset}}}

        & \metricrowstrut GT
        & \shortstack{4.120 \textcolor{gray}{$\pm$0.509}}
        & \shortstack{3.830 \textcolor{gray}{$\pm$0.490}}
        & \shortstack{4.077 \textcolor{gray}{$\pm$0.635}}
        & \shortstack{4.575 \textcolor{gray}{$\pm$0.522}}
        & \shortstack{4.452 \textcolor{gray}{$\pm$0.555}}
        & \shortstack{4.414 \textcolor{gray}{$\pm$0.880}}
        & 0.0025 & 0.0235 \\

        \datasetseparator
        % \multirow[c]{6}{*}{\makecell{\textbf{LICA}\\\textbf{dataset}}}

        % & \multicolumn{9}{l}{\textcolor{gray}{\textit{Intention-to-Design Generation Models}}} \\
        % & OpenCOLE
        % & \shortstack{3.359 \textcolor{gray}{$\pm$0.697}}
        % & \shortstack{2.219 \textcolor{gray}{$\pm$0.843}}
        % & \shortstack{3.843 \textcolor{gray}{$\pm$0.585}}
        % & \shortstack{4.296 \textcolor{gray}{$\pm$0.565}}
        % & \shortstack{4.090 \textcolor{gray}{$\pm$0.678}}
        % & \shortstack{2.837 \textcolor{gray}{$\pm$1.053}}
        % & 0.0035 & 0.0537 \\

        % \cmidrule(l){2-10}

        % & \multicolumn{9}{l}{\textcolor{gray}{\textit{Graphic Design Composition Models}}} \\
        & \metricrowstrut FlexDM
        & \shortstack{2.186 \textcolor{gray}{$\pm$0.529}}
        & \shortstack{1.790 \textcolor{gray}{$\pm$0.581}}
        & \shortstack{2.711 \textcolor{gray}{$\pm$0.811}}
        & \shortstack{3.373 \textcolor{gray}{$\pm$0.625}}
        & \shortstack{2.742 \textcolor{gray}{$\pm$0.730}}
        & \shortstack{1.889 \textcolor{gray}{$\pm$0.830}}
        & 0.3212 & 0.0574 \\

        & \metricrowstrut LaDeCo
        & \shortstack{2.889 \textcolor{gray}{$\pm$0.760}}
        & \shortstack{2.444 \textcolor{gray}{$\pm$0.810}}
        & \shortstack{3.294 \textcolor{gray}{$\pm$0.798}}
        & \shortstack{3.965 \textcolor{gray}{$\pm$0.634}}
        & \shortstack{3.501 \textcolor{gray}{$\pm$0.799}}
        & \shortstack{3.171 \textcolor{gray}{$\pm$1.101}}
        & 0.0715 & 0.0393 \\

        & \metricrowstrut PosterCopilot
        & \shortstack{2.929 \textcolor{gray}{$\pm$0.787}}
        & \shortstack{2.731 \textcolor{gray}{$\pm$0.983}}
        & \shortstack{3.395 \textcolor{gray}{$\pm$0.856}}
        & \shortstack{3.961 \textcolor{gray}{$\pm$0.679}}
        & \shortstack{3.581 \textcolor{gray}{$\pm$0.827}}
        & \shortstack{3.097 \textcolor{gray}{$\pm$1.244}}
        & 0.1372 & 0.0281 \\

        & \metricrowstrut VFLM
        & \shortstack{\underline{3.427} \textcolor{gray}{$\pm$0.657}}
        & \shortstack{\underline{2.933} \textcolor{gray}{$\pm$0.718}}
        & \shortstack{\textbf{3.977} \textcolor{gray}{$\pm$0.653}}
        & \shortstack{\underline{4.157} \textcolor{gray}{$\pm$0.586}}
        & \shortstack{\underline{4.022} \textcolor{gray}{$\pm$0.662}}
        & \shortstack{\underline{3.647} \textcolor{gray}{$\pm$1.007}}
        & \textbf{0.0001} & \underline{0.0216} \\

        \rowcolor{hlours}[\tabcolsep][\tabcolsep]\cellcolor{white}
        & \metricrowstrut\textbf{\ours}
        & \shortstack{\textbf{3.698} \textcolor{gray}{$\pm$0.710}}
        & \shortstack{\textbf{3.480} \textcolor{gray}{$\pm$0.731}}
        & \shortstack{\underline{3.970} \textcolor{gray}{$\pm$0.722}}
        & \shortstack{\textbf{4.378} \textcolor{gray}{$\pm$0.545}}
        & \shortstack{\textbf{4.217} \textcolor{gray}{$\pm$0.693}}
        & \shortstack{\textbf{4.224} \textcolor{gray}{$\pm$0.906}}
        & \underline{0.0482}
        & \textbf{0.0199} \\
        
        \cline{2-10}
        \multirow[c]{-6}{*}[2.5pt]{\makecell{\textbf{LICA}\\\textbf{dataset}}}

        & \metricrowstrut GT
        & \shortstack{4.376 \textcolor{gray}{$\pm$0.541}}
        & \shortstack{4.150 \textcolor{gray}{$\pm$0.459}}
        & \shortstack{4.512 \textcolor{gray}{$\pm$0.531}}
        & \shortstack{4.699 \textcolor{gray}{$\pm$0.463}}
        & \shortstack{4.749 \textcolor{gray}{$\pm$0.454}}
        & \shortstack{4.884 \textcolor{gray}{$\pm$0.358}}
        & 0.0040 & 0.0162 \\

        \specialrule{\heavyrulewidth}{0pt}{0pt}
    \end{tabular}

    % Crello VLM coverage: OpenCOLE, FlexDM, LaDeCo, Ours, and revised GT use
    % all available renders; PosterCopilot omits its four generation failures,
    % and VFLM uses the source-text-bearing designs. LICA VLM coverage: GT,
    % OpenCOLE, FlexDM, and LaDeCo use all 1,011 static designs, PosterCopilot
    % uses its 1,010 available renders, and VFLM uses the 956 source-text-bearing
    % designs; Ours uses all 1,011 available renders. In both
    % datasets, criterion (ii) alone excludes source-text-free designs during
    % dataset-level aggregation. The rule-based cells are rounded from
    % outputs/<method>.json; their valid sample counts can differ from the VLM
    % coverage because they require parseable text/layout outputs.
    % Best and second-best means within each dataset's graphic
    % design composition group are shown in bold and underlined; OpenCOLE and
    % GT are excluded from this ranking. LICA judge images use the original PNG
    % bytes, preserving native RGB/RGBA and alpha without resizing, re-encoding,
    % or compositing. Design Intention Alignment is not an input-equivalent generation
    % comparison because methods were generated under different conditioning
    % regimes. Reported standard deviations measure variation across designs,
    % not repeated judge calls.

\end{table}

\subsection{Experimental Setup}
\paragraph{Implementation Details.}
We initialize our Retriever, Planner, and Layout Generator with Qwen3-VL-8B-Instruct~\citep{bai2025qwen3vltechnicalreport} and fine-tune them using LoRA~\citep{hu2021loralowrankadaptationlarge}, while GPT-5.6 Luna~\citep{openai2026gpt56} is used as the Visual Verifier.
All experiments are conducted on four NVIDIA RTX 5880 GPUs.
Detailed training configurations are provided in Appendix~\ref{sec:training_config}.
% Qwen3-VL-8B-Instruct~\citep{bai2025qwen3vltechnicalreport} is used to initialize our Retriever, Planner, and Layout Generator, which are finetuned using LoRA~\citep{hu2021loralowrankadaptationlarge}, while Visual Verifier uses GPT-5.6 Luna~\citep{openai2026gpt56}. 
% All runs are conducted on 4 NVIDIA RTX 5880 48G GPUs.

\paragraph{Baselines and Datasets.}
% We evaluate all methods on the Crello-v4 dataset~\citep{yamaguchi2021canvasvaelearninggeneratevector} and compare against two groups of baselines: (1) \emph{intention-to-design} models, which directly translate user intentions into final designs, including OpenCOLE~\citep{inoue2024opencolereproducibleautomaticgraphic}; and (2) \emph{graphic design composition} models, which compose given visual and textual elements into final designs, including the open-source models FlexDM~\citep{inoue2023flexiblemultimodaldocumentmodels}, LaDeCo~\citep{lin2024elementsdesignlayeredapproach}, PosterCopilot~\citep{wei2025postercopilotlayoutreasoningcontrollable}, and VFLM~\citep{guo2026seeingimprovingvisualfeedback}.

%%% lica 사용시
We evaluate on the Crello-v4~\citep{yamaguchi2021canvasvaelearninggeneratevector} and LICA~\citep{hirsch2026licalayeredimagecomposition} datasets, training all learnable components on Crello and evaluating LICA in a zero-shot setting. We compare \ours against open-source graphic design composition baselines, including FlexDM~\citep{inoue2023flexiblemultimodaldocumentmodels}, LaDeCo~\citep{lin2024elementsdesignlayeredapproach}, PosterCopilot~\citep{wei2025postercopilotlayoutreasoningcontrollable}, and VFLM~\citep{guo2026seeingimprovingvisualfeedback}, in the setting of arranging given visual and textual elements into a final design. Detailed settings of baselines are provided in Appendix~\ref{sec:evaluation_setup}.

\paragraph{Evaluation Metrics.}
Following prior works~\citep{jia2024colehierarchicalgenerationframework,lin2024elementsdesignlayeredapproach,hong2026creatipostereditablecontrollablemultilayer,wei2025postercopilotlayoutreasoningcontrollable,liu2026designascode}, we evaluate both perceptual design quality and structural validity. We use GPT-5~\citep{openai2025gpt5} as a VLM judge to independently score six criteria from 1 to 5: (i) layout composition rationality, (ii) typography legibility and hierarchy, (iii) visible element presentation, (iv) color harmony and contrast, (v) visual style coherence, and (vi) design intention alignment. We additionally report two rule-based metrics, text occlusion (T.Occ.) and text unreadability (Rea.), which measure text covered by foreground elements and background complexity beneath text, respectively. Further evaluation details and a complementary user study are provided in Appendix~\ref{sec:metrics}.

% A comprehensive evaluation of graphic design should consider both the holistic perceptual quality of the completed render and measurable structural and geometric properties of the generated design. Drawing on prior multi-aspect evaluation protocols for graphic design ~\citep{jia2024colehierarchicalgenerationframework, lin2024elementsdesignlayeredapproach, hong2026creatipostereditablecontrollablemultilayer, wei2025postercopilotlayoutreasoningcontrollable, liu2026designascode}, we define six criteria for our evaluation. We use GPT-5~\citep{openai2025gpt5} as a VLM judge that independently scores each
% criterion from 1 to 5: (i) layout composition rationality, (ii) typography legibility and hierarchy, (iii) visible element presentation, (iv) color harmony and contrast, (v) visual style coherence, and (vi) design intention alignment. Together, these criteria assess the design's visual organization, communicative clarity, overall coherence, and consistency with its intended purpose.

% To complement this perceptual assessment, we also report two rule-based metrics: text occlusion (T.Occ.), which measures text covered by foreground elements, and text unreadability (Rea.), which measures background complexity beneath text. Further details on the evaluation protocol are provided in the Appendix. % supplementary material

\begin{figure}[t]
    \centering
    \includegraphics[width=1.0\linewidth]{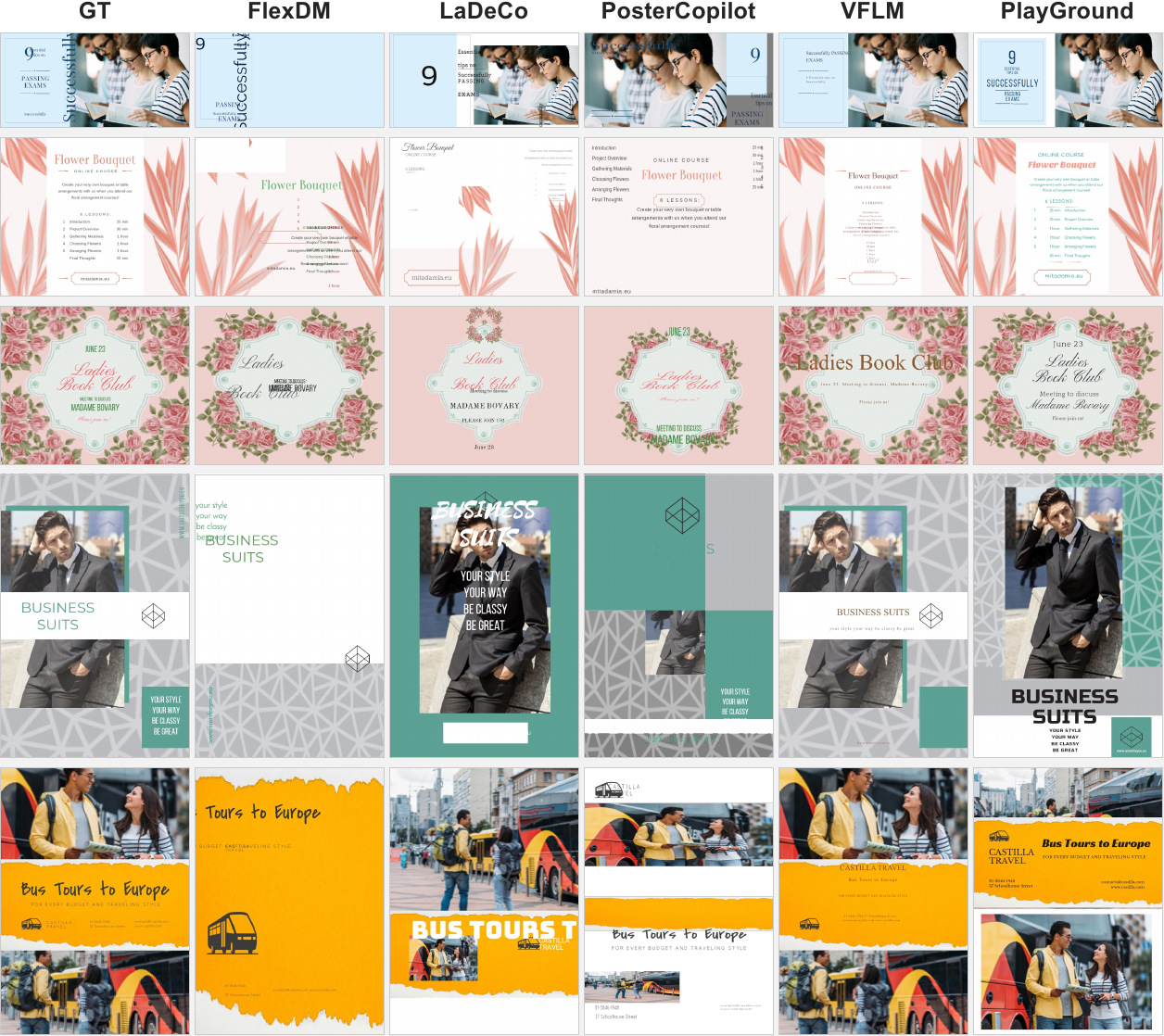}
    \vspace{-15pt}
    \caption{
    \textbf{Qualitative results on Crello dataset.}
    }
    \label{fig:main_qual}
    % \vspace{-10pt}
\end{figure}
\begin{figure}[t]
    \centering
    \includegraphics[width=1.0\linewidth]{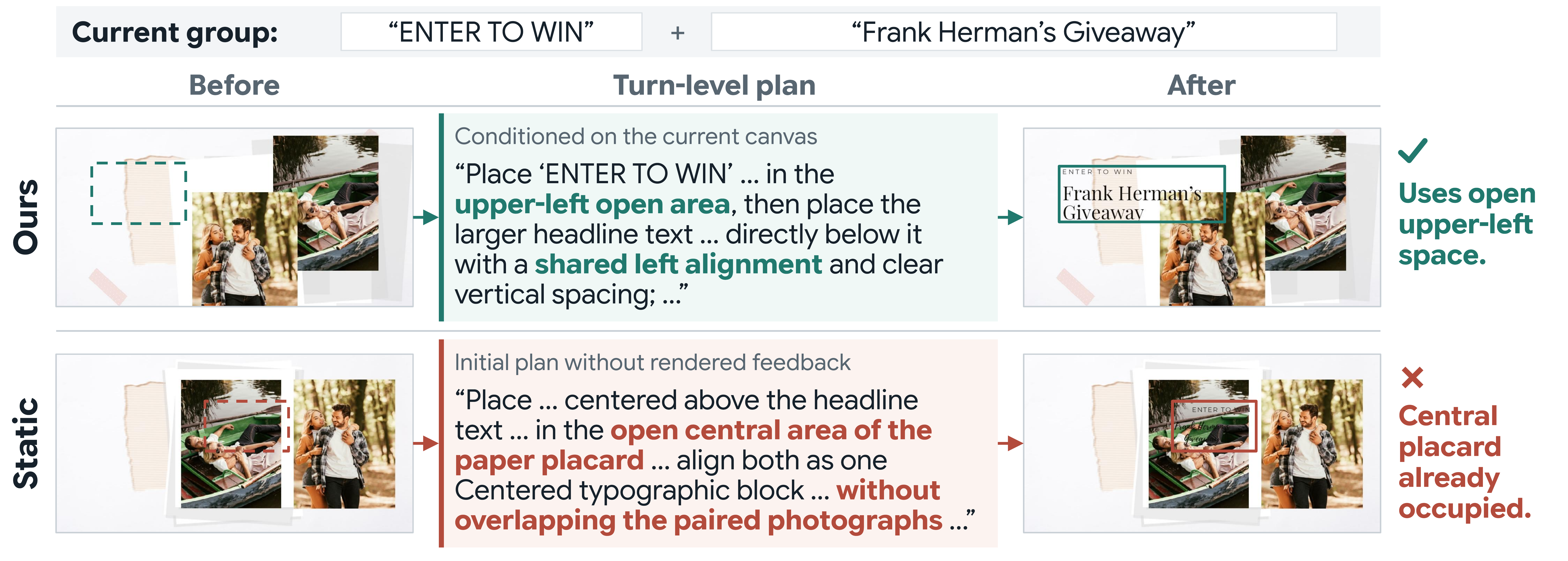}
    \vspace{-15pt}
    \caption{\textbf{Effect of our Progressive Planner.}
    Our Progressive Planner adapts to the rendered canvas and uses available space, while the Static Planner follows an outdated initial plan.}
    \label{fig:planner_qual}
    \vspace{-10pt}
\end{figure}
\subsection{Comparison with State-of-the-Art Methods}
Table~\ref{tab:main_quan} shows that \ours leads the baselines on all six VLM metrics on Crello and five of six on LICA, indicating strong overall graphic design quality.
Its performance on LICA without additional fine-tuning further supports strong zero-shot generalization.
The low standard deviations also indicate consistent design quality.
We attribute this consistency in part to our Progressive Planner, which forms each plan from the actual canvas state, aligning composition decisions with the evolving design.
While VFLM also exhibits lower variation on some metrics, we attribute this behavior to its constrained setting of rendering only text on a fixed ground-truth background.
For the rule-based metrics, \ours achieves lower T.Occ. and Rea. than FlexDM, LaDeCo, and PosterCopilot, indicating fewer text occlusions and less complex backgrounds beneath text.
Note that VFLM's lower T.Occ. is an expected artifact of its text-only rendering constraint.

Figure~\ref{fig:main_qual} compares the rendered designs of each method with the corresponding ground truth. \ours consistently produces coherent image--text compositions by capturing spatial and semantic relationships among input elements. It places text in appropriate free regions without obscuring salient visuals and effectively uses supporting elements to strengthen typography, as shown in the second and fourth examples. Compared with VFLM, which places text on a ground-truth visual background, \ours achieves competitive or better typography and hierarchy while jointly composing visual and textual elements. The third to fifth examples further show that \ours can produce plausible alternatives to the ground truth with strong balance and hierarchy. In contrast, several baselines exhibit fragmented compositions, out-of-canvas placement, or undesirable occlusions. These results highlight the benefit of \ours, which composes related elements as functional groups and adapts placement and stacking decisions to the evolving canvas.

% Figure~\ref{fig:main_qual} compares the rendered designs of each method with the corresponding ground-truth designs. \ours consistently produces coherent image-text compositions by understanding the spatial and semantic relationships among input elements. It places text in appropriate free regions without obscuring salient visual content and effectively uses supporting visual elements as containers or decorative structures to strengthen typography, as illustrated in the second and fourth examples. Compared with VFLM, which places text on the ground-truth visual background, \ours achieves competitive or better typography and visual hierarchy while jointly composing both visual and textual elements. Moreover, the third to fifth examples show that \ours can produce plausible alternative compositions that differ from the ground truth while maintaining strong visual balance and hierarchy. In contrast, several baselines exhibit fragmented compositions, out-of-canvas placement, or undesirable occlusions. These results highlight the benefit of our group-wise progressive pipeline, which composes related elements as coherent units and adapts subsequent placement and stacking decisions to the evolving canvas.

\subsection{Analysis}
\paragraph{Ablation Studies.}

\begin{wraptable}{r}{0.48\columnwidth}
    \vspace{-12pt}
    \centering
    \scriptsize
    \setlength{\tabcolsep}{7pt}
    \caption{\small\textbf{Ablation on pipeline components.}
    % We present an ablation study on our pipeline components. 
    VLM denotes the mean of six VLM scores on a Crello subset.
    $\triangle$ denotes a Static Planner.}
    \label{tab:abl_module}
    \vspace{-8pt}
    \resizebox{\linewidth}{!}{
        \begin{tabular}{l|cccc|c}
            \toprule
            
            \textbf{\#}
            & \textbf{Retriever}
            & \textbf{Planner}
            & \textbf{$K$}
            & \textbf{Verifier}
            & \textbf{VLM} $\uparrow$ \\
            
            \specialrule{\lightrulewidth}{\aboverulesep}{1pt}
    
            \hlrow \textbf{(I)} & Top-5 & \cmark & 5 & \cmark
            & \textbf{3.958} \\
            \textbf{(II)} & \xmark & \cmark & 5 & \cmark
            & 3.901 \\
            \textbf{(III)} & Lowest-5 & \cmark & 5 & \cmark
            & 3.867 \\
            \textbf{(IV)} & Top-5 & \cmark & 3 & \cmark
            & 3.923 \\
            \textbf{(V)} & Top-5 & \cmark & 1 & \cmark
            & 3.891 \\
            \textbf{(VI)} & Top-5 & \cmark & 5 & \xmark
            & 3.698 \\
            \textbf{(VII)} & Top-5 & $\triangle$ & 5 & \cmark
            & 3.894 \\
    
            \bottomrule
        \end{tabular}
    }
    \vspace{-12pt}
\end{wraptable}

Table~\ref{tab:abl_module} analyzes the contribution of each component. Comparing (I) and (II), retrieval improves performance, showing that reference designs provide useful guidance. More importantly, (III), which uses the lowest-ranked references under our metric, performs even worse than using no references, indicating that mismatched references can harm generation and supporting the effectiveness of our metric in retrieving structurally similar designs.

We next examine candidate generation and verification. Comparing (I), (IV), and (V), performance consistently improves as the number of candidates $K$ increases, showing the benefit of generating multiple layout candidates. Comparing (I) and (VI), removing the visual verifier and randomly selecting a candidate causes a clear drop, demonstrating the importance of evaluating rendered candidates, selecting the best one, and providing feedback for regeneration.
Finally, (VII) uses a separately trained Qwen3-VL-based~\citep{bai2025qwen3vltechnicalreport} Static Planner that generates the entire plan once at the beginning. Its lower performance than (I) demonstrates the benefit of progressive planning from intermediate renders rather than relying on a fixed initial plan. Fig.~\ref{fig:planner_qual} further illustrates this difference: our Planner uses the available upper-left space for text, whereas the static plan targets a central placard region already occupied by photographs. Further ablation settings are in Appendix~\ref{sec:ablation_settings}.

\paragraph{Effect of Per-Turn Granularity and Grouping.}

\begin{wraptable}{r}{0.48\columnwidth}
    \vspace{-13pt}
    \centering
    \scriptsize

    \caption{\small\textbf{Ablation on per-turn granularity and grouping.}
    VLM denotes the mean of six VLM scores on a Crello subset.}
    \label{tab:abl_group}
    \vspace{-8pt}

    \resizebox{\linewidth}{!}{
        \begin{tabular}{l|l|c}
            \toprule
            
            {\textbf{Grouping scheme}} & \textbf{Per-turn Unit}
            & \textbf{VLM} $\uparrow$ \\
            
            \specialrule{\lightrulewidth}{\aboverulesep}{1pt}
            
            \makebox[2em][l]{(I)} None & Element-wise & 3.464 \\
            \makebox[2em][l]{(II)} None & Full-layout  & 3.351 \\
            \makebox[2em][l]{(III)} LaDeCo-style & Semantic layer & \underline{3.532} \\
            \hlrow \makebox[2em][l]{(IV)} \textbf{Ours} & Functional Group & \textbf{3.578} \\
            
            \specialrule{\heavyrulewidth}{1pt}{0pt}
        \end{tabular}
    }
    \vspace{-15pt}
\end{wraptable}

% We study the effect of per-turn granularity and grouping using GPT-5.6 Luna in a zero-shot setting. To isolate these factors, we use only the Planner--Layout Generator loop without the Visual Verifier, and vary the set of elements handled at each turn.

% We first compare two variants without grouping. Processing one element per turn achieves higher design quality than placing all elements at once, but requires substantially longer inference due to the increased number of iterations. In contrast, both group-based variants achieve higher VLM scores than these non-grouping baselines, suggesting that jointly planning and placing functionally related elements is an effective unit for iterative composition.

% We further compare our functional grouping with the LaDeCo-style grouping~\citep{}, which follows its five predefined semantic layers and placement order. The LaDeCo-style variant is faster due to its predetermined layer structure, but achieves a lower VLM score than \ours. Our approach instead forms functional groups from six possible roles without fixing the number of groups. This achieves the highest design quality while remaining substantially faster than element-wise generation, providing a favorable balance between quality and inference efficiency.

We study per-turn granularity and grouping using GPT-5.6 Luna in a zero-shot setting with only the Planner--Layout Generator loop to isolate these factors. We compare (I) element-wise generation, (II) full-layout generation that places all elements at once, (III) LaDeCo-style semantic-layer grouping, and (IV) our functional grouping.

Without grouping, (I) element-wise generation outperforms (II) full-layout generation, suggesting that decomposing layout generation into sequential decisions is beneficial. However, both group-based variants (III, IV) further improve performance, showing that jointly planning and placing functionally related elements provides a more effective unit for iterative composition.
Compared with (III) LaDeCo-style grouping~\citep{lin2024elementsdesignlayeredapproach}, which uses five predefined semantic layers and a fixed placement order, \ours forms groups from six functional roles without fixing the number of groups. The higher VLM score of (IV) supports the benefit of this more flexible grouping strategy for our iterative generation pipeline.

\paragraph{Extension to Intention-to-Design.}
To address settings where design elements are not provided in advance, we further extend \ours to the intention-to-design task, where a design intention is directly converted into a final design. We prepend an element-generation stage using GPT-5.6 Luna \citep{openai2026gpt56} and Qwen-Image-2.1~\citep{qwenimage21} before applying \ours. Qualitative results and experimental details are provided in Appendix~\ref{sec:intention_to_design}.
\section{Conclusion}
\label{sec:conclusion}

We propose \textbf{\ours}, a progressive layout generation framework that adapts composition decisions to the evolving rendered canvas. At each turn, the Progressive Planner forms a functional group from the remaining elements and plans its composition based on the current canvas. A Structure-Aware Retriever provides global composition guidance by predicting a coarse layout structure and retrieving structurally similar references. To enable such progressive planning, we further introduce a two-stage data construction pipeline that constructs functional groups and derives supervision for group-wise turn-level planning from completed designs. Experiments on Crello and LICA demonstrate improved design quality and zero-shot generalization over existing composition methods, while ablations validate the effectiveness of our key components.

\subsection*{{AI use statement}}
We used generative AI tools in several parts of this work.
GPT-5.6 Luna was used to construct annotations for our two-stage data
construction pipeline and as the Visual Verifier during inference.
GPT-5 was used for VLM-based evaluation.
Generative AI tools were also used for auxiliary data processing and experiments,
including document title generation for LICA and the intention-to-design extension.
We additionally used generative AI tools for language polishing of the manuscript.
We reviewed all AI-assisted outputs and take full responsibility for the final text, claims, experimental results, and artifacts presented in this work.

\bibliography{iclr2027_conference}
\bibliographystyle{iclr2027_conference}

\newpage
\appendix

\section*{\Large Appendix}
\raggedbottom

\section{Additional Related Work}
\label{sec:additional_related_work}

This section provides additional discussion of MLLMs and broader graphic design layout generation beyond the iterative graphic design systems discussed in Sec.~\ref{sec:related_work}.

\paragraph{Multimodal Large Language Models.}
MLLMs have rapidly advanced in visual perception, language understanding, and reasoning, enabling their application to a wide range of multimodal tasks~\citep{bai2025qwen3vltechnicalreport, openai2024gpt4o,openai2025gpt5, openai2026gpt56}. 
Recently, their use has expanded beyond standalone task models to multi-stage systems in which multiple MLLMs collaborate within a unified pipeline~\citep{yao2025surveyagenticmultimodallarge}. 
In graphic design and layout generation, MLLMs are also increasingly used for data generation~\citep{guo2026seeingimprovingvisualfeedback} and evaluating generated outputs~\citep{deganutti2026graphicdesignbenchcomprehensivebenchmarkevaluating}.

\paragraph{Graphic Design Layout Generation.}
Beyond the iterative graphic design systems discussed in Sec.~\ref{sec:related_work}, we review two broader lines of work: (1) content-aware layout generation and (2) graphic design composition.

\textit{Content-aware layout generation.}
Content-aware layout generation arranges predefined elements such as text, logos, and underlays on a given background canvas~\citep{hsu2023posterlayoutnewbenchmarkapproach,zhang2024vascar,hsu2025postero}. Building on this setting, PosterLlama~\citep{seol2024posterllamabridgingdesignability} and PosterLLaVA~\citep{yang2024posterllavaconstructingunifiedmultimodal} leverage LLMs and MLLMs to support more flexible layout conditions. Retrieval-augmented approaches, such as RALF~\citep{horita2024retrievalaugmentedlayouttransformercontentaware} and LayoutCoT~\citep{shi2026layoutcotunleashingdeepreasoning}, retrieve reference layouts using background information or available layout constraints. In these settings, the background content or layout conditions provide spatial cues for plausible element placement. In contrast, our setting starts from a blank canvas and requires determining the positions and scales of all supplied elements without such observable spatial cues. We therefore infer coarse layout structure from the input elements themselves and retrieve structurally similar references.

\textit{Graphic design composition.}
Another line of work composes complete editable documents from multimodal elements.
FlexDM~\citep{inoue2023flexiblemultimodaldocumentmodels}, Graphist~\citep{cheng2024graphicdesignlargemultimodal}, LaDeCo~\citep{lin2024elementsdesignlayeredapproach}, and AesthetiQ~\citep{patnaik2025aesthetiqenhancinggraphiclayout} predict structured layouts for supplied visual and textual elements.
Other works, including COLE~\citep{jia2024colehierarchicalgenerationframework}, OpenCOLE~\citep{inoue2024opencolereproducibleautomaticgraphic}, CreatiPoster~\citep{hong2026creatipostereditablecontrollablemultilayer}, and IGD~\citep{qu2025igdinstructionalgraphicdesign}, additionally generate or retrieve visual elements as part of the design process.
Our work specifically focuses on composing a fixed inventory of supplied elements, without generating or retrieving additional elements.
While effective at generating editable documents, existing graphic design composition methods do not fully capture the iterative workflow of real-world design, as they provide limited support for revising composition decisions based on visual feedback from rendered results.

\section{Data Construction Details}
\label{sec:data_construction_details}

\subsection{Definition of Functional Group Roles}
We define a closed set of six functional roles to characterize the semantic and compositional functions of each group. As described in Sec.~\ref{sec:data_construction}, elements are first grouped according to their compositional relationships, and a role is then assigned to each resulting group according to the definitions below. 

\paragraph{Background.}
A \texttt{background} group establishes the canvas-wide base layer of a design and remains behind the main foreground content. Its defining characteristics are broad canvas coverage, placement beneath the main content in the stacking order, and a primarily foundational rather than independently semantic role. A background may consist of a single element or multiple adjacent, tiled, split, or overlapping elements that together form a continuous backdrop. Plain solid-color fields or visually indistinguishable elements are included when they satisfy these criteria, whereas foreground panels, cards, overlays, or independently meaningful visual elements are excluded even if they occupy a large portion of the canvas. When no supplied element or combination of elements forms such a base layer, the canvas itself is treated as the implicit background.

\paragraph{Visual/hero.}
A \texttt{visual/hero} serves as a semantic visual object or anchor in the composition, regardless of whether it is the primary or a secondary focal element.
Its defining characteristic is that it carries meaningful visual content rather than functioning purely as background or decoration.
Typical examples include photographs, products, people, scenes, illustrations, diagrams, and message-bearing visual containers such as signs, badges, placards, quote cards, emblems, and typographic-poster frames.
When such a visual also contains independently planned \texttt{title/headline}, \texttt{content}, \texttt{support}, or \texttt{decorative} elements, those elements remain in their own functional groups rather than being absorbed into the \texttt{visual/hero}.
Multiple components are treated as a single \texttt{visual/hero} only when they function as one inseparable visual object rather than as independently editable semantic layers.

\paragraph{Title/headline.}
A \texttt{title/headline} is an attention-leading typographic unit that establishes a distinct semantic and visual anchor in the design.
It typically includes titles, headlines, slogans, taglines, or tightly coupled subtitles, including those placed within a \texttt{visual/hero} container.
Small visual elements such as underlines, highlights, ornaments, or simple backings are included in the same group when they are structurally attached to and primarily serve the headline.
Text whose main purpose is informational, functional, or supportive rather than attention-leading is assigned to another role.

\paragraph{Content.}
A \texttt{content} group conveys information whose primary purpose is to be read or understood.
It includes body text, supporting copy, quotes, captions, descriptions, dates, times, venues, schedules, factual information, and non-promotional pricing.
Text belongs to \texttt{content} when it neither functions as the dominant headline nor primarily serves branding, contact, action, or other utility purposes.

\paragraph{Support.}
A \texttt{support} group provides identity, action, attribution, or other functional information that supports delivery of the main message.
Typical examples include logos, brand identifiers, sponsor marks, websites, contact information, social handles, addresses, QR codes, calls-to-action, buttons, promotional offers, discounts, legal notices, credits, and footer material.
Its defining characteristic is functional utility rather than visual prominence, so a \texttt{support} group may still occupy a visually salient position in the composition.

\paragraph{Decorative.}
A \texttt{decorative} group provides non-semantic visual embellishment that contributes to the overall composition without primarily carrying content, identity, or action.
It includes independently composed ornaments, accent systems, atmospheric overlays, and standalone decorative motifs that span the design or have no clear semantic host.
Decorative elements that specifically support another functional group---such as a border, divider, frame, highlight, accent shape, pattern, or ornamental icon attached to a \texttt{title/headline}, \texttt{visual/hero}, \texttt{content}, or \texttt{support}---are instead included in that group.
Multiple ornaments that jointly form one coherent decorative system are grouped together rather than treated as separate \texttt{decorative} groups.

\subsection{Detailed Data Construction Pipeline}
\label{sec:detailed_data_construction}
\begin{figure}[p]
    \centering
    \includegraphics[width=0.98\linewidth]{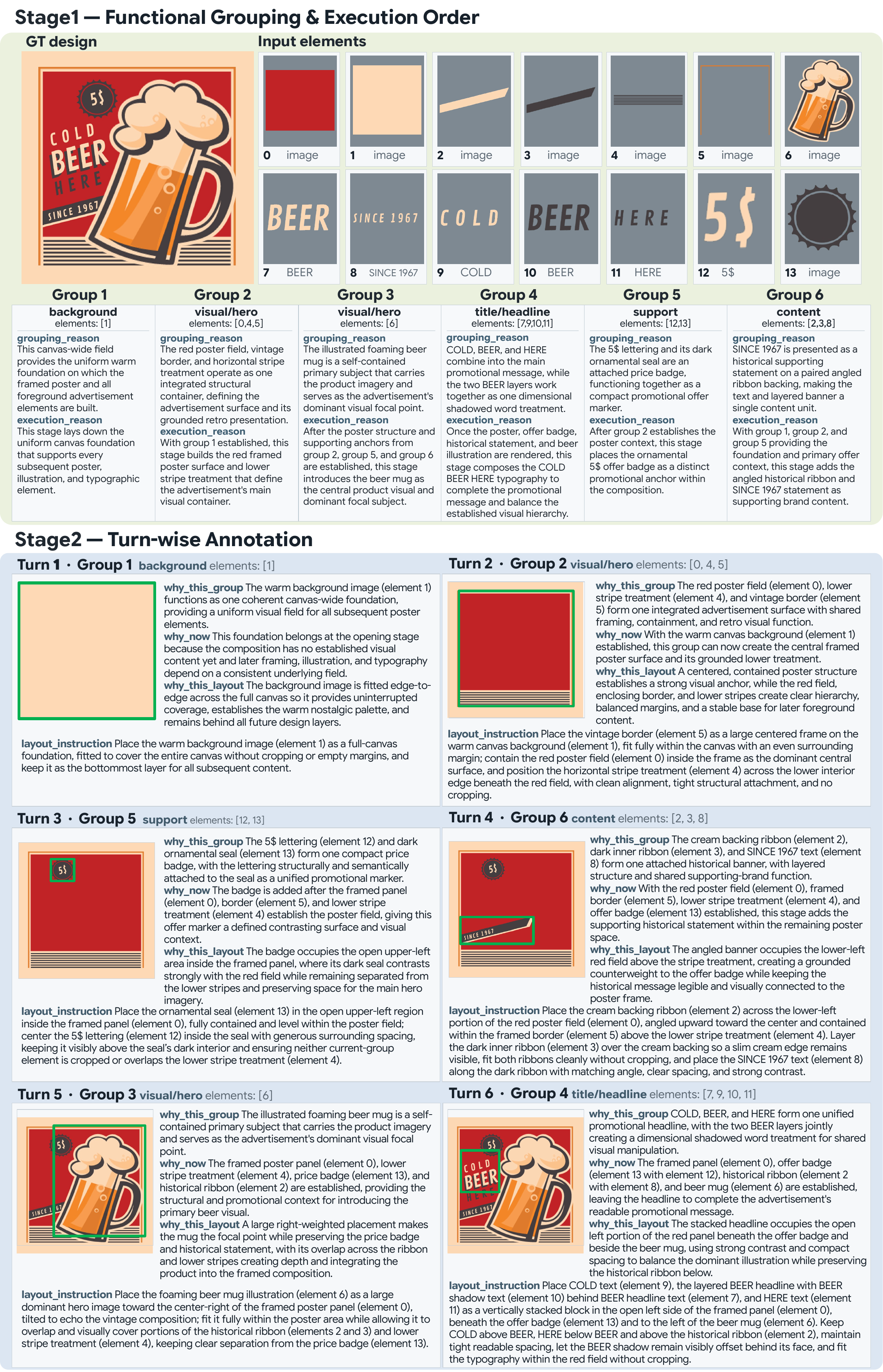}
    \vspace{-5pt}
    \caption{\textbf{Detailed Data Construction Pipeline.} Stage~1 constructs functional groups and their execution order, and Stage~2 derives a turn-level plan for each group.}
    \label{fig:annotation_example}
    \vspace{-10pt}
\end{figure}

Fig.~\ref{fig:annotation_example} provides a detailed example of our two-stage data construction pipeline.
In Stage~1, the input elements are organized into functional groups with their roles and execution order.
In Stage~2, each execution-ordered group is converted into a turn-level plan containing the grouping, execution, and layout rationales together with a layout instruction.
The resulting sequence provides group-wise supervision for progressively constructing the complete design.

\subsection{Group and Execution Statistics}
\begin{figure}[H]
    \centering
    \vspace{-5pt}
    \begin{subfigure}[t]{0.49\linewidth}
        \centering
        \includegraphics[width=\linewidth]{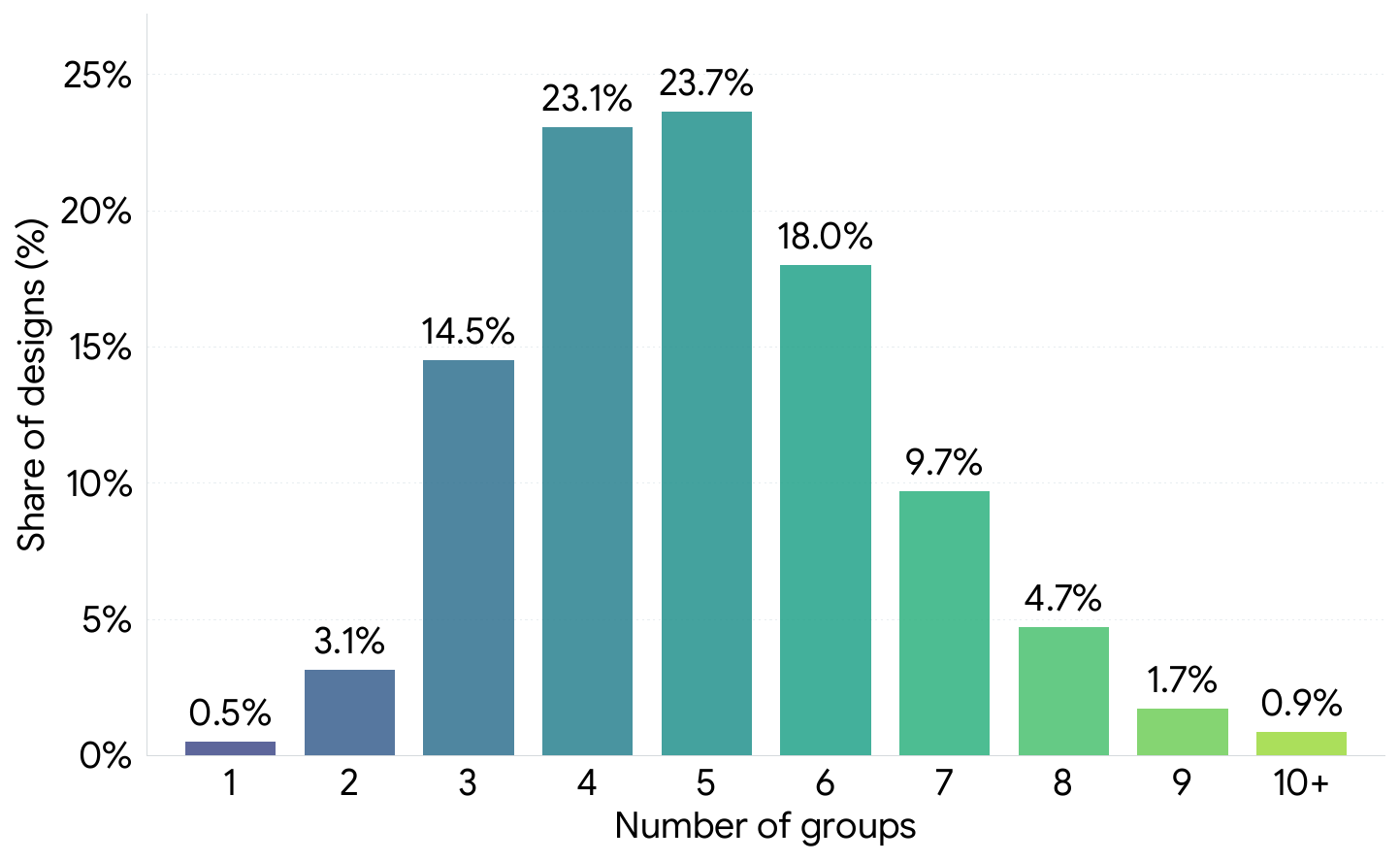}
        \caption{Number of functional groups per design.}
        \label{fig:groups_per_design}
    \end{subfigure}
    \hfill
    \begin{subfigure}[t]{0.49\linewidth}
        \centering
        \includegraphics[width=\linewidth]{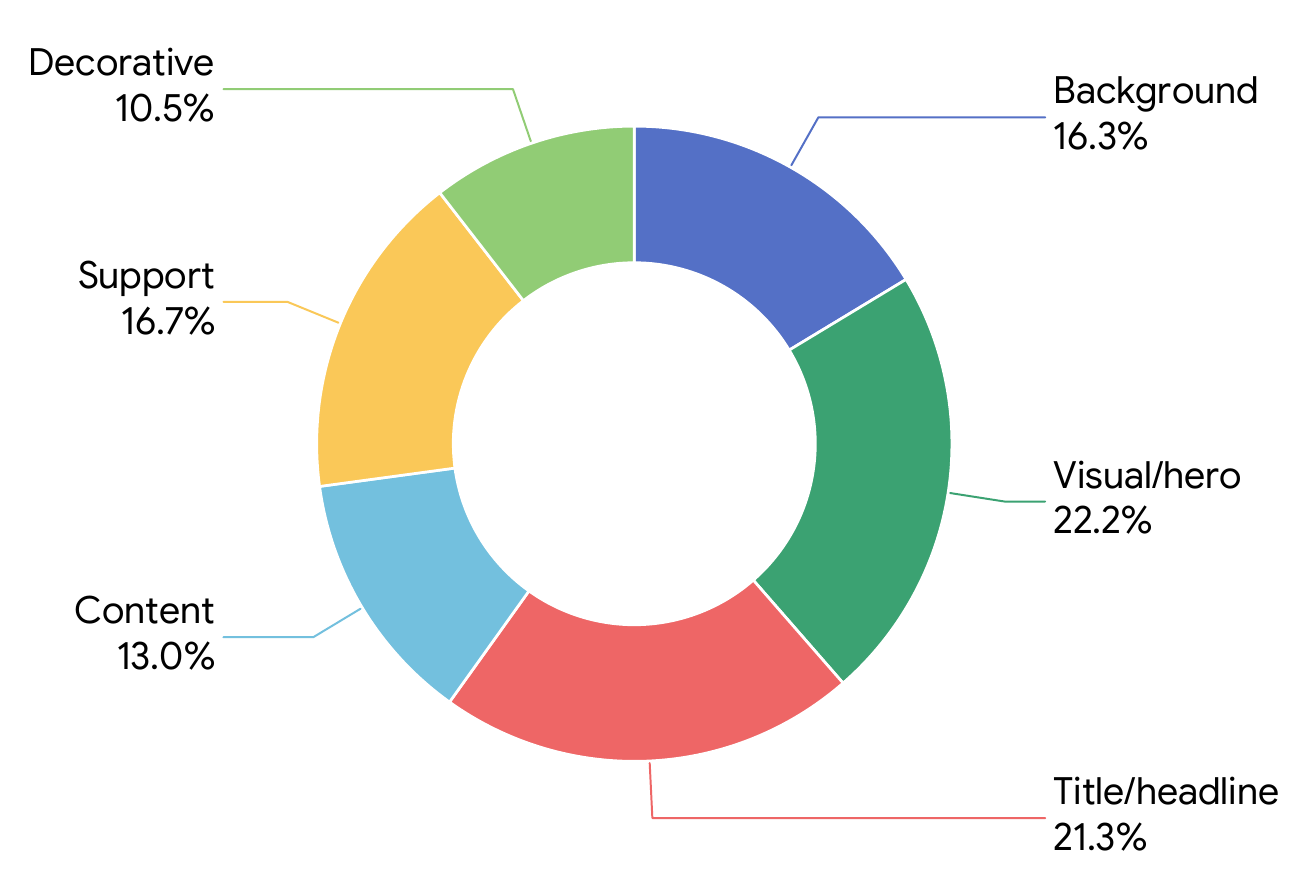}
        \caption{Distribution of functional group roles.}
        \label{fig:role_distribution}
    \end{subfigure}
    \caption{\textbf{Statistics of the constructed functional groups.}}
    \label{fig:group_statistics}
    \vspace{-10pt}
\end{figure}

We analyze the functional groups and execution orders constructed by our annotation pipeline to characterize the supervision used for group-wise layout generation. As shown in Fig.~\ref{fig:groups_per_design}, designs are typically decomposed into multiple functional groups, with four to six groups accounting for the majority of the data. Fig.~\ref{fig:role_distribution} further shows that all six predefined functional roles are represented, with no single role dominating the resulting groups. Together, these statistics show that the constructed supervision decomposes complete layouts into multiple functionally diverse group-level decisions rather than treating all elements as a single prediction target.

\begin{figure}[H]
    \centering
    \vspace{-5pt}
    \begin{subfigure}[t]{0.49\linewidth}
        \centering
        \includegraphics[width=\linewidth]{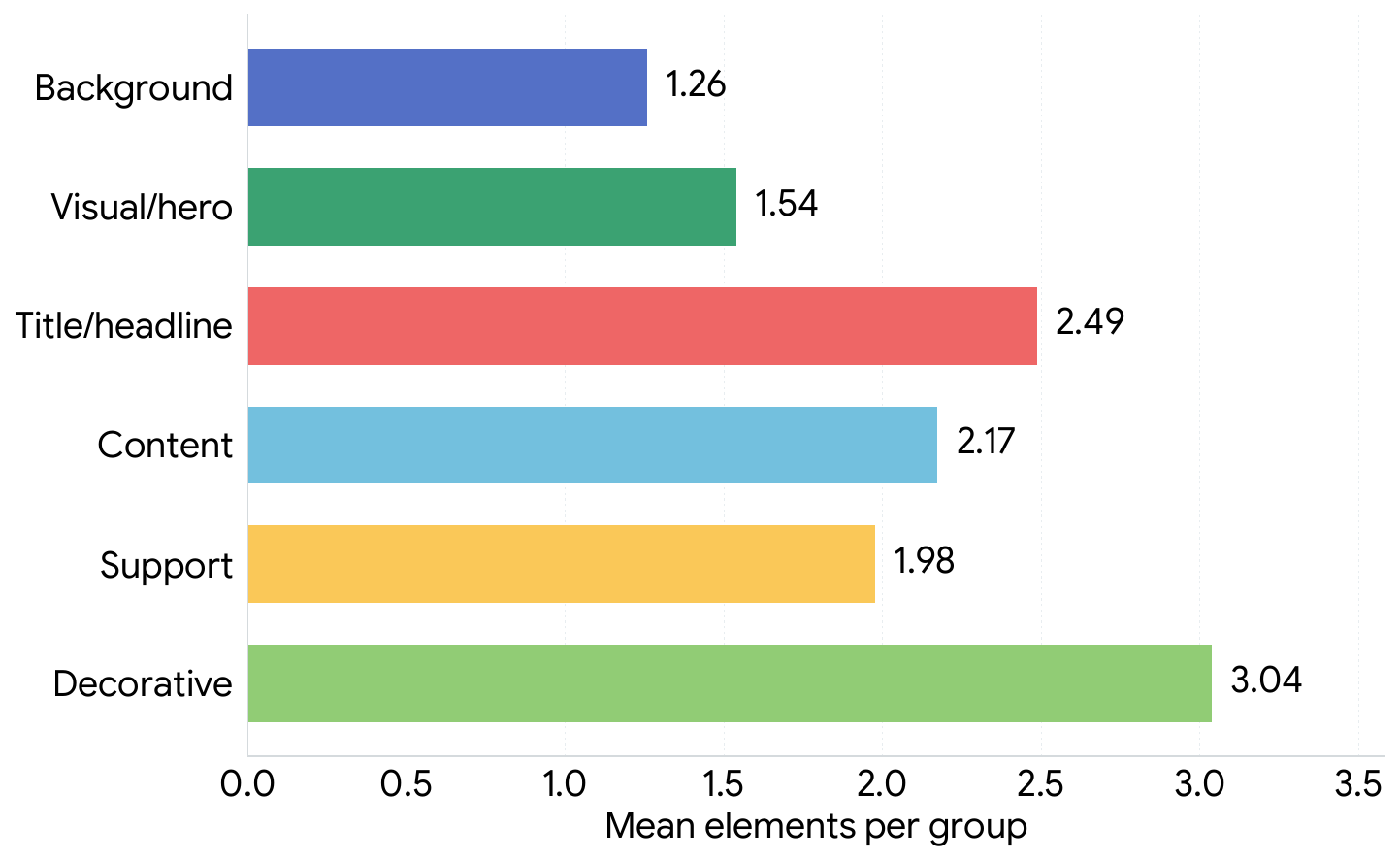}
        \caption{Mean group size by functional role.}
        \label{fig:elements_per_role}
    \end{subfigure}
    \hfill
    \begin{subfigure}[t]{0.49\linewidth}
        \centering
        \includegraphics[width=\linewidth]{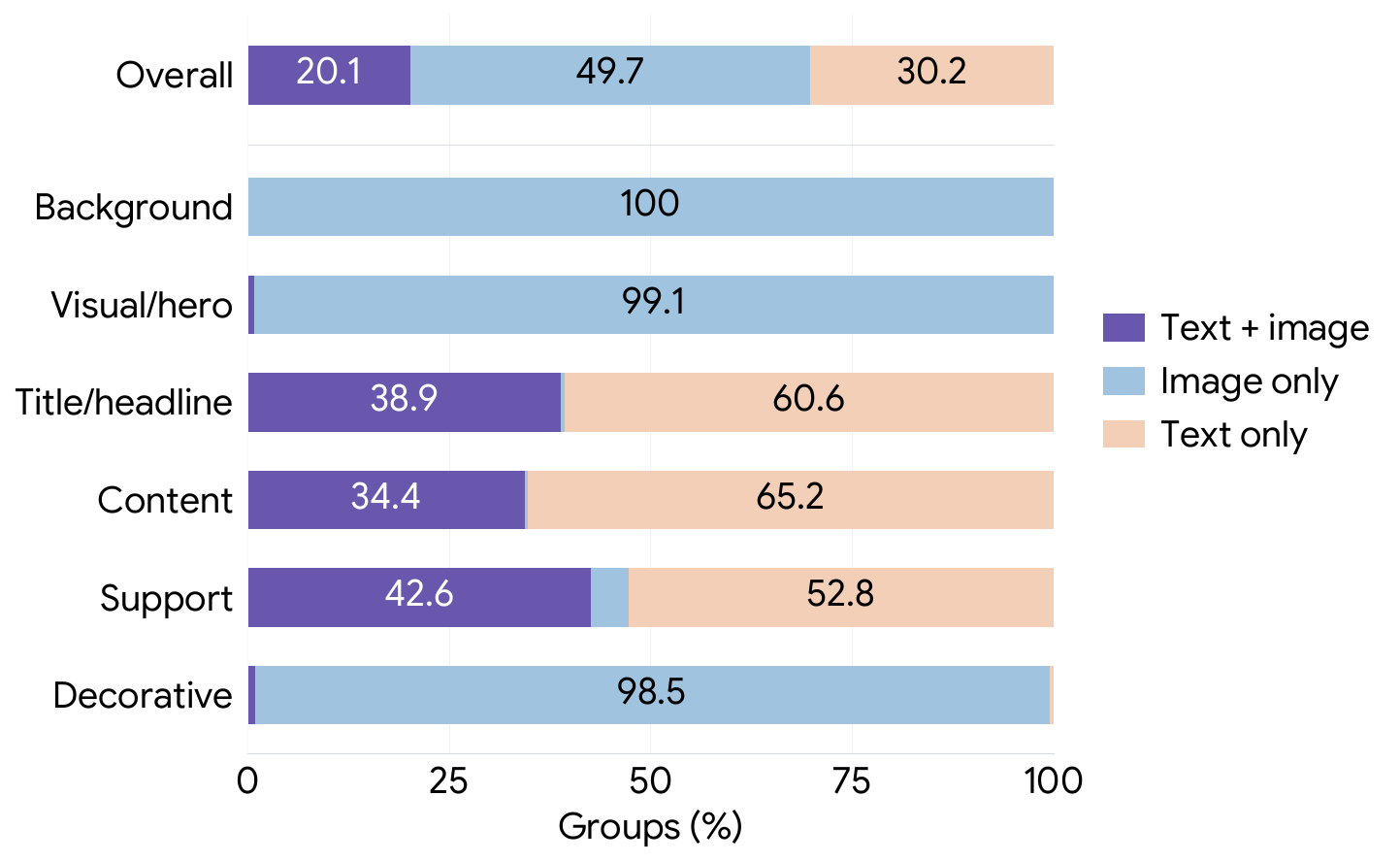}
        \caption{Modality composition by functional role.}
        \label{fig:group_modality}
    \end{subfigure}
    \caption{\textbf{Group size and modality composition by functional role.}}
    \label{fig:group_composition}
    \vspace{-10pt}
\end{figure}

The constructed groups also vary substantially in their internal composition. As shown in Fig.~\ref{fig:elements_per_role}, the average number of elements per group differs across functional roles: background and visual/hero groups are typically compact, whereas decorative, title/headline, and content groups tend to contain more elements. 

Fig.~\ref{fig:group_modality} further shows that our grouping is not constrained by modality. While background, visual/hero, and decorative groups are almost exclusively image-only, substantial portions of title/headline (38.9\%), content (34.4\%), and support (42.6\%) groups contain both text and image elements. This indicates that functionally associated text and visual elements can form a common compositional unit and be planned together.

\begin{figure}[h]
    \centering
    \includegraphics[width=0.7\linewidth]{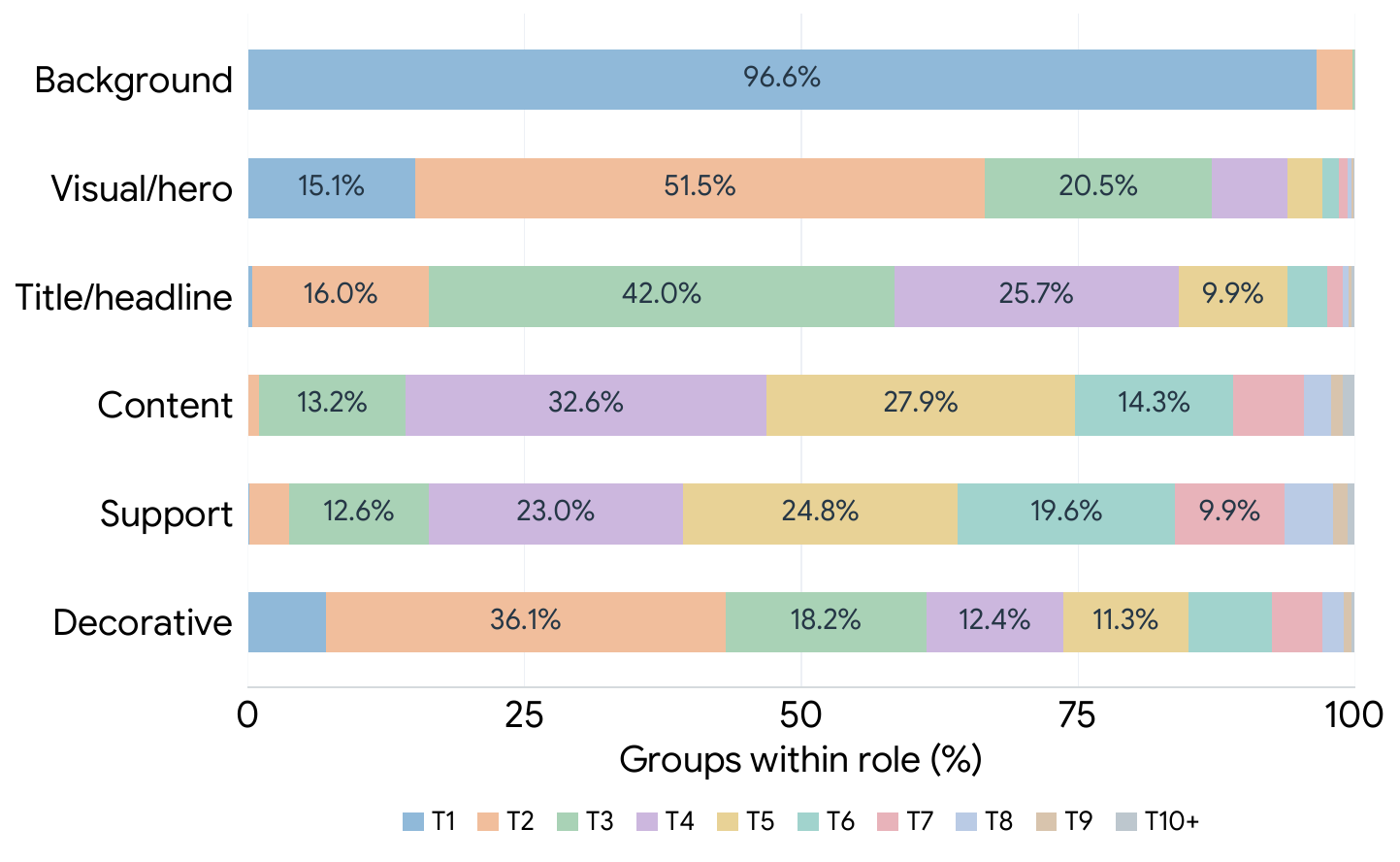}
    \vspace{-10pt}
    \caption{\textbf{Execution-turn distribution by functional role.}
    Each row shows the distribution of execution turns for the corresponding functional role. T1, T2, $\ldots$ denote the first, second, $\ldots$ group-wise planning turns, and T10+ denotes turn 10 or later.}
    \label{fig:execution_by_role}
    \vspace{-5pt}
\end{figure}

The execution annotations also exhibit a clear role-dependent structure. As shown in Fig.~\ref{fig:execution_by_role}, background groups are overwhelmingly placed at the beginning of the trajectory, and visual/hero groups are concentrated in early turns. In contrast, content and support groups are distributed more heavily toward the middle and later stages, while decorative groups span a broader range of turns. These patterns indicate that the annotated execution order is closely associated with the functional role of each group, yielding structured construction trajectories in which foundational visual components are generally established before subsequent content and supporting elements.

\subsection{Prompts used for Data Construction}
\paragraph{Stage 1 prompt.}
An abbreviated Stage~1 prompt is provided below.
\texttt{\{GT\_IMAGE\}}, \texttt{\{CANVAS\}},
\texttt{\{DOCUMENT\_TITLE\}}, \texttt{\{DOCUMENT\_INTENTION\}}, and
\texttt{\{ALL\_ELEMENTS\}} denote sample-specific inputs substituted at runtime.
Image elements are provided as interleaved image blocks within the element inventory.
Ellipses mark content omitted for brevity and are not part of the actual prompt.

\begin{vlmpromptbox_stage1}
    \begingroup
\fontsize{6.5}{7.2}\selectfont
\ttfamily
\setlength{\parindent}{0pt}
\setlength{\parskip}{1.5pt}

You are an expert graphic designer specializing in visual layout analysis and
document composition.

Your primary task is to semantically group every supplied element into coherent
visual design blocks from a designer's perspective, then determine their
dependency-aware execution order for a future multi-turn layout planner.

Treat grouping as a design understanding problem rather than a geometric
clustering problem. Elements should be grouped into the coarsest
dependency-consistent layout blocks that a professional designer would plan and
render in one turn. Avoid both element-level fragmentation and module-level
over-merging.

\medskip
\textbf{[GT RENDERED DESIGN]}

\{GT\_IMAGE\}

Treat the GT rendered design as the primary oracle evidence for semantic
grouping, functional roles, structural attachment, common manipulation, visual
hierarchy, and planning dependency.

Do not restate the observed final layout or mention exact coordinates, bboxes,
dimensions, or pixel-level placement.

\medskip
\textbf{[CANVAS]}

\{CANVAS\}

[\ldots\ background identification rules omitted \ldots]

\medskip
\textbf{[DOCUMENT TITLE]}

\{DOCUMENT\_TITLE\}

\medskip
\textbf{[DOCUMENT INTENTION]}

\{DOCUMENT\_INTENTION\}

\medskip
\textbf{[ALL ELEMENTS]}

\{ALL\_ELEMENTS\}

Use element bboxes as structural evidence for canvas coverage, containment,
and background identification. Geometry and source-list order should otherwise
remain secondary evidence.

The \texttt{order} field indicates the ground-truth global stacking order.
Use it primarily for front-to-back relationships and internal stacking, not as
the default grouping or execution order.

[\ldots]

\medskip
\textbf{[TASK]}

1. Partition all supplied element IDs into semantic/layout groups, assigning
every element to exactly one group.

2. Assign contiguous integer \texttt{group\_id} values starting at 0.

3. Assign exactly one closed-set functional role to every group.

4. Determine a dependency-aware \texttt{execution\_order} for the future
multi-turn planner.

5. Explain why each set of elements forms one coherent semantic/layout group.

6. Explain why each group should be planned at its selected execution stage.

\medskip
\textbf{[GROUP GRANULARITY REQUIREMENTS]}

Form the coarsest block that remains semantically and executionally homogeneous.

Merge elements only when they are normally positioned and revised as one unit
and share the same planning dependency. A shared panel, bounded region,
alignment system, reading flow, or local composition is supporting evidence,
but is not sufficient by itself.

A single panel, card, badge, placard, or bounded module may contain multiple
groups. Separate a visual or structural anchor from content that can be planned
independently relative to it.

Split elements when subsets have distinct semantic functions, anchors,
alignment systems, or execution dependencies.

Spatial proximity or shared containment alone does not justify merging.

Decorative elements attached to a specific semantic block should be grouped
with that block; create a separate \texttt{decorative block} only for
independently composed decoration serving the overall composition.

[\ldots]

\medskip
\textbf{[EXECUTION ORDER REQUIREMENTS]}

\texttt{execution\_order} is the order in which the sequential planner should
plan and render the groups.

Prefer the following dependency structure when supported by the design:

1. canvas-filling foundations and backgrounds,

2. groups that establish stable visual or typographic anchors,

3. groups that depend on those rendered anchors,

4. secondary information and support groups,

5. finishing accents or foreground decorations.

Within a bounded module, plan a visual container or hero anchor before title,
content, support, or decorative groups whose placement depends on it.

For each group, \texttt{execution\_reason} should explain why it is planned at
that stage and, when applicable, which rendered anchor it depends on or
establishes for later groups.

[\ldots]

\medskip
\textbf{[CLOSED-SET GROUP ROLES]}

Choose exactly one role from:

- \texttt{background block}

- \texttt{visual/hero block}

- \texttt{title/headline block}

- \texttt{content block}

- \texttt{support block}

- \texttt{decorative block}

[\ldots\ detailed role definitions omitted \ldots]

Determine group membership from planner actions and dependencies before
assigning roles. Multiple groups may share the same non-background role only
when they form genuinely separate local compositions with independent anchors
or execution dependencies.

\medskip
\textbf{[STRICT OUTPUT CONTRACT]}

Return exactly one valid JSON object with top-level keys
\texttt{groups} and \texttt{execution\_order}.

Each group contains:
\texttt{group\_id}, \texttt{role}, \texttt{elements}, and
\texttt{grouping\_reason}.

Each execution-order entry contains:
\texttt{group\_id} and \texttt{execution\_reason}.

Across all groups, \texttt{elements} must form an exact partition of the
supplied element IDs, and every valid \texttt{group\_id} must appear exactly
once in \texttt{execution\_order}.

[\ldots\ additional JSON-format validation and output example omitted \ldots]

\endgroup
\end{vlmpromptbox_stage1}

\paragraph{Stage 2 prompt.}
An abbreviated Stage~2 prompt is provided below.
\texttt{\{GT\_IMAGE\}}, \texttt{\{CURRENT\_RENDER\_IMAGE\}}, and
\texttt{\{NEXT\_STATE\_RENDER\_IMAGE\}} denote the final, current, and oracle next-state renders.
The remaining placeholders provide sample-specific document context, previous-group state, and current-turn group information.
Image elements are supplied as interleaved image blocks within the element inventories.

\begin{vlmpromptbox_stage2}
    \begingroup
\fontsize{6.5}{7.2}\selectfont
\ttfamily
\setlength{\parindent}{0pt}
\setlength{\parskip}{1.5pt}

You are an expert sequential graphic-layout oracle annotator.

Your task is to generate one turn-level layout plan for a fixed design group in
a sequential layout generation process.

The semantic grouping and execution order are already determined by a previous
oracle decomposition stage. Do not reconsider or modify them. Infer how the
fixed current group should be placed and rendered at this execution stage.

Generate a layout plan that a sequential layout planner can directly follow.

\medskip
\textbf{[DOCUMENT CONTEXT]}

\textbf{[DOCUMENT TITLE]}

\{DOCUMENT\_TITLE\}

\textbf{[DOCUMENT INTENTION]}

\{DOCUMENT\_INTENTION\}

\textbf{[CANVAS]}

\{CANVAS\}

\medskip
\textbf{[FINAL GT DESIGN]}

\{GT\_IMAGE\}

Use the final GT design only as global context for visual hierarchy, design
intent, style, and major compositional relationships. Do not use elements that
are not yet rendered as current anchors.

\medskip
\textbf{[CURRENT RENDER]}

\{CURRENT\_RENDER\_IMAGE\}

CURRENT RENDER represents the sequential design state before adding the current
group. Use it to identify existing anchors, available space, hierarchy, and
current spatial relationships.

\medskip
\textbf{[PREVIOUSLY EXECUTED GROUPS]}

\{PREVIOUS\_GROUPS\}

\textbf{[PREVIOUSLY RENDERED ELEMENTS]}

\{PREVIOUS\_ELEMENT\_LINES\}

These groups and elements are already present in CURRENT RENDER and provide the
existing visual context for the current turn.

\medskip
\textbf{[ORACLE NEXT-STATE RENDER]}

\{NEXT\_STATE\_RENDER\_IMAGE\}

ORACLE NEXT-STATE RENDER shows the result after adding only the fixed current
group to CURRENT RENDER.

Use the visual difference between CURRENT RENDER and ORACLE NEXT-STATE RENDER
as the primary evidence for the current group's target region, anchor
relationship, relative scale, spacing, alignment, overlap, containment, crop or
fit behavior, and intra-group stacking.

\medskip
\textbf{[FIXED CURRENT GROUP]}

\{CURRENT\_GROUP\}

\textbf{[FIXED EXECUTION REASON]}

\{CURRENT\_EXECUTION\_REASON\}

The current group, its role, element membership, grouping rationale, and
execution logic are fixed and must not be changed.

\medskip
\textbf{[CURRENT GROUP ELEMENTS]}

\{CURRENT\_ELEMENT\_LINES\}

These are the only elements added in this turn.

\medskip
\textbf{[EVIDENCE USE]}

Use:
- the fixed group and execution reason for grouping and execution semantics,

- CURRENT RENDER and previously rendered elements for existing anchors,

- the CURRENT-to-NEXT render difference for current-group placement,

- FINAL GT DESIGN only for global composition and design-intent context.

[\ldots]

\medskip
\textbf{[TASK]}

Generate exactly one turn-local layout plan for the fixed current group.

Plan and add only the current-group elements. Use qualitative and relational
layout language; do not output coordinates, bounding boxes, percentages,
dimensions, pixel distances, or other raw geometric measurements.

Do not mention GT, oracle evidence, annotation, or internal metadata in the
output.

\medskip
\textbf{[RATIONALE REQUIREMENTS]}

The rationale must contain exactly:

- \texttt{why\_this\_group}: why the fixed elements form one coherent planning
unit,

- \texttt{why\_now}: why the group belongs at this execution stage given the
already established composition,

- \texttt{why\_this\_layout}: why the recovered placement and arrangement are
appropriate for the current canvas.

Keep these three functions distinct. Ground explanations in semantic
relationships, visual anchors, hierarchy, readability, spacing, alignment,
containment, overlap, balance, or whitespace as appropriate.

Preserve the original grouping and execution decisions and do not propose
alternative groups, stages, or layouts.

Each rationale should normally be one sentence and must not exceed two
sentences.

[\ldots]

\medskip
\textbf{[LAYOUT INSTRUCTION REQUIREMENTS]}

\texttt{layout\_instruction} must be an imperative instruction describing how
to add only the current group.

Describe, when relevant:

1. target region and primary anchor,

2. intra-group arrangement,

3. relative scale, alignment, and spacing,

4. overlap, containment, stacking, and crop or fit behavior.

Describe what the planner should do rather than why it is appropriate.
Mention every current-group element at least once using descriptive references,
e.g., \texttt{headline text (element 7)}.

When referring to an existing anchor, identify a concrete previously rendered
element. Do not reference nonexistent elements.

[\ldots]

\medskip
\textbf{[STRICT OUTPUT CONTRACT]}

Return exactly one valid JSON object with top-level key
\texttt{turn\_plan}.

\texttt{turn\_plan} must contain exactly:

- \texttt{group\_id}

- \texttt{role}

- \texttt{elements}

- \texttt{rationale}

- \texttt{layout\_instruction}

\texttt{rationale} must contain exactly:

- \texttt{why\_this\_group}

- \texttt{why\_now}

- \texttt{why\_this\_layout}

The \texttt{group\_id}, \texttt{role}, and \texttt{elements} must exactly match
the fixed current group.

Return concise English and valid JSON only.

[\ldots\ additional validation rules and output example omitted \ldots]

\endgroup
\end{vlmpromptbox_stage2}

\section{Main Pipeline Details}
\label{sec:main_pipeline_details}

\subsection{Overall Generation Algorithm}

\subsubsection{Overall Algorithm}
\begin{algorithm}[H]
    \caption{\ours Inference}
    \label{alg:playground}
    \begin{algorithmic}[1]

\Require
Element inventory $\mathcal{E}$, document metadata $\mathcal{M}$,
reference pool $\mathcal{D}_{\mathrm{ref}}$, renderer $\mathcal{R}$

\Require
Prompts $\mathcal{P}_\text{ret},\mathcal{P}_\text{plan},
\mathcal{P}_\text{pred},\mathcal{P}_\text{ver}$;
trained modules $f_{\theta_\text{ret}}$,
$f_{\theta_\text{plan}}$, $f_{\theta_\text{pred}}$;
verifier $f_\text{ver}$;
number of candidates $K$;
maximum verification rounds $R_{\max}=3$

\Ensure
Selected layout predictions $\{\mathcal{Y}_t\}_{t=1}^{T}$
and final canvas $C_T$

\vspace{2pt}
\State $\Phi
\gets
f_{\theta_\text{ret}}
(\mathcal{P}_\text{ret},\mathcal{M},\mathcal{E})$
\Comment{\textbf{1. Structure prediction}}

\State $\mathcal{Z}
\gets$ top-5 designs from $\mathcal{D}_\text{ref}$
ranked by Eq.~\ref{eq:retrieval_score}
\Comment{\textbf{2. Reference retrieval}}

\State $C_0 \gets$ empty canvas,\quad
$\mathcal{E}^\text{rem}_1 \gets \mathcal{E}$,\quad
$t\gets1$

\While{$\mathcal{E}^\text{rem}_t \neq \emptyset$}

    \State
    $(G_t,p_t)
    \gets
    f_{\theta_\text{plan}}
    \left(
    \mathcal{P}_\text{plan},
    \mathcal{M},
    \mathcal{Z},
    C_{t-1},
    \mathcal{E}^\text{rem}_t
    \right)$
    \Comment{\textbf{3. Planning} $\rightarrow$ form and plan next group}
    
    \State Let $\ell_t$ denote the layout instruction in $p_t$ and set $\ell_t^{(1)} \gets \ell_t$

    \For{$r=1,\ldots,R_{\max}$}
        \State Set the layout instruction in $p_t$ to $\ell_t^{(r)}$
    
        \State
        $\left\{
        \mathcal{Y}_t^{(r,k)}
        \right\}_{k=1}^{K}
        \gets
        \left\{
        f_{\theta_{\mathrm{pred}}}
        \left(
        \mathcal{P}_{\mathrm{pred}},
        \mathcal{M},
        \mathcal{Z},
        C_{t-1},
        G_t,
        p_t
        \right)
        \right\}_{k=1}^{K}$
        \Statex \Comment{\textbf{4. Layout generation} $\rightarrow$ $K$ candidates}

        \State
        $\mathcal{C}_t^{(r)}
        \gets
        \left\{
        \mathcal{R}
        \left(
        C_{t-1},
        \mathcal{Y}_t^{(r,k)}
        \right)
        \right\}_{k=1}^{K}$
        \Comment{Render candidate layouts}

        \State
        $\left(
        \bar{C}_t^{(r)},
        d_t^{(r)}
        \right)
        \gets
        f_\text{ver}
        \left(
        \mathcal{P}_\text{ver},
        G_t,
        \ell_t^{(r)},
        C_{t-1},
        \mathcal{C}_t^{(r)}
        \right)$
        \Comment{\textbf{5. Verification} $\rightarrow$ round $r$}

        \If{$d_t^{(r)}=\texttt{accept}$}
            \State $C_t \gets \bar{C}_t^{(r)}$
            \State $\mathcal{Y}_t \gets$
            prediction corresponding to $\bar{C}_t^{(r)}$
            \State \textbf{break}
        \ElsIf{$r<R_{\max}$}
            \State
            $\ell_t^{(r+1)}
            \gets$
            revised instruction returned by $f_{\mathrm{ver}}$
            \Comment{Retry the same group}
        \EndIf

    \EndFor

    \If{$d_t^{(r)}\neq\texttt{accept}$ for all
    $r=1,\ldots,R_{\max}$}
        \State Visual Verifier selects
        $C_t$ from
        $\{\bar{C}_t^{(1)},\ldots,\bar{C}_t^{(R_{\max})}\}$
        \State $\mathcal{Y}_t \gets$
        prediction corresponding to $C_t$
        \Comment{Final selection across rounds}
    \EndIf

    \State
    $\mathcal{E}^\text{rem}_{t+1}
    \gets
    \mathcal{E}^\text{rem}_t
    \setminus
    \mathcal{E}_t$
    \Comment{\textbf{6. State update} $\rightarrow$ remove placed group}

    \State $t\gets t+1$
    \Comment{$C_t$ becomes the previous canvas for the next turn}

\EndWhile

\State $T\gets t-1$
\State \Return $\{\mathcal{Y}_t\}_{t=1}^{T},\,C_T$
\Comment{\textbf{7. Final editable design}}

\end{algorithmic}
\end{algorithm}

Algorithm~\ref{alg:playground} summarizes the complete inference procedure of \ours. The Structure-Aware Retriever first retrieves five reference designs, which remain fixed throughout generation. At each turn, the Planner forms the next functional group and its plan from the current canvas and remaining elements, after which the Layout Generator produces $K$ candidate layouts for the selected group. The Visual Verifier then either accepts the best rendered candidate or revises the layout instruction for regeneration, for up to $R_{\max}=3$ rounds. The accepted canvas and updated remaining elements define the next state, and this process repeats until all input elements are placed.

\subsubsection{Visualization of Overall Process}
\begin{figure}[p]
    \centering
    \includegraphics[width=0.98\linewidth]{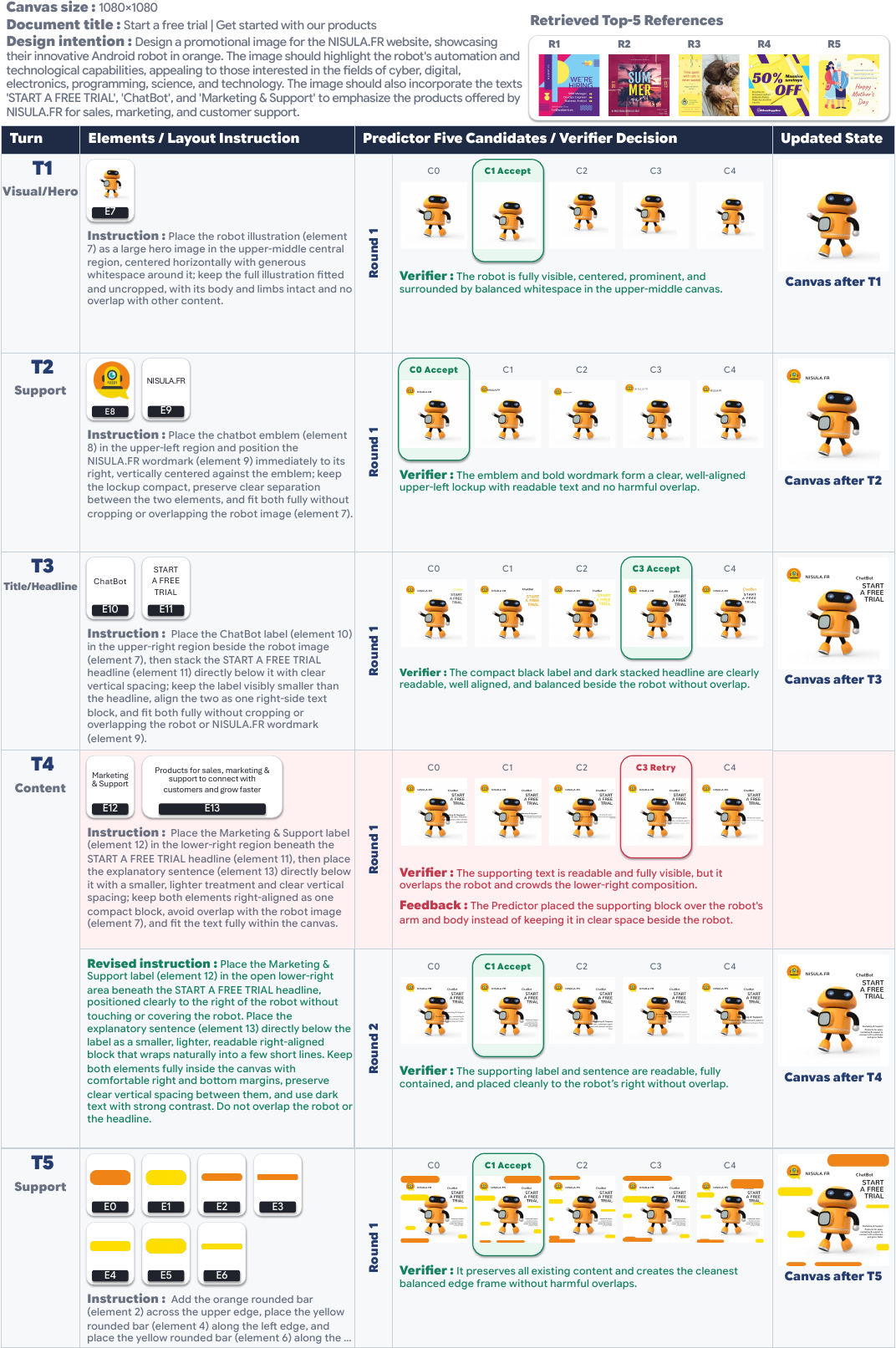}
    \vspace{-5pt}
    \caption{\textbf{Detailed example of Progressive Design Generation.} }
    \label{fig:pipeline_example}
    \vspace{-10pt}
\end{figure}
Fig.~\ref{fig:pipeline_example} visualizes a complete generation trajectory of \ours.
The retrieved references are fixed before generation, and each turn shows the selected functional group and layout instruction, the $K$ candidate layouts produced by the Layout Generator, the Visual Verifier decision, and the resulting updated canvas.
As generation progresses from turn 1 to turn 5, the accepted canvas is repeatedly carried forward as the visual state for the next planning turn.

The example also illustrates the feedback loop within a turn.
At turn 4, the verifier rejects the initially selected candidate because the newly placed text overlaps the existing robot and crowds the lower-right composition.
It therefore revises the layout instruction and regenerates the same group; the second-round candidate is then accepted after placing the text in a clearer region.
This example highlights how progressive planning and turn-level verification jointly construct the design while adapting to the evolving canvas.

\subsection{Structure-Aware Retriever}

\subsubsection{Retrieval Score}
\label{sec:retrieval-score}

\paragraph{Role-aware group IoU.}
For the query, we use the predicted role--region structure
$\Phi_q$ from Eq.~\ref{eq:retriever_output}.
For each candidate $c\in\mathcal{D}_{\mathrm{ref}}$, we use its Stage~1
role--region structure
$\Phi_c=\{(r_{c,j},B_{c,j})\}_{j=1}^{J_c}$.
Before comparison, each group region is clipped to the corresponding canvas
bounds and normalized by the canvas size:
\begin{equation}
\bar B_{d,j}
=
\left(
\frac{x_{d,j}}{w_d},
\frac{y_{d,j}}{h_d},
\frac{\Delta x_{d,j}}{w_d},
\frac{\Delta y_{d,j}}{h_d}
\right),
\qquad d\in\{q,c\},
\end{equation}
where $w_d$ and $h_d$ denote the canvas width and height, respectively, and
$(x_{d,j},y_{d,j},\Delta x_{d,j},\Delta y_{d,j})$ denotes the clipped group
region in $(x,y,\text{width},\text{height})$ form.

For $J_q$ query groups and $J_c$ candidate groups, we define the role-aware
pairwise overlap as
\begin{equation}
M_{ij}
=
\mathbf{1}[r_{q,i}=r_{c,j}]
\operatorname{IoU}(\bar B_{q,i},\bar B_{c,j}).
\end{equation}
Let $\Pi_{q,c}$ denote the set of one-to-one matchings between query and
candidate groups, allowing groups to remain unmatched.
Since multiple groups may share the same functional role, we use Hungarian matching to find the one-to-one correspondence between role-compatible groups that maximizes the total region IoU.
Normalizing by $\max(J_q,J_c)$ penalizes unmatched groups rather than averaging
only over the matched subset:
\begin{equation}
S_{\text{role-IoU}}(q,c)
=
\frac{1}{\max(J_q,J_c)}
\max_{\pi\in\Pi_{q,c}}
\sum_{(i,j)\in\pi} M_{ij}.
\label{eq:role-score}
\end{equation}

\paragraph{Canvas aspect ratio.}
Let $a_d=w_d/h_d$ denote the aspect ratio of design $d$.
This term preserves canvas-geometry information that is removed when group
regions are normalized independently by canvas size.
We define
\begin{equation}
S_{\text{canvas}}(q,c)
=
\frac{\min(a_q,a_c)}{\max(a_q,a_c)}.
\label{eq:canvas-score}
\end{equation}

\paragraph{Word and image count similarity.}
Let $W_d$ denote the total number of words across all text elements and $I_d$
the number of image elements.
This term compares the amount of textual and visual content between the query
and candidate:
\begin{equation}
S_{\text{word/image}}(q,c)
=
\frac{\min(W_q,W_c)+\min(I_q,I_c)}
     {\max(W_q+I_q,W_c+I_c)}.
\label{eq:count-score}
\end{equation}
When $W_q+I_q=W_c+I_c=0$, we define
$S_{\text{word/image}}(q,c)=0$.

\paragraph{Weights and reference selection.}
The weights in Eq.~\ref{eq:retrieval_score} are fixed to
$(\lambda_{\text{role-IoU}},\lambda_{\text{canvas}},\lambda_{\text{word/image}})
=(0.10,0.45,0.45)$ and are used consistently across all experiments.
Thus, the retrieval similarity is
\begin{equation}
\operatorname{sim}(q,c)
=
0.10\,S_{\text{role-IoU}}(q,c)
+
0.45\,S_{\text{canvas}}(q,c)
+
0.45\,S_{\text{word/image}}(q,c).
\label{eq:retrieval-score-supp}
\end{equation}
To obtain the Top-5 references used in Sec.~4.2, we first rank candidates by $\operatorname{sim}(q,c)$. Since Crello contains visually and textually similar designs across its train and test splits, we apply a lexical deduplication filter to avoid retrieving near-duplicate references that could provide an overly direct template for the query. This ensures that the retrieved examples serve as structural priors rather than near-identical design matches.
For two designs $d$ and $d'$, we define their lexical overlap as
\begin{equation}
D(d,d')
=
\frac{\sum_t\min(m_d(t),m_{d'}(t))}
     {\max(W_d,W_{d'})},
\label{eq:text-dedup}
\end{equation}
where $m_d(t)$ denotes the number of occurrences of word $t$ across all text
elements in design $d$.
Repeated occurrences are counted, while word order and text-element boundaries
are ignored.
We define $D(d,d')=0$ when $W_d=W_{d'}=0$.

A candidate $c$ is excluded if $D(q,c)\geq0.7$ or if
$D(z,c)\geq0.7$ for any previously selected reference
$z\in\mathcal{Z}_q$.
The first five remaining candidates form $\mathcal{Z}_q$.

\subsubsection{Retriever Behavior Analysis}
\begin{figure}[H]
    \centering
    \includegraphics[width=0.9\linewidth]{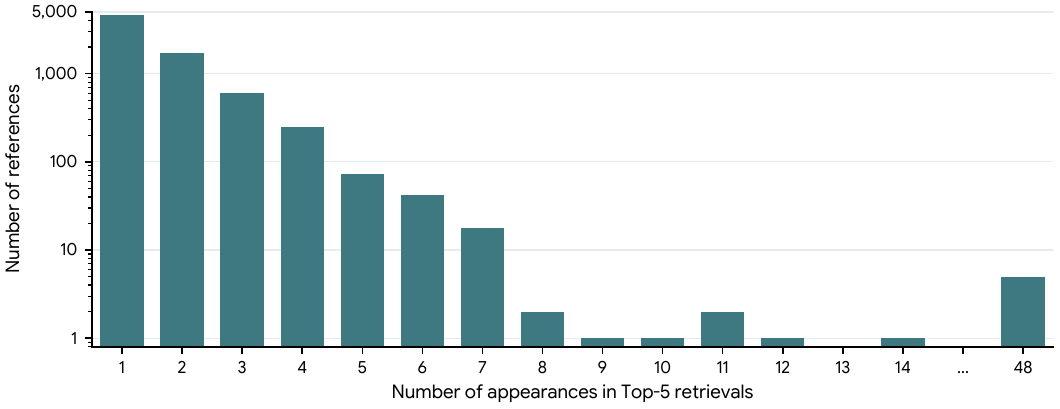}
    \vspace{-10pt}
    \caption{
    \textbf{Distribution of reference reuse across test queries.}
    For each reference appearing in the Top-5 retrieval results, we count how many times it is selected across all test queries. The histogram shows the distribution of these selection frequencies across references, with the vertical axis shown on a logarithmic scale.
    }
    \label{fig:retrieval_selection_frequency}
    \vspace{-5pt}
\end{figure}

We analyze how often individual references are reused across test queries to examine whether retrieval collapses to a small set of frequently selected designs.
For each reference appearing in the Top-5 retrieval results, we count the number of test queries for which it is selected.
As shown in Fig.~\ref{fig:retrieval_selection_frequency}, most references are selected for only one or a few queries, and the number of references decreases rapidly as the selection frequency increases.
This indicates that the Retriever produces diverse reference choices across queries rather than repeatedly relying on a small set of references.

Only a few references are selected frequently.
The most frequently selected reference appears in 48 queries, which mainly correspond to designs containing a single image spanning the canvas.
Since these queries share nearly identical coarse layout structures, repeatedly retrieving the same reference is expected.
Overall, the distribution shows that the Retriever maintains diverse reference selection while consistently reusing similar references for structurally similar queries.

\subsubsection{Qualitative Retrieval Results}
\begin{figure}[H]
    \centering
    \includegraphics[width=0.95\linewidth]{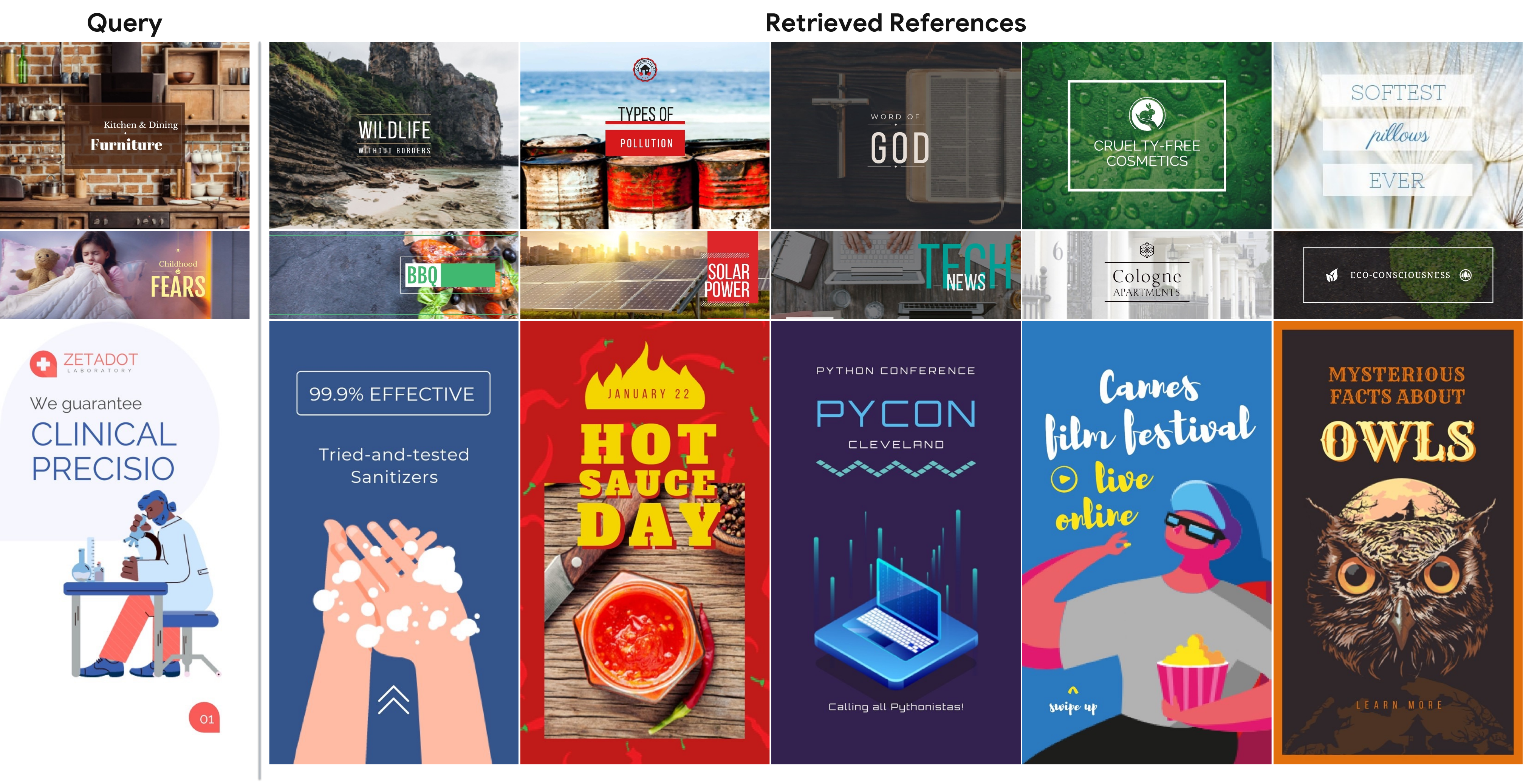}
    \vspace{-10pt}
    \caption{\textbf{Qualitative examples of retrieved references.} For each query, we show the five highest-ranked references retrieved from the reference pool. The retrieved designs exhibit similar coarse layout structures and canvas geometry.}
    \label{fig:retrieval_qualitative_examples}
    \vspace{-5pt}
\end{figure}

We visualize representative retrieval results in
Fig.~\ref{fig:retrieval_qualitative_examples}.
Across diverse queries, the retrieved references show similar coarse layout structures to the query, despite differences in visual content and style.
Their canvas aspect ratios also closely match those of the corresponding
queries, consistent with the canvas-similarity term in our retrieval score.
These examples illustrate the structural references provided to the subsequent
planning stage.

\begin{figure}[H]
    \centering
    \includegraphics[width=0.95\linewidth]{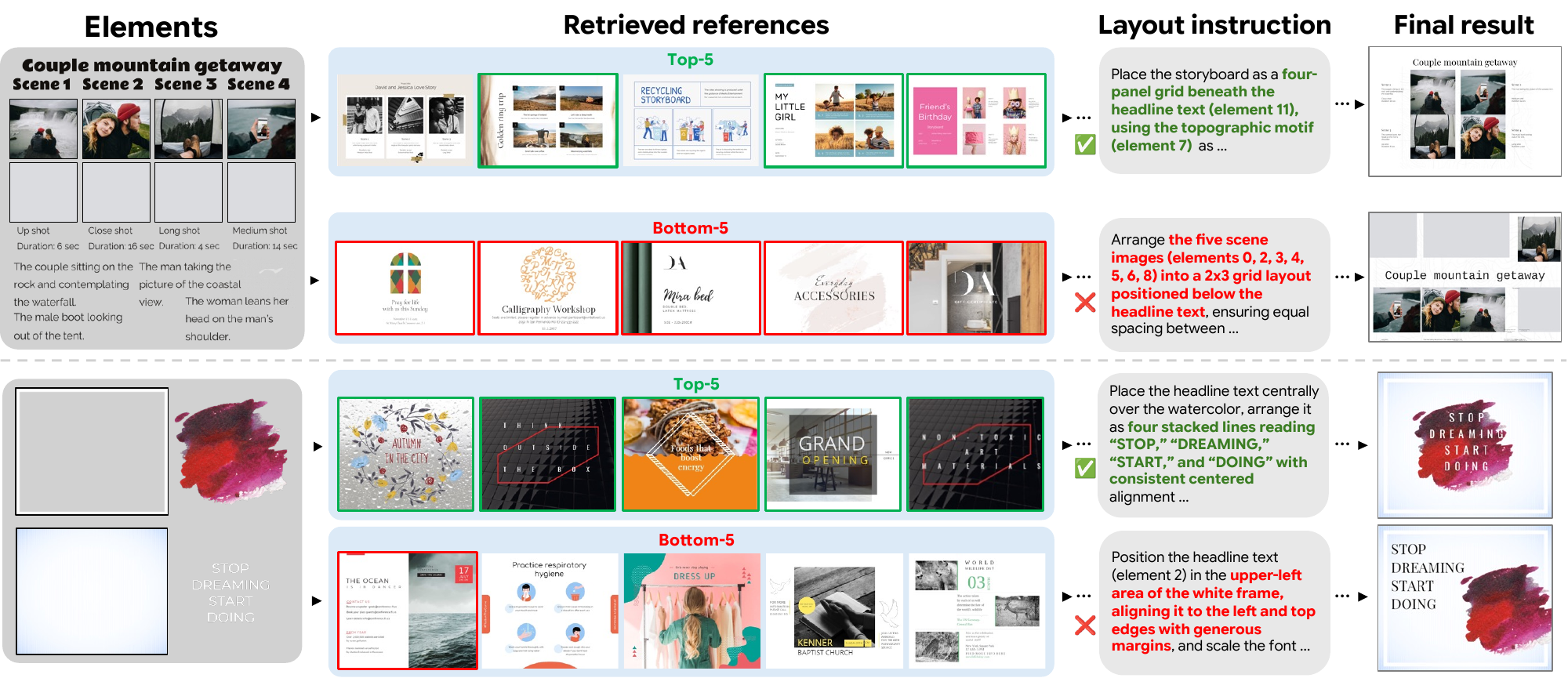}
    \vspace{-5pt}
    \caption{\textbf{Qualitative comparison of retrieval ranking.} For the same input elements, we compare the Top-5 references with the lowest-ranked five references among candidates sharing the query canvas size, together with the resulting layout instructions and final layouts.}
    \label{fig:retrieval_ranking_comparison}
    \vspace{-10pt}
\end{figure}

We further examine how retrieval ranking affects downstream generation in
Fig.~\ref{fig:retrieval_ranking_comparison}.
Consistent with the quantitative ablation in
Table~\ref{tab:abl_module}, where Top-5 references outperform low-ranked
references, the qualitative results show that reference quality also changes
the resulting planning behavior and final composition.
For visualization, we select the Bottom-5 as the lowest-ranked references among
candidates with the same canvas size as the query, so that the comparison is
not dominated by differences in canvas geometry.
Top-ranked references provide structural cues that better match the inferred
query structure, leading to more suitable layout instructions and final layouts,
whereas low-ranked references provide less relevant guidance and result in
noticeably different compositions.

\subsection{Training Configurations}
\label{sec:training_config}
\begin{table}[H]
\centering
\caption{
\textbf{Training configurations.}
Configurations used to fine-tune the Structure-Aware Retriever, Planner, and Layout Generator.
}
\label{tab:training_config}

\small
\renewcommand{\arraystretch}{1.08}
\setlength{\tabcolsep}{3pt}

\begin{tabularx}{\linewidth}{
    @{}
    >{\raggedright\arraybackslash}p{0.27\linewidth}
    YYY
    @{}
}
\toprule

\textbf{Configuration}
& \textbf{Structure-Aware Retriever}
& \textbf{Progressive Planner}
& \textbf{Layout Generator} \\
\midrule

\rowcolor{gray!12}
\multicolumn{4}{@{}l@{}}{\textbf{Model and adaptation}} \\

Base model
& Qwen3-VL-8B-Instruct
& Qwen3-VL-8B-Instruct
& Qwen3-VL-8B-Instruct \\

LoRA rank / $\alpha$ / dropout
& 32 / 64 / 0.05
& 32 / 64 / 0.05
& 32 / 64 / 0.05 \\

LoRA target
& LLM
& LLM
& LLM \\

Trainable non-LoRA module
& Merger
& Merger
& Merger \\

\addlinespace[2pt]
\rowcolor{gray!12}
\multicolumn{4}{@{}l@{}}{\textbf{Optimization}} \\

Epochs
& 5 & 6 & 10 \\

Global batch size
& 32 & 64 & 64 \\

Per-device batch size
& 1 & 1 & 1 \\

Gradient accumulation
& 8 & 16 & 16 \\

LoRA learning rate
& $1 \times 10^{-4}$
& $1 \times 10^{-4}$
& $1 \times 10^{-4}$ \\

Merger learning rate
& $1 \times 10^{-5}$
& $1 \times 10^{-5}$
& $1 \times 10^{-5}$ \\

Weight decay
& 0.1 & 0.1 & 0.1 \\

Warmup ratio
& 0.03 & 0.03 & 0.03 \\

LR scheduler
& Cosine
& Cosine
& Cosine \\

\addlinespace[2pt]
\rowcolor{gray!12}
\multicolumn{4}{@{}l@{}}{\textbf{Visual input resolution}} \\

Element image min. pixels
& $192 \times 32 \times 32$
& $16 \times 32 \times 32$
& $16 \times 32 \times 32$ \\

Element image max. pixels
& $384 \times 32 \times 32$
& $784 \times 32 \times 32$
& $784 \times 32 \times 32$ \\

Canvas max. pixels
& --
& $1536 \times 32 \times 32$
& $1536 \times 32 \times 32$ \\

Reference max. pixels
& --
& $512 \times 32 \times 32$
& $512 \times 32 \times 32$ \\

\addlinespace[2pt]
\rowcolor{gray!12}
\multicolumn{4}{@{}l@{}}{\textbf{Training system}} \\

Distributed training
& DDP
& DeepSpeed ZeRO-3
& DeepSpeed ZeRO-3 \\

Precision
& BF16 + TF32
& BF16 + TF32
& BF16 + TF32 \\

Gradient checkpointing
& Yes
& Yes
& Yes \\

FlashAttention-2 / Liger
& Yes / Yes
& Yes / Yes
& Yes / Yes \\

\bottomrule
\end{tabularx}
\end{table}
We report the detailed training configurations for the Structure-Aware Retriever, Progressive Planner, and Layout Generator in Table~\ref{tab:training_config}. All three modules are initialized from Qwen3-VL-8B-Instruct and fine-tuned with LoRA, with module-specific optimization and visual input settings.

\subsection{Prompts for Each Module}
\begin{vlmpromptbox_retriever}
    You are the coarse-layout Retrieval Agent for a graphic-design generation pipeline.

\medskip
\textbf{[DESIGN REQUEST]}

Title: \{title\}

Intention: \{intention\}

\medskip
\textbf{[CANVAS]}

Width: \{canvas\_width px\}

Height: \{canvas\_height px\}

\medskip
\textbf{[SUPPLIED ELEMENTS]}
\{element\_lines\}

\medskip
\textbf{[TASK]}

Infer the small set of functional groups that the supplied elements are likely to form in the complete design, and estimate one coarse canvas region for each group. This is not the final design plan. The output is used only to retrieve structurally similar training examples.

Use the request, every supplied element, image content, text content, and canvas aspect ratio. Think globally about likely background, focal visual, headline, information, CTA, branding, footer, structural container, and decoration regions. Regions may overlap when intended design layers overlap, and a canvas-filling background may use the full canvas.

\medskip
\textbf{[CLOSED-SET GROUP ROLES]}

Choose each group role from:

- \texttt{background block}

- \texttt{visual/hero block}

- \texttt{title/headline block}

- \texttt{content block}

- \texttt{support block}

- \texttt{decorative block}

[\ldots\ detailed role definitions omitted \ldots]

\medskip
\textbf{[NON-NEGOTIABLE RETRIEVAL CONTRACT]}

- Predict structure from the supplied raw inputs only.

- Select each group's primary design function. Roles may repeat and not all roles must appear.

- Explain the semantic or structural reason for each grouping in one non-empty \texttt{grouping\_reason}.

- Return exactly one syntactically valid JSON object and nothing else.

- Before responding, verify that group IDs are contiguous, all four region values are numeric pixel values, and every region stays inside the canvas.

[\ldots\ additional validation details omitted \ldots]

\medskip
\textbf{[OUTPUT SCHEMA]}

\{

\ \ ``groups'': [

\ \ \ \ \{

\ \ \ \ \ \ ``group\_id'': 0,

\ \ \ \ \ \ ``role'': ``role label'',

\ \ \ \ \ \ ``grouping\_reason'': ``why these elements form one functional group'',

\ \ \ \ \ \ ``group\_canvas\_region'': \{

\ \ \ \ \ \ \ \ ``left'': 0, ``top'': 0, ``width'': 0, ``height'': 0

\ \ \ \ \ \ \}

\ \ \ \ \}

\ \ ]

\}

\end{vlmpromptbox_retriever}

\begin{vlmpromptbox_planner}
    You are an expert sequential graphic-layout planner.

\medskip
\textbf{[DOCUMENT TITLE]}

\{title\}

\textbf{[DOCUMENT INTENTION]}

\{intention\}

\textbf{[CANVAS]}

width: \{canvas\_width\}

height: \{canvas\_height\}

\medskip
\textbf{[COMPLETE IMMUTABLE ELEMENT INVENTORY]}

Every element has an immutable integer ID. Image elements are represented by multimodal asset slots.

\{element\_lines\}

Assign every listed element exactly once across the conversation. Never omit, duplicate, invent, merge, or silently repair an element. Asset images are semantic inputs, not completed-layout evidence.

\medskip
\textbf{[REFERENCE DESIGNS]}

\{retrieved\_examples\}

The five images above are ordered by deterministic visual-structure similarity. Use them only as broad composition priors. Do not describe or identify them, copy hidden annotations, or output reference metadata.

\medskip
\textbf{[TASK]}

Infer and return exactly one next dependency-ready group, not a plan for the complete design. The element inventory is the only source of target group membership and role; the title, intention, canvas, and reference designs provide context but do not reveal the target grouping.

Choose the coarsest dependency-consistent unassigned group that forms one coherent unit and can be planned and rendered in one turn. Avoid both element-level fragmentation and over-merging across independently planned units.

Conversation order is the predicted inter-group execution and rendering order. Follow a dependency-aware human design workflow: place a supplied canvas-wide foundation first when present, then stable visual, container, or typographic anchors before groups positioned relative to them. Among dependency-ready groups, prefer the group whose placement is most stable and least likely to cause downstream rework or unnecessary overlap or occlusion.

[\ldots\ additional grouping and execution constraints omitted \ldots]

\medskip
\textbf{[RATIONALE GUIDANCE]}

- \texttt{why\_this\_group}: explain why the selected elements form one coherent, jointly placeable semantic unit.

- \texttt{why\_now}: explain why this group should be rendered next based on its dependencies, available rendered anchors, anchoring value, placement stability, and reduced downstream rework or unnecessary occlusion.

- \texttt{why\_this\_layout}: justify the qualitative layout from the selected elements, canvas, current conversation state, and broad reference priors, relating it to hierarchy, reading flow, readability, contrast, balance, whitespace, or occlusion avoidance where relevant.

Each rationale field should normally be one sentence and must not exceed two sentences. Do not offer alternatives or describe the selected group as already rendered.

\medskip
\textbf{[LAYOUT INSTRUCTION]}

\texttt{layout\_instruction} must specify one decisive, executable action that adds only the selected group's newly assigned elements. Earlier elements may be referenced as fixed alignment, containment, spacing, contrast, depth, or overlap anchors, but must not be reassigned or modified.

Where relevant, describe qualitative region, alignment, anchoring, reading order, relative scale, spacing, whitespace, overlap, containment, crop or fit behavior, depth, typography, color, and contrast-sensitive placement. Never output bounding boxes, exact coordinates, normalized coordinates, pixel measurements, or position tokens.

Mention every selected element at least once using a descriptive reference such as \texttt{headline text (element 7)}. Every referenced element ID must belong either to the selected group or to a group from an earlier assistant response.

[\ldots]

\medskip
\textbf{[CLOSED-SET GROUP ROLES]}

Choose exactly one role from:

- \texttt{background block}

- \texttt{visual/hero block}

- \texttt{title/headline block}

- \texttt{content block}

- \texttt{support block}

- \texttt{decorative block}

[\ldots\ detailed role definitions omitted \ldots]

\medskip
\textbf{[STRICT OUTPUT CONTRACT]}

Return exactly one valid JSON object and nothing else. The top-level object must contain exactly \texttt{turn\_plan}. \texttt{turn\_plan} must contain exactly \texttt{group\_id}, \texttt{role}, \texttt{elements}, \texttt{rationale}, and \texttt{layout\_instruction}. \texttt{rationale} must contain exactly \texttt{why\_this\_group}, \texttt{why\_now}, and \texttt{why\_this\_layout}.

[\ldots\ additional output-format checks omitted \ldots]

\medskip
\textbf{[SUBSEQUENT TURNS]}

Continue the same sequential planning conversation. All prior assistant turn plans are validated and immutable.

\textbf{[CURRENT CANVAS AFTER PREVIOUS TURNS]}

\{current\_canvas\_image\}

The image above is the cumulative rendered state after all prior turn plans. Treat all existing placements as fixed and plan only the next group relative to them.

\textbf{[REMAINING IMMUTABLE ELEMENT INVENTORY]}

\{remaining\_element\_lines\}

Select exactly one dependency-ready group using only remaining elements. Do not reuse, modify, or reassign any previously planned element, and do not expose any later group.

Return exactly one valid JSON object following the same \texttt{turn\_plan} schema established in the initial instruction.

\end{vlmpromptbox_planner}

\begin{vlmpromptbox_generator}
    You are the group-wise layout Predict Agent for a complete graphic design.

\medskip
\textbf{[CANVAS]}

Width: \{canvas\_width px\}

Height: \{canvas\_height px\}

\medskip
\textbf{[ALL SUPPLIED ELEMENTS]}

\{all\_element\_lines\}

\medskip
\textbf{[RETRIEVED REFERENCE IMAGES]}

The following accepted Top-5 reference images are provided in retrieval ranking order. Use their visual composition and layout as references.

\{retrieved\_examples\}

\medskip
\textbf{[TASK]}

Lay out ONLY the elements listed in the current group's target elements. Treat the supplied current group as the only available Planner decision. Follow its role and \texttt{layout\_instruction}, and infer the current elements' exact placement from the canvas size, complete element inventory, supplied assets, and retrieved reference images. Do not place or predict any non-current element.

\textbf{[CURRENT GROUP]}

\{current\_group\}

Now predict the current group elements: 
\{target\_group\_elements\}

\medskip
\textbf{[SUBSEQUENT TURNS]}

current canvas state: \{current\_canvas\_state\}

current group:

\{current\_group\}

Now predict the current group elements: \{target\_group\_elements\}

\end{vlmpromptbox_generator}

\begin{vlmpromptbox_verifier}
    You are a strict graphic-layout referee.

One layout group has just been placed on a design canvas. You will see several candidate canvases that differ only in how the current group was predicted.

Your task is to (1) select the best rendered candidate, (2) diagnose any remaining problem using prediction metadata, and (3) choose whether to accept or retry with a better plan.

\medskip
\textbf{[ITERATION STATE]}

Current judge call: \{current\_iteration of max\_iterations\}

Actions returned by earlier judge calls:

\{tool\_history\}

Judge the candidates actually shown; do not assume a prior correction succeeded.

Number of candidates in this call: \{n\_candidates\}

\medskip
\textbf{[CANVAS]}

width: \{canvas\_width\}

height: \{canvas\_height\}

\medskip
\textbf{[CURRENT GROUP]}

\{current\_group\_json\}

\textbf{[ELEMENTS BEFORE THIS GROUP]}

\{previous\_element\_lines\}

\textbf{[ELEMENTS IN THIS GROUP]}

\{current\_group\_element\_lines\}

\textbf{[CANVAS BEFORE THIS TURN]}

<image>

\medskip
\textbf{[CANDIDATES]}

\textbf{[CANDIDATE 0]}

<image>

\textbf{[PREDICTION --- CANDIDATE 0]}

\{candidate\_0\_prediction\}

[\ldots\ candidates 1--3 omitted \ldots]

\textbf{[CANDIDATE 4]}

<image>

\textbf{[PREDICTION --- CANDIDATE 4]}

\{candidate\_4\_prediction\}

\medskip
\textbf{[STEP 1 --- SELECT THE BEST CANDIDATE]}

Rank candidates only by their rendered images, not prediction metadata. Choose one using this priority: (1) preserve readable text, important subjects, and the existing canvas's meaning; (2) follow the current \texttt{layout\_instruction}; (3) achieve strong readability and composition.

Penalize overlap only when it hides important content or breaks an element's role. Allow intentional layering of backgrounds, decoration, framing, and depth. If tied, choose the safer candidate, then the lower index. Give one short reason.

\medskip
\textbf{[STEP 2 --- DIAGNOSE THE SELECTED CANDIDATE]}

After selecting, use its prediction metadata and the current plan only to diagnose visible problems. Distinguish harmful occlusion from intentional layering. Earlier-group elements are fixed; resolve conflicts by revising current-group placement, scale, style, or internal stacking.

\medskip
\textbf{[STEP 3 --- CHOOSE ONE ACTION]}

Choose exactly one:

- \texttt{accept}

- \texttt{retry\_with\_feedback}

Use \texttt{accept} when the selected candidate is already good enough to commit.

Use \texttt{retry\_with\_feedback} when the selected candidate can be improved by giving the Predictor a revised \texttt{layout\_instruction}. Return only a rewritten layout instruction; group ID, role, elements, and rationale are reused. The Predictor regenerates only the current group.

Use relational visual language, not coordinates, bbox values, x/y positions, or exact numeric dimensions. Do not ask the Predictor to reorder elements placed in earlier groups.

[\ldots\ additional retry guidance omitted \ldots]

\medskip
\textbf{[OUTPUT]}

For \texttt{accept}:

\{

\ \ ``choice'': <candidate index>,

\ \ ``reason'': ``<one short sentence>'',

\ \ ``action'': ``accept''

\}

For \texttt{retry\_with\_feedback}:

\{

\ \ ``choice'': <candidate index>,

\ \ ``reason'': ``<one short sentence>'',

\ \ ``action'': ``retry\_with\_feedback'',

\ \ ``retry\_reason'': ``<what the Predictor failed to execute>'',

\ \ ``layout\_instruction'': ``<the revised natural-language plan for a better result>''

\}

\medskip
\textbf{[FINAL SELECTION AFTER MAXIMUM ROUNDS]}

If no verification round is accepted, make the final selection from the per-round finalists.

\textbf{[CANVAS]}

width: \{canvas\_width\}

height: \{canvas\_height\}

\{previous\_canvas\_block\}

\textbf{[CURRENT GROUP]}

\{current\_group\_json\}

\textbf{[ELEMENTS IN THE CURRENT GROUP]}

\{target\_group\_elements\}

\textbf{[FINAL CANDIDATES]}

\{candidate\_block\}

Choose the single best final candidate. Select the candidate that follows the current group's \texttt{layout\_instruction} most faithfully while producing the best overall visual result. Judge only the candidates shown; you must choose one candidate.

Return exactly one valid JSON object:

\{

\ \ ``choice'': <candidate index from 0 to \{max\_candidate\}>,

\ \ ``reason'': ``<one short sentence>'',

\ \ ``action'': ``accept''

\}

\end{vlmpromptbox_verifier}

\section{Experimental Details}
\label{sec:experimental_details}

\subsection{Evaluation Setup}
\label{sec:evaluation_setup}
\paragraph{Input configuration.}
We adapt each baseline to its native input formulation while keeping the
underlying Crello elements consistent whenever possible.
FlexDM, LaDeCo, and \ours receive the original image and text elements as
separate editable inputs.
PosterCopilot represents all elements as images; therefore, we first render each
text element using its ground-truth text attributes and provide the resulting
text renders as image elements together with the original visual elements.
VFLM is specialized for text placement on a fixed background, so we render all
ground-truth non-text elements to construct the background canvas and provide
only the text elements for layout prediction.
These settings follow the input assumptions of each method while using the
same source design content.

\paragraph{Dataset-specific preprocessing.}
LICA~\citep{hirsch2026licalayeredimagecomposition} does not provide document titles, which are required as metadata inputs to our pipeline. We therefore generate a title for each LICA sample using Qwen3.5-27B~\citep{qwen3.5}. We use the ground-truth rendered image together with the available LICA metadata, including the description, aesthetics, tags, and user intent, to generate a short document title. The generated title is used only as auxiliary metadata during LICA inference.

\subsection{Evaluation Metrics}
\label{sec:metrics}
Our quantitative evaluation uses both VLM-based and rule-based metrics.
The VLM-based metrics assess perceptual design quality and alignment with the
design intention from completed renders, while the rule-based metrics measure
structural and spatial properties of the generated layouts.

\subsubsection{VLM-based Metrics}
\label{sec:vlm_metrics}
\paragraph{Judge and evaluation setup.}
We use GPT-5~\citep{openai2025gpt5}
(\texttt{gpt-5-2025-08-07}) as the VLM judge.
For each design, the judge receives the final rendered image and the
corresponding design intention, without access to source elements,
ground-truth layouts, intermediate renders, or method identity.
For Crello-v4, we use the intention annotations provided by
OpenCOLE~\citep{inoue2024opencolereproducibleautomaticgraphic}, while LICA uses
its original \texttt{user\_intent} annotations.
The judge independently assigns an integer score from 1 to 5 to each of the six
criteria using the fixed evaluation prompt and structured output format.

\paragraph{Evaluation criteria.}
Each criterion defines anchors for integer scores from 1 to 5, and the judge scores the six criteria independently. Criteria (i) through (v) focus on the perceptual quality of the visible design, while criterion (vi) additionally evaluates consistency with the provided design intention. We define the six evaluation criteria as follows:

\begin{itemize}
    \item \emph{(i) Layout composition rationality} covers balance, alignment, grouping, spacing, whitespace, visual flow, and density at the level of the full design. Overlap and alignment enter this score only when they affect the visible composition.

    \item \emph{(ii) Typography legibility and hierarchy} covers readability at the presented scale, typographic choices, contrast between text and the background, spacing, line breaks, reading order, hierarchy, clipping, overlap, and broken glyphs. It does not check exact wording or spelling.

    \item \emph{(iii) Visible element presentation} covers framing, cropping, scale, clarity, visibility, and compositing of non-text imagery, graphics, icons, shapes, and decorations. It does not assess the artistic merit of an element or its fidelity to an unseen source.

    \item \emph{(iv) Color harmony and contrast} covers palette relationships, color balance, and separation between foreground and background. If insufficient contrast makes text difficult to read, the resulting loss of legibility is counted under typography and is not counted again here.

    \item \emph{(v) Visual style coherence} checks whether typography, imagery, decorations, backgrounds, and effects form a consistent visual language. It excludes color harmony, thematic relevance, originality, and personal preference.

    \item \emph{(vi) Design intention alignment} compares the intention with evidence visible in the render. It assesses whether the design realizes the stated purpose, message, format, and salient content or style cues, while excluding source preservation, requirements that cannot be observed in the render, similarity to the ground truth, general visual polish, and character-level text transcription.
\end{itemize}

\paragraph{Coverage and aggregation.}
VFLM~\citep{guo2026seeingimprovingvisualfeedback} performs inference only on designs containing text; therefore, text-free
samples are excluded from all VFLM aggregates.
For the typography criterion, we exclude samples without text for all
methods, while retaining those samples for the other five criteria when
applicable.
Whether a sample contains text is determined from the source-layout metadata,
regardless of text visibility in the generated render.
For each criterion, we report the arithmetic mean and sample standard deviation
across evaluated designs.
\paragraph{Full evaluation prompt.}
The exact prompt used for VLM evaluation is provided below. \texttt{\{FINAL\_RENDER\_IMAGE\}} marks the position of the final render image, while \texttt{\{DESIGN\_INTENTION\}} is replaced by the design intention for each sample. The structured-output schema is specified separately through the API and is therefore not included in the prompt below.
\begin{vlmpromptbox_vlm_eval}
    You are a rigorous evaluator of completed poster and graphic-layout renders.

\par\medskip
Evaluate the final render using the six criteria below. For criteria 1--5,
judge the image as if the design intention were unavailable. Do not reward,
penalize, or reinterpret visible execution based on whether it matches the
intention. Use the design intention only for
design\_intention\_alignment.

\par\medskip
Score the six criteria independently with integers from 1 to 5, using only
their criterion-specific anchors. Do not let one overall impression determine
every score. Count the same visible defect under multiple criteria only when it
causes a distinct failure under each criterion's definition.

\par\medskip
\textbf{General rules:}

\par
- Judge at the presented viewing scale. Do not assume content becomes legible
or clean when zoomed in.

\par
- Do not penalize deliberate whitespace, overlap, asymmetry, cropping, or
stylistic contrast unless it visibly harms function or coherence.

\par
- Do not equate visual complexity, decoration, or realism with quality.

\par
- Score every criterion. Never return null, N/A, not\_evaluable, or an omitted
field.

\par\medskip
\textbf{SCORING CRITERIA}

\par
1. layout\_composition\_rationality

\par
Evaluate macro-level spatial organization: balance, alignment, grouping,
spacing, whitespace, distribution of visual weight, regional flow, and density
or crowding. Do not score text legibility, color harmony, style coherence, or
pixel-level artifacts here. Judge overlap and alignment by their visible
compositional impact, not as raw counts.

\par
- 5: Spatial organization is deliberate and balanced, with clear grouping and
flow and no meaningful awkward gap or crowding.

\par
- 4: Organization is sound, with only minor local imbalance, spacing,
alignment, or grouping issues.

\par
- 3: The layout remains functional, but noticeable imbalance, awkward gaps,
weak grouping, or crowding requires revision.

\par
- 2: Major imbalance, fragmentation, or crowding disrupts navigation and
substantially weakens the layout.

\par
- 1: Spatial organization is chaotic or effectively absent, with no usable
grouping or flow.

\par\medskip
2. typography\_legibility\_and\_hierarchy

\par
Evaluate visible typography as a communication system: legibility at the
presented scale, apparent font size and choice, text-background contrast, line
breaking, letter and line spacing, reading order, hierarchy, clipping, overlap,
and visibly broken or blurred glyphs. Judge apparent typographic execution, not
exact wording or spelling.

\par
- 5: All visually important text is readily legible, and typographic levels
create an immediate, effective reading order.

\par
- 4: Text and hierarchy are clear overall, with only minor issues in secondary
text, spacing, contrast, or emphasis.

\par
- 3: The main text remains usable, but noticeable legibility, line-break,
spacing, contrast, or hierarchy issues require revision.

\par
- 2: Important visible text is difficult to read or the typographic hierarchy
is substantially confusing.

\par
- 1: The visible text system is functionally unusable, or no functional visible
typography is present.

\par\medskip
3. visible\_element\_presentation

\par
Evaluate the visible presentation of non-text imagery, graphics,
illustrations, icons, shapes, and decorations: framing, harmful crop,
off-canvas placement, relative scale, distortion, blur or pixelation,
occlusion, focal-subject visibility, collisions, seams, halos, transparency,
and compositing integration. Judge visible presentation and integration, not
subject matter or intrinsic artistic merit. Text-specific defects belong to
typography\_legibility\_and\_hierarchy.

\par
- 5: Visible non-text elements are cleanly framed, identifiable, appropriately
scaled, and integrated, with no meaningful presentation or compositing problem.

\par
- 4: Element presentation is strong, with only minor localized framing, scale,
clarity, or integration issues.

\par
- 3: Most visible elements remain usable, but noticeable crop, occlusion,
distortion, collision, clarity, or integration problems require revision.

\par
- 2: Major presentation problems substantially obstruct or weaken one or more
visually prominent elements.

\par
- 1: Prominent visual elements are severely cropped, obscured, distorted,
degraded, off-canvas, or colliding to the point of functional failure.

\par\medskip
4. color\_harmony\_and\_contrast

\par
Evaluate the visible color system: palette harmony, hue, saturation and
brightness relationships, foreground-background separation, distribution and
balance of color, and unwanted clashes. Judge color as an internally
functioning system, not theme appropriateness or personal taste. When contrast
specifically makes text hard to read, score that functional consequence under
typography\_legibility\_and\_hierarchy rather than duplicating the penalty here.

\par
- 5: The palette is harmonious and controlled, with effective visual
separation and no meaningful clash or imbalance.

\par
- 4: Color relationships are strong, with only minor localized imbalance,
dullness, excess intensity, or weak separation.

\par
- 3: The palette remains usable, but noticeable clashes, muddy relationships,
uneven distribution, or weak contrast require revision.

\par
- 2: Major color conflicts or contrast failures substantially weaken visual
organization and coherence.

\par
- 1: The color system is severely dysfunctional, with pervasive clashes,
muddiness, or loss of visual separation.

\par\medskip
5. visual\_style\_coherence

\par
Evaluate whether the visible typeface personality, imagery treatment,
illustration or icon language, decorations, background, and effects form a
deliberate, mutually consistent visual system. Judge cross-element consistency
and integration, not personal taste, theme appropriateness, originality, color
harmony itself, or the intrinsic beauty of individual assets.

\par
- 5: All visible treatments form a polished, coherent visual language with no
meaningful stylistic conflict.

\par
- 4: The visual language is strong, with only minor localized inconsistency or
unresolved styling.

\par
- 3: Styling is workable, but noticeable conflicts or inconsistencies make the
design feel partly unresolved.

\par
- 2: Major treatment conflicts substantially weaken stylistic coherence.

\par
- 1: The design lacks a usable visual language or contains severe stylistic
conflicts that make it feel visibly broken.

\par\medskip
6. design\_intention\_alignment

\par
Evaluate how well the final render visibly realizes the supplied design
intention: its main purpose or topic, intended message, format or use, and
salient explicit theme, mood, style, or content cues. Judge alignment jointly
from the intention and visible render. Do not assess source-asset preservation,
unseen requirements, exact ground-truth similarity, general visual polish, or
character-perfect text transcription here.

\par
- 5: The main purpose and message are immediately recognizable; all salient
explicit cues are visibly fulfilled with no meaningful contradiction.

\par
- 4: The intention is clearly realized and most salient cues are fulfilled,
with only a minor omission or weakness.

\par
- 3: The broad topic or purpose is recognizable, but important cues are only
partially realized or remain ambiguous.

\par
- 2: The render has only a limited connection to the intention; major cues are
absent or contradicted and the intended message is hard to recover.

\par
- 1: The render is unrelated to or strongly contradicts the intention; the
intended purpose is not recognizable.

\par\medskip
For each criterion, provide one integer score and one English evidence sentence
of no more than 30 words. State only the most decisive evidence; do not restate
the criterion, anchor, or score.

\par\medskip
For criteria 1--5, evidence must cite only visible image properties and must not
mention or rely on the design intention. For criterion 6, evidence must connect
a salient intention cue to its visible fulfillment or failure.

\par\medskip
Return only the schema-constrained result. Do not provide a total, average,
overall score, failure tag, or additional commentary.

\par\bigskip
\{FINAL\_RENDER\_IMAGE\}

\par\bigskip
\textbf{DESIGN INTENTION} \textemdash{} USE ONLY FOR CRITERION 6

\par
\{DESIGN\_INTENTION\}

\end{vlmpromptbox_vlm_eval}

\subsubsection{Rule-based Metrics}
Following prior element-to-design works~\citep{cheng2024graphicdesignlargemultimodal,lin2024elementsdesignlayeredapproach,wei2025postercopilotlayoutreasoningcontrollable},
we report eight rule-based metrics covering structural properties
(Val, Ove, and Ali), agreement with the reference layout
(IoU, IOPR, and ARD), and text--background interactions
(T.Occ. and Rea.).
Let $\mathcal{E}=\{e_i\}_{i=1}^{N}$ denote the element inventory on a
canvas of size $W\times H$, and let $b_i=(x_i,y_i,w_i,h_i)$ denote
the predicted bounding box of element $e_i$.
An asterisk denotes the corresponding reference quantity.

Table~\ref{tab:main_quan} reports T.Occ. and Rea. alongside the
VLM-based metrics, while Table~\ref{tab:rule_based_quan} reports the
remaining six metrics.
We do not report these six metrics for VFLM because it places text
on a fixed ground-truth background rather than predicting the layout
of the complete element inventory.
IOPR is omitted for FlexDM and LaDeCo because their stacking orders
are not explicitly predicted.
Layout IoU and ARD are reported only on Crello, as reliable
image-element reference-box correspondence is unavailable in our
LICA evaluation setup.

\paragraph{Quantitative comparison.}
Table~\ref{tab:rule_based_quan} shows that \ours achieves the highest
layout IoU and lowest ARD on Crello among the compared methods,
indicating closer agreement with the reference layouts in element
geometry.
It also achieves the lowest IOPR on both datasets among methods
that predict stacking order.
Since rule-based metrics measure specific structural properties rather than overall design quality, we report them as complementary diagnostics alongside the VLM-based evaluation.

\begin{table}[h]
    \centering
    \setlength{\tabcolsep}{1pt}
    \caption{\textbf{Rule-based Quantitative Comparison.}
    We report the mean of six rule-based metrics, Val, Ove, Ali, IoU, IOPR, and ARD.
    A dash marks a metric that we do not report.
    \textbf{Bold} and \underline{underline} mark the best and the second best within each dataset, excluding GT.}
    \label{tab:rule_based_quan}
    \vspace{-9pt}
    \scriptsize
    \renewcommand{\arraystretch}{0.95}
    
    \begin{tabular}{@{}
        >{\centering\arraybackslash}m{0.07\textwidth}|
        >{\raggedright\arraybackslash\hspace{3pt}}m{0.115\textwidth}|
        *{6}{>{\centering\arraybackslash\tiny}m{0.12\textwidth}}}
        \toprule

        \multicolumn{1}{c|}{\multirow[c]{2}{*}{\textbf{Dataset}}}
        & \multicolumn{1}{c|}{\multirow[c]{2}{*}{\textbf{Method}}}
        & \multicolumn{6}{c}{\textbf{Rule-based metrics}} \\

        &
        & {\scriptsize\mbox{Val $\uparrow$}}
        & {\scriptsize\mbox{Ove $\downarrow$}}
        & {\scriptsize\mbox{Ali $\downarrow$}}
        & {\scriptsize\mbox{IoU $\uparrow$}}
        & {\scriptsize\mbox{IOPR $\downarrow$}}
        & {\scriptsize\mbox{ARD $\downarrow$}} \\
        % & {\scriptsize\mbox{T.Occ. $\downarrow$}}
        % & {\scriptsize\mbox{Rea. $\downarrow$}} \\

        \datasetseparator
        % \multirow[c]{6}{*}{\makecell{\textbf{Crello}\\\textbf{dataset}}}
        
        % & \multicolumn{7}{l}{\textcolor{gray}{\textit{Intention-to-Design Generation Models}}} \\
        % & OpenCOLE
        % & -- & -- & -- & -- & -- & -- \\
        % & 0.0034 & 0.0737 \\

        % \cmidrule(l){2-8}
        
        % & \multicolumn{7}{l}{\textcolor{gray}{\textit{Graphic Design Composition Models}}} \\
        & \metricrowstrut FlexDM
        & 0.8602 & 0.4699 & 0.0012 & \underline{0.2059} & --
        & 0.1245 \\
        % & 0.1519 & 0.0583 \\

        & \metricrowstrut LaDeCo
        & 0.9227 & \textbf{0.2026} & 0.0012 & 0.1297 & --
        & 0.1749 \\
        % & 0.0570 & 0.0432 \\

        & \metricrowstrut PosterCopilot
        & \textbf{0.9755} & 0.3323 & \textbf{0.0007} & 0.1759 & \underline{0.0871}
        & \underline{0.0191} \\
        % & 0.0369 & 0.0385 \\

        & \metricrowstrut VFLM
        & -- & -- & -- & -- & -- & --
        \\
        % & \textbf{0.0002} & \underline{0.0368} \\

        & \cellcolor{hlours}\metricrowstrut\textbf{\ours}
        & \cellcolor{hlours}\underline{0.9309}
        & \cellcolor{hlours}\underline{0.2164}
        & \cellcolor{hlours}\underline{0.0011}
        & \cellcolor{hlours}\textbf{0.2253}
        & \cellcolor{hlours}\textbf{0.0446}
        & \cellcolor{hlours}\textbf{0.0097} \\
        % & \cellcolor{hlours}\underline{0.0041}
        % & \cellcolor{hlours}\textbf{0.0345} \\

        \cline{2-8}
        \multirow[c]{-6}{*}[2.5pt]{\makecell{\textbf{Crello}\\\textbf{dataset}}}

        & \metricrowstrut GT
        & 0.9046 & 0.2107 & 0.0011 & 1.0000 & 0.0000
        & 0.0000 \\
        % & 0.0025 & 0.0235 \\
        % GT (before revision):
        % 0.9068 & 0.2213 & 0.0011 & 1.0000 & 0.0000 & 0.0000 & 0.0026 & 0.0260

        \datasetseparator
        % \multirow[c]{6}{*}{\makecell{\textbf{LICA}\\\textbf{dataset}}}
        
        % & \multicolumn{7}{l}{\textcolor{gray}{\textit{Intention-to-Design Generation Models}}} \\
        % & OpenCOLE
        % & -- & -- & -- & -- & -- & -- \\
        % & 0.0035 & 0.0537 \\

        % \cmidrule(l){2-8}

        % & \multicolumn{7}{l}{\textcolor{gray}{\textit{Graphic Design Composition Models}}} \\
        & \metricrowstrut FlexDM
        & \underline{0.9299} & 1.8654 & \underline{0.0006} & -- & --
        & -- \\
        % & 0.3212 & 0.0574 \\

        & \metricrowstrut LaDeCo
        & 0.8652 & \textbf{0.2471} & 0.0008 & -- & --
        & -- \\
        % & 0.0715 & 0.0393 \\

        & \metricrowstrut PosterCopilot
        & \textbf{0.9729} & 0.4514 & \textbf{0.0004} & -- & \underline{0.0926}
        & -- \\
        % & 0.1372 & 0.0281 \\

        & \metricrowstrut VFLM
        & -- & -- & -- & -- & -- & --
        \\
        % & 0.0001 & 0.0218 \\

        & \cellcolor{hlours}\metricrowstrut\textbf{\ours}
        & \cellcolor{hlours}0.8036 
        & \cellcolor{hlours}\underline{0.3715} 
        & \cellcolor{hlours}0.0008 
        & \cellcolor{hlours}--
        & \cellcolor{hlours}\textbf{0.0343} 
        & \cellcolor{hlours}-- \\
        % & \cellcolor{hlours}--
        % & \cellcolor{hlours}-- \\

        \cline{2-8}
        \multirow[c]{-6}{*}[2.5pt]{\makecell{\textbf{LICA}\\\textbf{dataset}}}

        & \metricrowstrut GT
        & 0.8815 & 0.2406 & 0.0006 & -- & 0.0000
        & -- \\
        % & 0.0040 & 0.0162 \\

        \specialrule{\heavyrulewidth}{0pt}{0pt}
    \end{tabular}

    % Numbers are the `average` field of outputs/<method>.json (Crello) and
    % outputs/lica_<method>.json (LICA), rounded to four decimals.
    % Aggregated sample counts (Crello): 2,371/2,375 for PosterCopilot,
    % 2,314/2,375 for VFLM, 2,375/2,375 for the rest.
    % (LICA): 1,010/1,011 for PosterCopilot, 953/1,011 for VFLM,
    % 1,011/1,011 for the rest.
    % IOPR additionally drops the 48 Crello designs whose ground truth has
    % fewer than two valid elements, so its denominator is 2,327 (2,323 for
    % PosterCopilot).
    % FlexDM and LaDeCo reuse the input stacking order rather than predicting
    % one, so their IOPR would be 0.0000 by construction and is reported as
    % not applicable instead.

\end{table}

\paragraph{Validity (Val $\uparrow$).}
Validity measures the proportion of elements whose predicted area
is at least a minimum fraction of the canvas area
\citep{hsu2023posterlayoutnewbenchmarkapproach}:
\begin{equation}
\mathrm{Val}
=
\frac{|\mathcal{E}_{\mathrm{val}}|}{N},
\qquad
\mathcal{E}_{\mathrm{val}}
=
\left\{
e_i \;\middle|\; w_i h_i \geq \tau W H
\right\},
\end{equation}
where $\tau=10^{-3}$.

\paragraph{Overlap (Ove $\downarrow$).}
Overlap measures the total pairwise bounding-box IoU among valid
elements, normalized by the total number of elements
\citep{hsu2023posterlayoutnewbenchmarkapproach}:
\begin{equation}
\mathrm{Ove}
=
\frac{1}{N}
\sum_{\substack{i<j\\ e_i,e_j\in\mathcal{E}_{\mathrm{val}}}}
\operatorname{IoU}(b_i,b_j).
\end{equation}
This geometric metric does not distinguish intentional layering
from harmful occlusion.

\paragraph{Alignment (Ali $\downarrow$).}
Alignment measures the consistency of element edges and centers.
We consider six canvas-normalized coordinates per element:
left, horizontal center, right, top, vertical center, and bottom.
Let $c_i^a$ denote coordinate $a$ of element $e_i$, and let
$\Delta_i^a=\min_{j\neq i}|c_i^a-c_j^a|$ denote its distance to
the closest corresponding coordinate of another element.
Following \citet{li2020attributeconditionedlayoutganautomatic}, we compute
\begin{equation}
\mathrm{Ali}
=
\frac{1}{N}
\sum_{i=1}^{N}
g\!\left(\min_a\Delta_i^a\right),
\qquad
g(\Delta)=-\log_{10}(1-\Delta).
\end{equation}
Lower scores indicate closer alignment along at least one of the
six coordinates.

\paragraph{Layout IoU (IoU $\uparrow$).}
Layout IoU measures bounding-box agreement between corresponding
predicted and reference elements
\citep{wei2025postercopilotlayoutreasoningcontrollable}.
Let $\mathcal{E}_{\mathrm{match}}$ denote the elements with an
available predicted--reference correspondence.
We compute
\begin{equation}
\mathrm{IoU}
=
\frac{1}{|\mathcal{E}_{\mathrm{match}}|}
\sum_{e_i\in\mathcal{E}_{\mathrm{match}}}
\operatorname{IoU}(b_i,b_i^{*}).
\end{equation}
The same matched subset is used for ARD.

\paragraph{Inverse Order Pair Ratio (IOPR $\downarrow$).}
IOPR measures ordering inversions between overlapping elements
relative to the reference design
\citep{cheng2024graphicdesignlargemultimodal,wei2025postercopilotlayoutreasoningcontrollable}.
Here, $\zeta_i$ denotes the final global stacking order, including
inter-group ordering, with larger values rendered in front.
We count pairs that overlap in the predicted layout and have
opposite stacking orders in the prediction and reference,
normalizing by the total number of element pairs:
\begin{equation}
\mathrm{IOPR}
=
\frac{2}{N(N-1)}
\sum_{i<j}
\mathbf{1}
\left[
\operatorname{IoU}(b_i,b_j)>0
\;\wedge\;
(\zeta_i-\zeta_j)(\zeta_i^{*}-\zeta_j^{*})<0
\right].
\end{equation}

\paragraph{Aspect Ratio Difference (ARD $\downarrow$).}
ARD measures the discrepancy between predicted and reference
aspect ratios independently of element position.
Following \citet{wei2025postercopilotlayoutreasoningcontrollable},
we use an angular aspect-ratio difference based on
CIoU \citep{zheng2020distance}:
\begin{equation}
\mathrm{ARD}
=
\frac{1}{|\mathcal{E}_{\mathrm{match}}|}
\sum_{e_i\in\mathcal{E}_{\mathrm{match}}}
\left(
\arctan\frac{w_i^{*}}{h_i^{*}}
-
\arctan\frac{w_i}{h_i}
\right)^2.
\end{equation}

\paragraph{Text Occlusion (T.Occ. $\downarrow$).}
Text occlusion measures the fraction of rendered glyph pixels
hidden by foreground elements.
Let $\mathcal{T}\subseteq\mathcal{E}$ denote the text elements.
For each $e_i\in\mathcal{T}$, let $A_i$ denote its glyph pixels
rendered using the predicted box and typographic attributes,
before foreground occlusion.
Let $O_i\subseteq A_i$ denote the pixels opaquely covered by
elements with a larger global stacking order.
We compute
\begin{equation}
\mathrm{T.Occ.}
=
\frac{1}{|\mathcal{T}|}
\sum_{e_i\in\mathcal{T}}
\frac{|O_i|}{|A_i|}.
\end{equation}

\paragraph{Text Unreadability (Rea $\downarrow$).}
Rea measures background complexity beneath text
\citep{hsu2023posterlayoutnewbenchmarkapproach}.
We first render all non-text elements in their predicted stacking
order to obtain a text-free background image $I$.
Let $G$ denote the Sobel gradient magnitude of its 8-bit grayscale
representation, and let $R$ denote the set of canvas pixels covered
by the union of predicted text boxes.
We compute
\begin{equation}
\mathrm{Rea}
=
\frac{1}{|R|}
\sum_{p\in R}
\frac{G(p)}{\max_q G(q)}.
\end{equation}
Lower scores indicate smoother background regions beneath text.

\subsubsection{User Study}
\begin{figure}[h]
    \centering
    \includegraphics[width=0.98\linewidth]{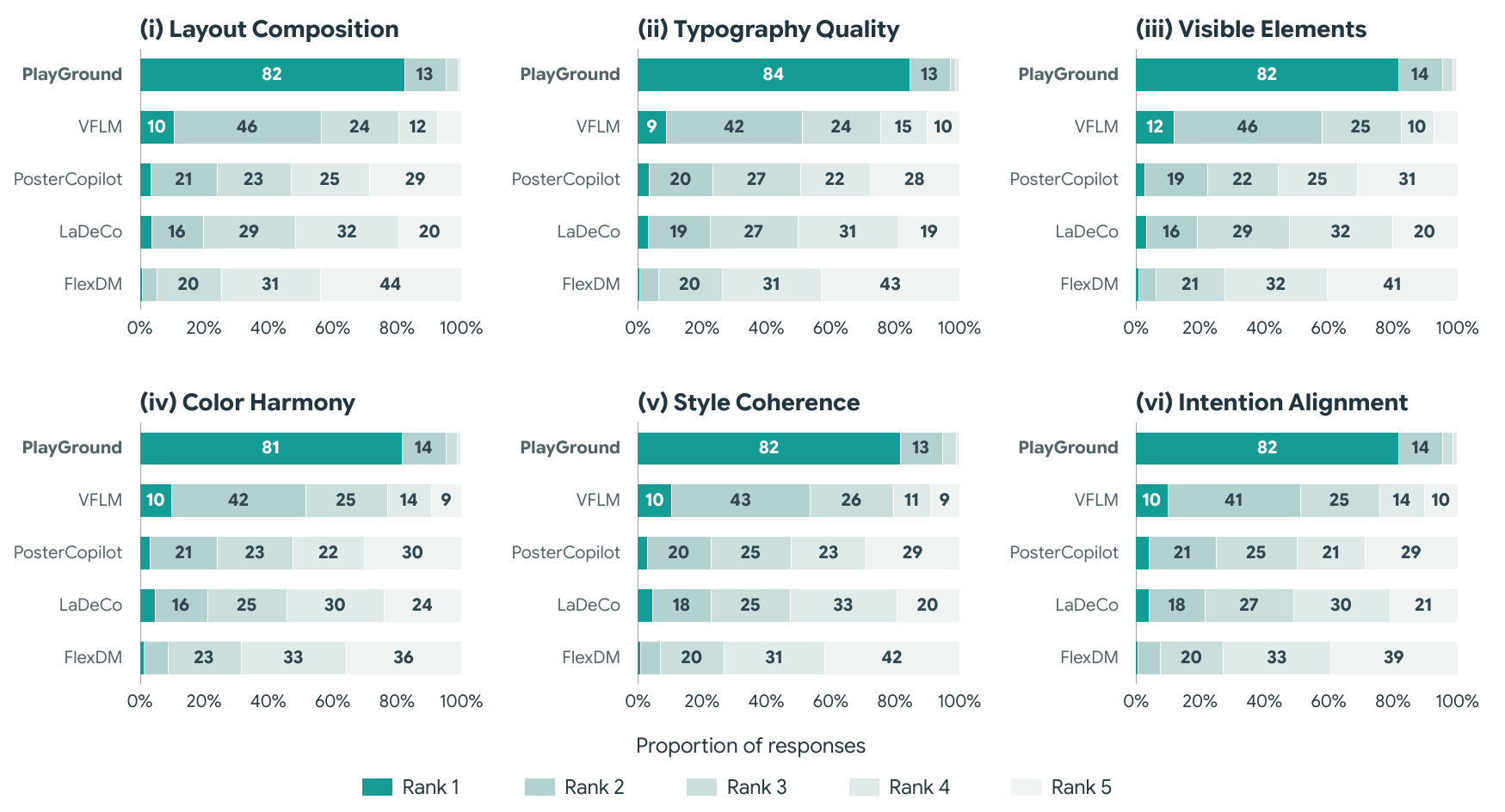}
    \vspace{-10pt}
    \caption{\textbf{User Study Results.} Human ranking distributions across six evaluation criteria.}
    \label{fig:user_study}
    \vspace{-15pt}
\end{figure}

We conducted a user study comparing \ours against four baselines: FlexDM~\citep{inoue2023flexiblemultimodaldocumentmodels}, LaDeCo~\citep{lin2024elementsdesignlayeredapproach}, PosterCopilot~\citep{wei2025postercopilotlayoutreasoningcontrollable}, and VFLM~\citep{guo2026seeingimprovingvisualfeedback}. The study involved 32 participants and 25 randomly selected samples from the Crello-v4 test split. For each evaluated sample, participants ranked the five methods' outputs from 1 (best) to 5 (worst) independently for each criterion. We used the six criteria defined for our VLM-based evaluation in Sec.~\ref{sec:vlm_metrics}: (i) layout composition rationality, (ii) typography legibility and hierarchy, (iii) visible element presentation, (iv) color harmony and contrast, (v) visual style coherence, and (vi) design intention alignment.

The user study results are presented in Fig.~\ref{fig:user_study}. \ours ranks first in approximately 81--84\% of responses and among the top two in approximately 95--97\% across all six criteria. Notably, \ours jointly arranges both visual and textual elements yet outperforms VFLM, which places only text on a fixed ground-truth background. We attribute this advantage in part to progressive planning from the current rendered canvas, which allows text placement to adapt to available space and previously placed elements. The human evaluation results are also consistent with the VLM-based results reported in Table~\ref{tab:main_quan}.

\subsection{Ablation Settings}
\label{sec:ablation_settings}
\paragraph{Evaluation subset.}
We conduct the ablation studies on a balanced subset of 600 samples from the Crello-v4 test split. The dataset is divided into six design categories, and we sample 100 designs from each category. Each selected sample contains at least eight elements.

\paragraph{Single-candidate verification.}
For experiments with $K=1$, candidate comparison is no longer applicable.
We therefore use a dedicated Visual Verifier prompt that directly evaluates the
single rendered candidate and decides between \texttt{accept} and
\texttt{retry\_with\_feedback}.
This preserves the verifier's feedback-and-regeneration behavior while removing
its multi-candidate selection role.

\section{Additional Qualitative Results}
\label{sec:additional_qualitative_results}
\begin{figure}[p]
    \centering
    \includegraphics[width=0.98\linewidth]{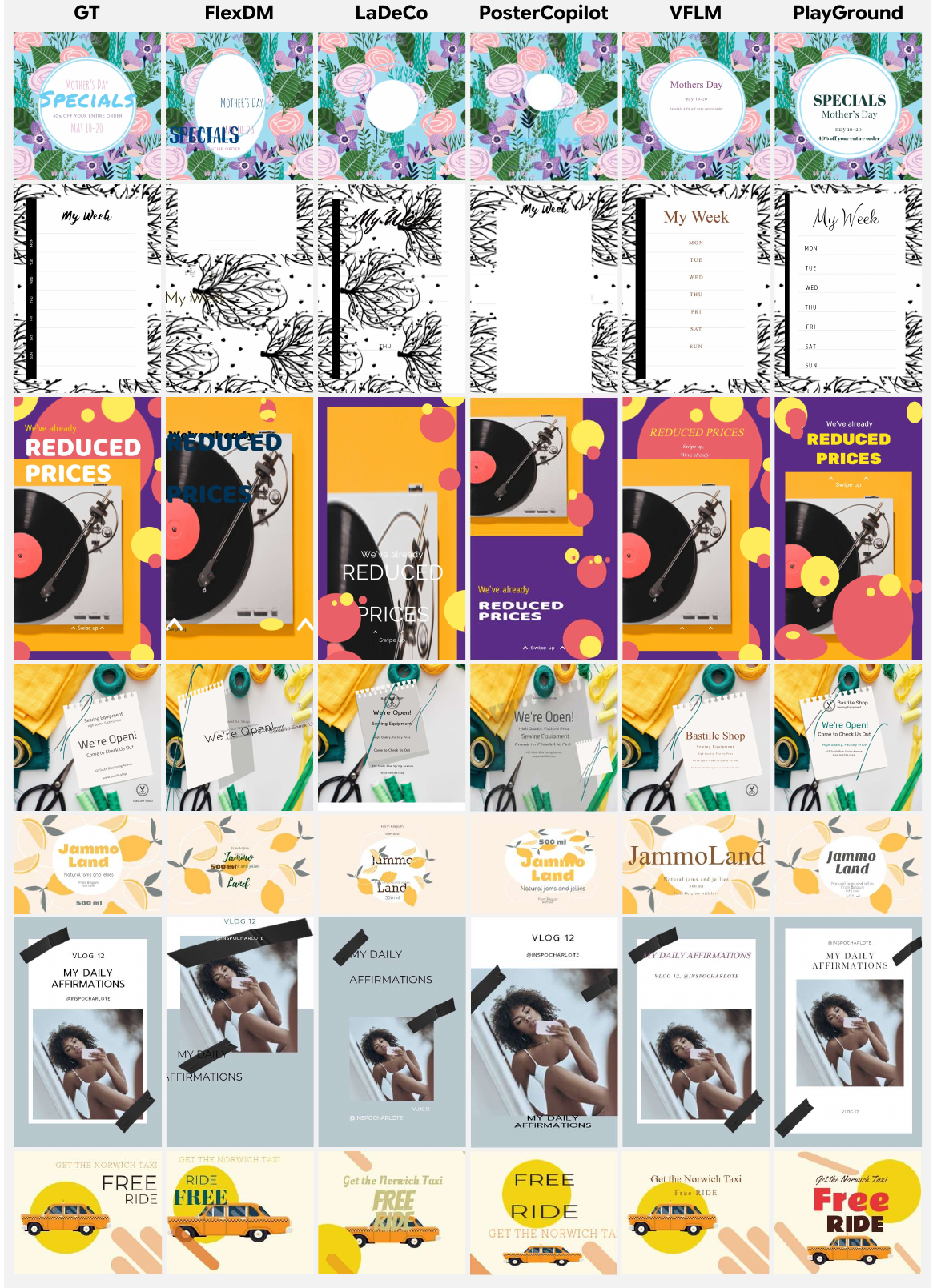}
    \vspace{-5pt}
    \caption{\textbf{More qualitative results on Crello dataset.}}
    \label{fig:qaul_crello_1}
    \vspace{-10pt}
\end{figure}

\begin{figure}[p]
    \centering
    \includegraphics[width=0.98\linewidth]{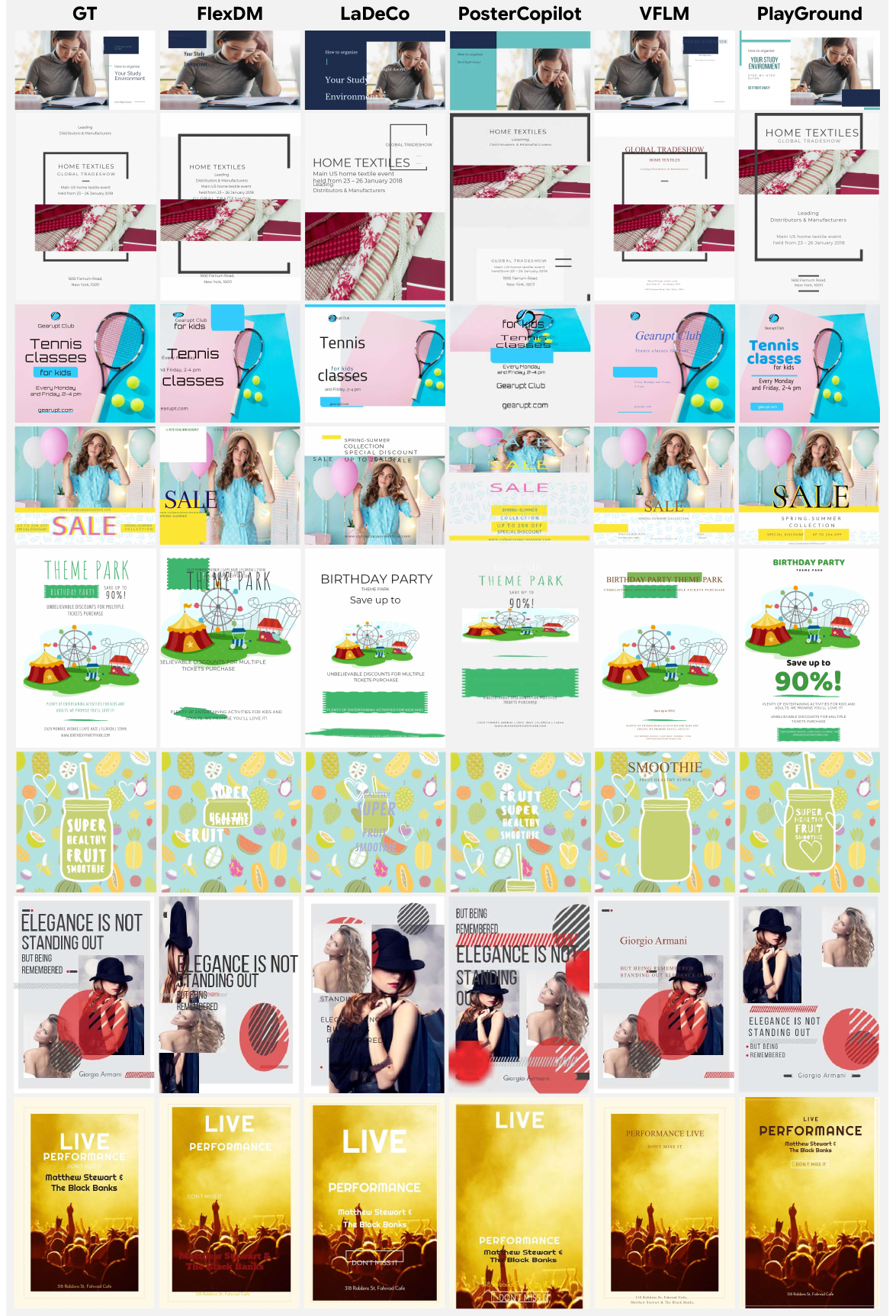}
    \vspace{-5pt}
    \caption{\textbf{More qualitative results on Crello dataset.}}
    \label{fig:qaul_crello_2}
    \vspace{-10pt}
\end{figure}

\begin{figure}[p]
    \centering
    \includegraphics[width=0.98\linewidth]{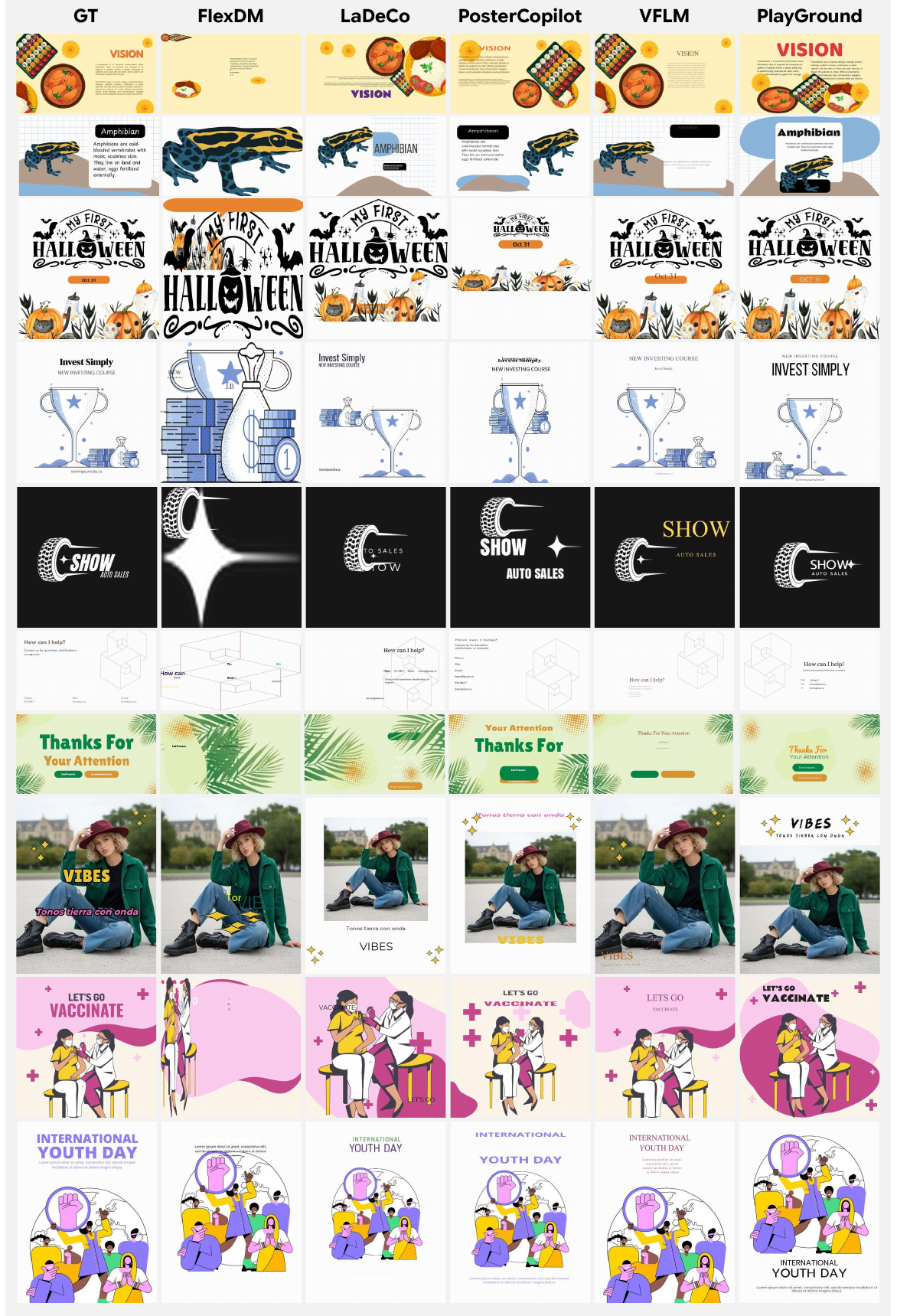}
    \vspace{-5pt}
    \caption{\textbf{More qualitative results on LICA dataset.}}
    \label{fig:qaul_lica_3}
    \vspace{-10pt}
\end{figure}

We present additional qualitative comparisons on Crello-v4 and LICA datasets between \ours and other baselines in Fig.~\ref{fig:qaul_crello_1}--\ref{fig:qaul_lica_3}.

\section{Extension to Intention-to-Design Task}
\label{sec:intention_to_design}
Our main setting assumes that the visual and textual elements to be composed are
provided in advance. In practice, however, users may specify only a design
intention without having the corresponding elements. To examine whether \ours can
operate in this setting, we extend it to intention-to-design generation, where
the required elements are first generated from the intention and then composed
into a final editable design.

To do so, we prepend an element-generation stage to \ours. Given a design
intention, GPT-5.6 Luna~\citep{openai2026gpt56} first derives the required textual
elements and descriptions of the visual elements. Qwen-Image-2.1~\citep{qwenimage21}
then generates the corresponding visual elements from these descriptions. The resulting
visual and textual elements are then provided directly to \ours, which performs
its standard functional grouping and progressive layout generation without
modifying the core composition pipeline.

\begin{figure}[p]
    \centering
    \includegraphics[width=0.98\linewidth, height=20cm]{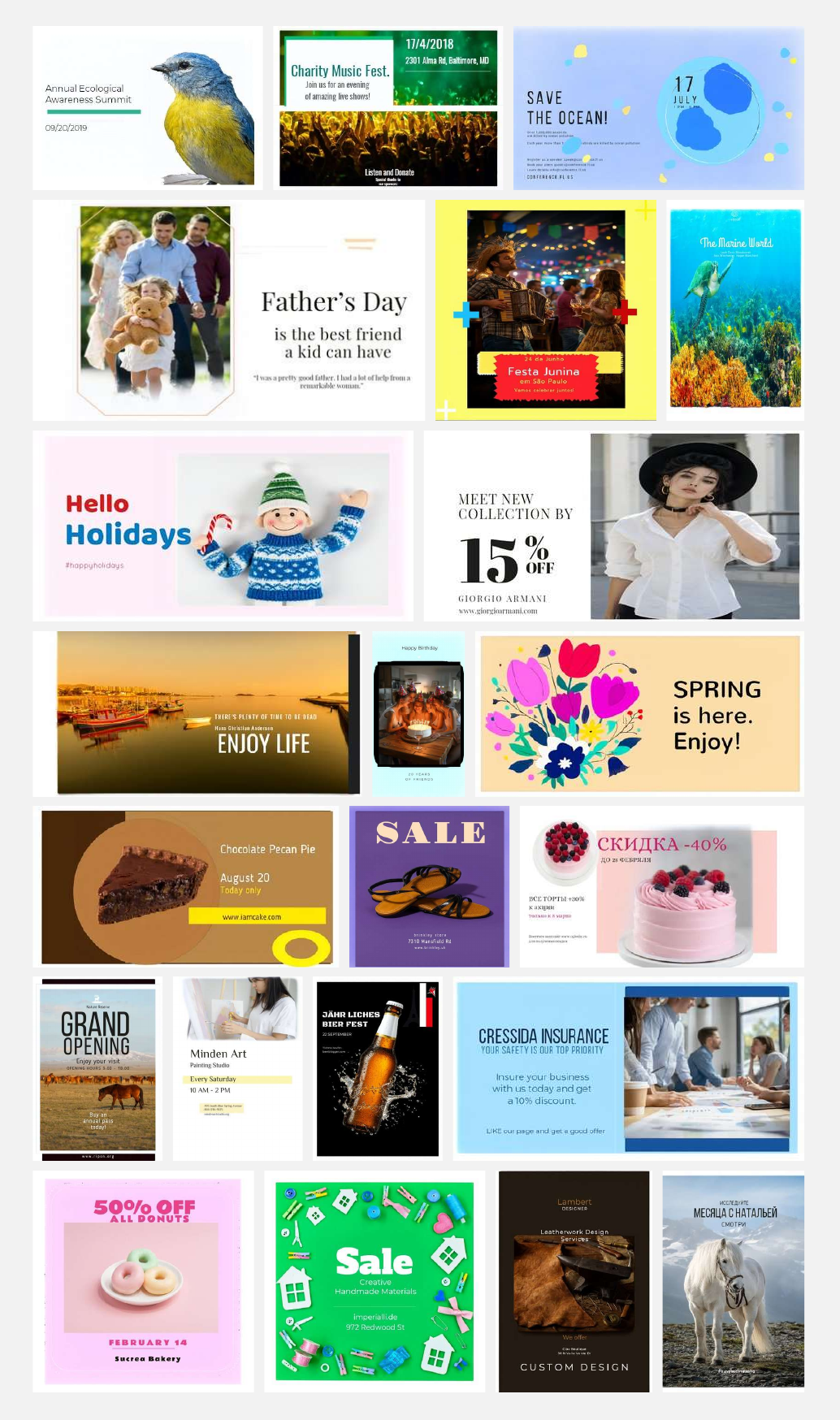}
    \vspace{-5pt}
    \caption{\textbf{Qualitative results of \ours on the intention-to-design task.} 
    % Visual and textual elements are first generated from the input intention and then composed by \ours into an editable design.
    }
    \label{fig:i2d_qual}
    \vspace{-10pt}
\end{figure}

We apply this extension to Crello, with qualitative examples shown in
Fig.~\ref{fig:i2d_qual}. These results demonstrate the modular extensibility of \ours, showing that the same composition model can be reused as a downstream design module when the required elements are produced by an external generation stage.

% \begin{vlmpromptbox_i2d}
%     \input{suppl/prompt/i2d_prompt}
% \end{vlmpromptbox_i2d}


\end{document}